\documentclass[num-refs]{wiley-article}

\usepackage{graphicx}
\usepackage[space]{grffile}
\usepackage{latexsym}
\usepackage{textcomp}
\usepackage{longtable}
\usepackage{tabulary}
\usepackage{booktabs,array,multirow}
\usepackage{amsfonts,amsmath,amssymb}
\usepackage{natbib}
\usepackage{url}
\usepackage{hyperref}
\hypersetup{colorlinks=false,pdfborder={0 0 0}}
\usepackage{etoolbox}
\usepackage{xspace}
\makeatletter

\DeclareRobustCommand\onedot{\futurelet\@let@token\@onedot}
\def\@onedot{\ifx\@let@token.\else.\null\fi\xspace}

\def\etal{\emph{et al}\onedot}
\makeatother

\newif\iflatexml\latexmlfalse

\AtBeginDocument{\DeclareGraphicsExtensions{.pdf,.PDF,.png,.png,.png,.PNG,.tif,.TIF,.jpg,.JPG,.jpeg,.JPEG,.svg,.SVG}}

\usepackage[utf8]{inputenc}
\usepackage[english]{babel}

\usepackage{siunitx}
\usepackage{subcaption}
\usepackage{svg}

\newcommand{\mysecref}[1]{Section~\ref{#1}}
\newcommand{\myfigref}[1]{Figure~\ref{#1}}
\newcommand{\myeqnref}[1]{Equation~(\ref{#1})}
\newcommand{\mytabref}[1]{Table~\ref{#1}}

\iflatexml

\else

\paperfield{Near-Infrared Hyperspectral Imaging for Wheat Protein Regression and Grain Variety Classification}
\abbrevs{Acc, Accuracy; CNN, Convolutional Neural Network; CV, Cross-Validation; HSI, Hyperspectral Imaging; NIR, Near-Infrared; OLS-R, Ordinary Least Squares Regression; PLS-DA, Partial Least Squares Discriminant Analysis; PLS-R, Partial Least Squares Regression; RGB, Red, Green, and Blue; RMSE, Root Mean Squared Error; SEM, Standard Error of The Mean; SG, Savitzky-Golay; SNV, Standard Normal Variate; sYX, Standard Error of Estimate.}
\corraddress{Ole-Christian Galbo Engstr\o m, Department of Computer Science, University of Copenhagen Faculty of Science, Universitetsparken 1, 2100 Copenhagen, Denmmark.}
\corremail{ocge@foss.dk}
\presentadd{R\&D, Foss Analytical A/S, Nils Foss Allé 1, 3400 Hiller\o d, Denmark.}
\fundinginfo{FOSS Analytical A/S; Innovation Fund Denmark, Grant Number: 1044-00108B.}
\fi

\papertype{Technical Report}

\title{Analyzing Near-Infrared Hyperspectral Imaging for Protein Content Regression and Grain Variety Classification Using Bulk References and Varying Grain-to-Background Ratios}

\author[1,2,3]{Ole-Christian Galbo Engstr\o m}
\author[2]{Erik Schou Dreier}
\author[3]{Birthe M\o ller Jespersen}
\author[1,4]{Kim Steenstrup Pedersen}

\affil[1]{Department of Computer Science (DIKU), University of Copenhagen, Denmark.}
\affil[2]{FOSS Analytical A/S, Denmark.}
\affil[3]{Department of Food Science (UCPH FOOD), University of Copenhagen, Denmark.}
\affil[4]{Natural History Museum of Denmark (NHMD), University of Copenhagen, Denmark.}

\runningauthor{Engstr\o m \etal}

\begin{document}

\maketitle
\selectlanguage{english}
\begin{abstract}
Based on previous work, we assess the use of NIR-HSI images for calibrating models on two datasets, focusing on protein content regression and grain variety classification. Limited reference data for protein content is expanded by subsampling and associating it with the bulk sample. However, this method introduces significant biases due to skewed leptokurtic prediction distributions, affecting both PLS-R and deep CNN models. We propose adjustments to mitigate these biases, improving mean protein reference predictions. Additionally, we investigate the impact of grain-to-background ratios on both tasks. Higher ratios yield more accurate predictions, but including lower-ratio images in calibration enhances model robustness for such scenarios.

\end{abstract}

\section{Introduction}
{\label{sec:introduction}}
Hyperspectral imaging (HSI) is an imaging technique that extracts spatial and spectral features where each pixel contains an entire wavelength spectrum. The hyperspectral image can be analyzed with deep learning models for image analysis where traditionally convolutional neural networks (CNNs) excel \cite{o2020deep, rawat2017deep}. Alternatively, from relevant spatial locations, spectral information can be extracted from the hyperspectral image for subsequent analysis with chemometric methods such as partial least-squares \cite{wold1966estimation}, which can be implemented for either regression (PLS-R) \cite{wold1983food, wold2001pls} or classification through discriminant analysis (PLS-DA) \cite{barker2003partial}.

This work concerns near-infrared hyperspectral imaging (NIR-HSI) for protein content regression analysis of wheat grain kernels and grain variety classification for wheat and rye grain kernels. Protein content regression is a typical chemometric problem that lends itself naturally to PLS-R due to the linear relationship between spectral absorbance in the near-infrared (NIR) wavelengths and protein content \cite{osborne2006near}. However, deep learning for image analysis performs well within agricultural domains \cite{benos2021machine, kamilaris2018deep, liakos2018machine, meshram2021machine, ren2020survey}. Grain variety classification can be solved with spectral features using PLS-DA or PLS2-DA or spatial features alone using CNNs, but simultaneously utilizing spectral and spatial features will increase performance \cite{engstrom2023improving, dreier2022hyperspectral}. Recently, CNNs utilizing the entire spatio-spectral data cube have been applied to grain quality analysis \cite{dreier2022hyperspectral, engstrom2023improving, engstrom2021predicting} with Engstrøm \etal~\cite{engstrom2021predicting} designing domain-specific CNNs for protein content prediction on NIR-HSI images of bulk wheat kernel samples and Dreier \etal~\cite{dreier2022hyperspectral} designing CNNs for grain variety classification. Engstrøm \etal~\cite{engstrom2023improving} improve upon both results and compare them with those achievable by PLS-R, PLS-DA, and PLS2-DA, respectively.

Engstrøm \etal~\cite{engstrom2023improving} show that a domain-specific CNN can outperform a PLS-R model - even when the latter is subject to an extensive search for the best preprocessing method. However, linear models such as PLS-R have a significant advantage over deep learning models such as CNNs because they can be calibrated on significantly less data. In comparison, deep learning methods require substantial amounts of data to converge meaningfully. Obtaining reference values for the images can be a costly and time-consuming process. It is particularly costly to obtain accurate reference values for the protein contents of individual grain kernels as the protein content may vary by several percentage points between each kernel - even when they originate from the same ear of the plant. Data augmentation techniques can be applied to inflate the training data, thus mitigating the need to analyze the protein content of every grain kernel. For protein content regression, one such approach is to measure the mean protein content of an entire bulk sample and then use the mean protein value as the reference for each bulk subsample originating from that bulk sample. Unless the bulk sample consists of grain kernels with identical protein values, the mean protein value of the bulk sample is unlikely to be equal to the mean protein value of the bulk subsamples, yielding inaccurate reference values for the training data. However, the inaccuracies will be two-sided; on average, the mean reference value will be correct per definition. Engstrøm \etal~\cite{engstrom2023improving, engstrom2021predicting} have shown that this approach allows for meaningful convergence of both CNNs and PLS-R.

Dreier \etal~\cite{dreier2022hyperspectral} show that models for grain variety classification trained on images with a high grain density, that is the grain-to-background ratio, $(\geq 0.5)$ fail to accurately predict on images with a lower grain density $(<0.5)$ than that which they were calibrated on. Consequently, it is interesting to see how grain variety classification models perform on a lower grain density when calibrated on these. Engstrøm \etal~\cite{engstrom2021predicting} use a grain density of $0.1$ for models calibrated to regress protein content and models calibrated to classify grain variety alike.
 
In this work, we delve deep into the results achieved by Engstrøm \etal~\cite{engstrom2023improving}. We use statistical approaches to investigate the implications of using the bulk mean protein value as a reference for subsamples when calibrating regression models, revealing significant biases in the predictions of both CNNs and PLS-R models due to the shapes of the predicted protein distributions. We also offer two approaches for the mitigation of these biases - one approach for PLS-R only and one approach for both CNNs and PLS-R. Additionally, we analyze the effects of grain density for both protein content regression using the mean bulk reference and grain variety classification. We show that for both tasks, a higher grain density is preferable, but in the case of grain variety classification, calibrating the model with a mixture of low and high grain density samples will allow it to predict much more accurately the low grain density samples while achieving accuracy on high grain density samples comparable with previous work.

This report presents the dataset and models in \mysecref{sec:dataset_and_models} and a nomenclature in \mysecref{sec:nomenclature}. In \mysecref{sec:mean_reference}, we analyze the implications of using the bulk sample mean protein content as a reference for predictions on subsamples of the bulk. Here, we also show how to correct the subsequent bulk mean protein content predictions linearly, yielding significantly better bulk mean protein content predictions. In \mysecref{sec:distributions}, our analysis of the distributions of predicted protein content in the bulk subsamples reveals why using the bulk sample mean as a reference for its subsamples leads to biased results. We analyze the effects of the grain density for both protein content regression and grain variety classification in \mysecref{sec:grain_density}. All results are concluded in \mysecref{sec:conclusion}.

\section{Datasets and models}
\begin{figure}[htbp]
    \centering
    \includegraphics[width=0.8\linewidth]{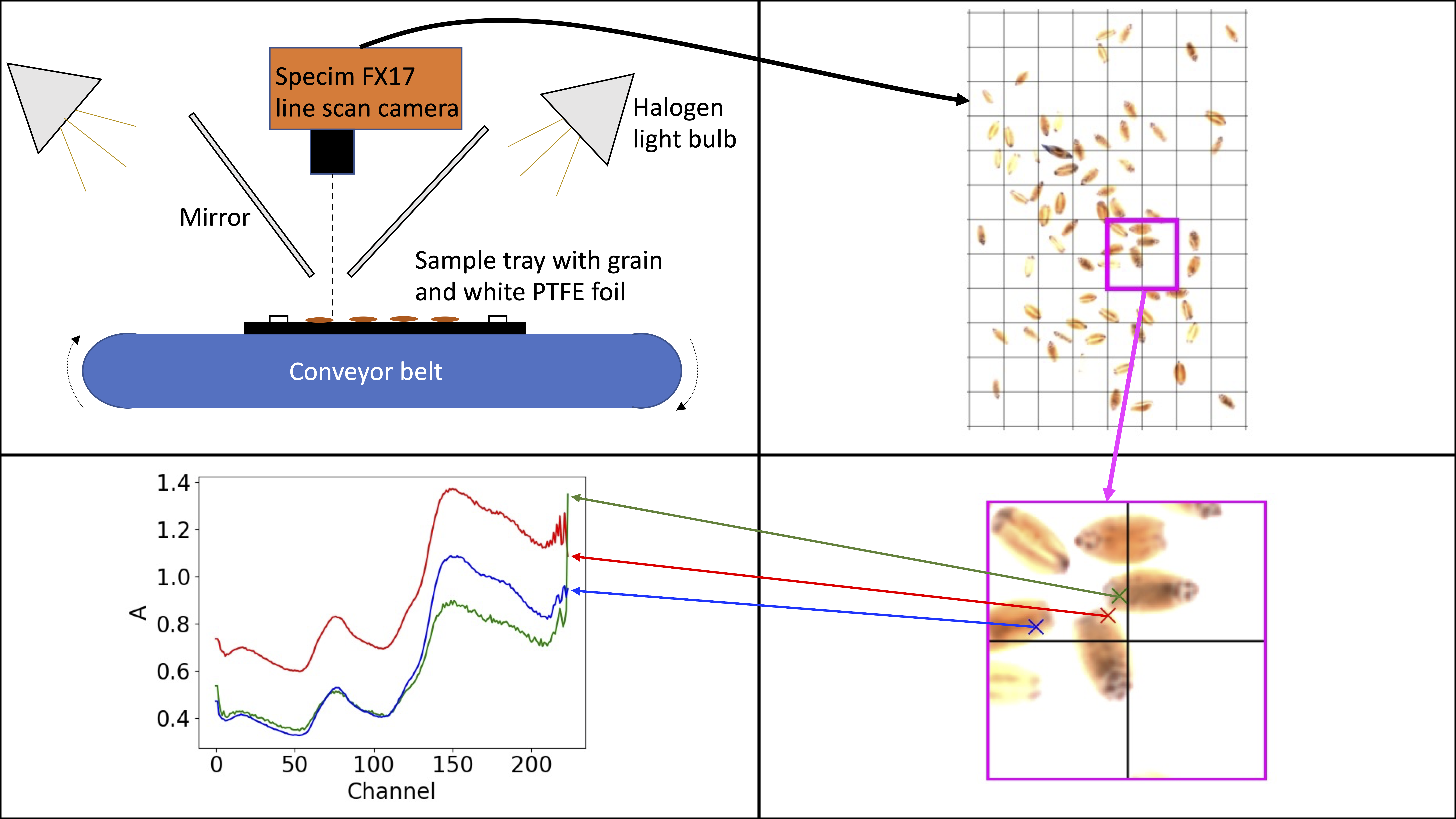}
    \caption{The hyperspectral imaging setup. This figure shows, clockwise from top-left, the imaging setup, a hyperspectral image of sparsely packed grain from one camera acquisition, a $128 \times 128$ pixels hyperspectral image crop, and the pseudo absorbance spectra of three pixels containing grain. The image is taken from Engstrøm \etal~\cite{engstrom2021predicting}.}
    \label{fig:imaging_setup}
\end{figure}

\begin{figure}[htbp]
    \centering
    \includegraphics[width=0.25\linewidth]{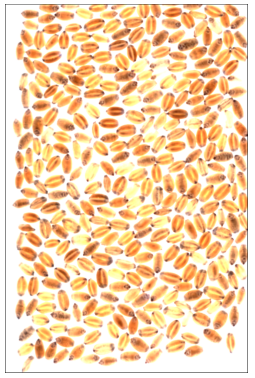}
    \includegraphics[width=0.25\linewidth]{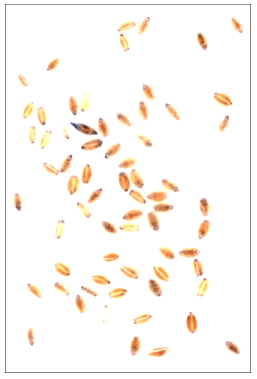}
    \caption{This figure shows examples of densely (left) and sparsely (right) packed grain images. Each of the $63$ bulk samples is distributed along $8$ sparsely packed images and approximately $10-14$ densely packed images \cite{engstrom2021predicting}. The image is taken from Engstrøm \etal~\cite{engstrom2021predicting}.}
    \label{fig:GrainImages}
\end{figure}

\begin{figure}[tbp]
    \centering
    \includegraphics[width=0.4\linewidth]{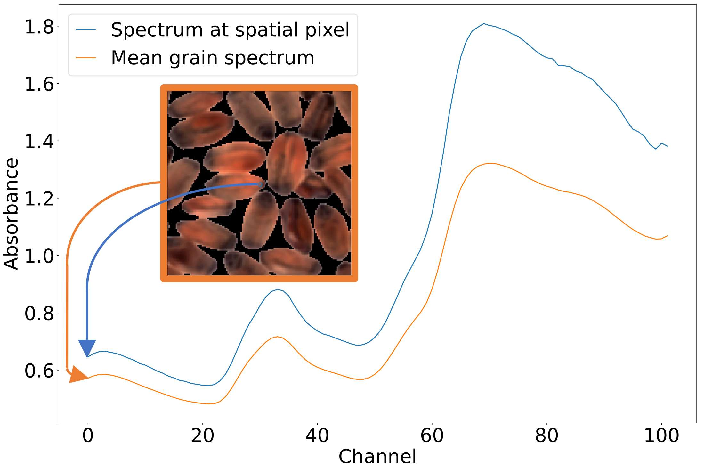}
    \caption{A masked hyperspectral image crop of spatial size $128 \times 128$ pixels along with the spectrum at one of the pixels and the mean spectrum of the crop computed over all the grain pixels, both containing $102$ wavelength channels. Taken from Engstrøm \etal~\cite{engstrom2023improving}. The discrepancy between grain colors in this Figure and those in \myfigref{fig:imaging_setup} and \myfigref{fig:GrainImages} are due to differences in the calculations that transform hyperspectral images to pseudo-RGB and are only for visualization purposes.}
    \label{fig:grain_and_spectra_image}
\end{figure}

{\label{sec:dataset_and_models}}
We use the two datasets of hyperspectral images of bulks of grain kernels captured with a Specim FX17 near-infrared hyperspectral line scan camera \cite{SpecimFX17} introduced in \cite{engstrom2021predicting} and \cite{dreier2022hyperspectral}, respectively. \myfigref{fig:imaging_setup} shows the imaging setup used by Engstrøm \etal~\cite{engstrom2021predicting}. Dreier \etal~\cite{dreier2022hyperspectral} use a similar setup. The two datasets contain images of densely and sparsely packed bulk grain samples. An example of a densely and a sparsely packed image from Engstrøm \etal~\cite{engstrom2021predicting} is shown in \myfigref{fig:GrainImages}. We preprocess the images according to the strategy devised by Engstrøm \etal~\cite{engstrom2023improving}. This preprocessing includes cropping the images in sizes of $128 \times 128$ pixels with an overlap of $64$ pixels in both spatial dimensions, spectral binning to $102$ wavelength channels, and discarding any image crops where less than 10\% of the pixels contain grain. Additionally, a mean grain spectrum is associated with each image crop, which is used for calibrating PLS. The mean grain spectrum is computed over the pixels that contain grain as determined by a binary segmentation mask using an Otsu threshold \cite{Otsu79a}. Initially, the hyperspectral image contains $224$ uniformly distributed wavelength channels. However, due to a reduced sensitivity in the camera's sensors near the edges of its spectrum \cite{SpecimFX17}, the first and last $10$ wavelength channels are discarded. Additionally, the spectral channels are averaged in neighboring pairs as the camera's average spectral resolution is $8$ nm, with its sensors sampling at every $3.5$ nm \cite{SpecimFX17}. Thus, after preprocessing, the spectral dimension contains $102$ uniformly distributed wavelength channels in the $938-1662$ nm part of the NIR spectrum, which contains molecular vibration bands relevant to grain quality analysis \cite{osborne2006near}, in particular to the protein content that is of interest to this study \cite{caporaso2018protein, holford1992nitrogen, strong1981nitrogen}. \myfigref{fig:grain_and_spectra_image} shows a hyperspectral image crop with its associated mean grain spectrum and the grain spectrum at an arbitrary grain pixel.

The first dataset, Dataset \#1, is of wheat grain kernels, where the associated task is to regress the protein content in the grain kernels. The dataset has associated bulk mean protein reference values measured with a FOSS Infratec\texttrademark Nova \cite{InfratecNova}. This dataset consists of $63$ distinct bulk samples of wheat grain kernels, of which $50$ are used for five-fold cross-validation and $13$ for testing. The subsequent cropping yields $69,630$ image crops for five-fold cross-validation and $17,783$ for testing. The second dataset, Dataset \#2, comprises seven wheat varieties and a single rye variety. Each image contains exactly a single variety. The task of this dataset is to classify the grain variety on a given image crop. \mytabref{tab:dataset1_splits} and \mytabref{tab:dataset2_splits} summarize the datasets.

\begin{table}[]
\centering
\begin{tabular}{@{}lll@{}}
\toprule
\rowcolor[HTML]{C0C0C0} 
Dataset split & \# bulk subsamples in total (and per split) & \# bulk samples in total (and per split) \\ \midrule
5-fold CV     & 69,630 (13,734 / 13,676 / 14,672 / 13,618 / 13,930)            & 50 (10 / 10 / 10 / 10 / 10)             \\
Test          & 17,783             & 13             
\end{tabular}
\caption{Dataset  \#1. The number of bulk subsamples (image crops and their mean grain spectra) and bulk samples in total and for each dataset split. There is only one test split.}
\label{tab:dataset1_splits}
\end{table}

\begin{table}[]
\centering
\begin{tabular}{@{}ll@{}}
\toprule
\rowcolor[HTML]{C0C0C0} 
Dataset split & \# bulk subsamples \\ \midrule
Training    & 15,376 \\
Validation  & 7,967  \\
Test        & 3,274  \\
\end{tabular}
\caption{The number of bulk subsamples (image crops and their mean grain spectra) and bulk samples for each dataset split. Each split contains all eight grain varieties. The distribution of grain varieties is approximately uniform in each dataset split.}
\label{tab:dataset2_splits}
\end{table}

While Engstrøm \etal~\cite{engstrom2023improving} experiment with a plethora of preprocessing techniques for PLS-R and PLS2-DA and model designs for CNNs, we focus our attention on the best performing combination of preprocessing and model for each of PLS-R, PLS2-DA, and CNN. For PLS-R on spectral data from Dataset \#1, the best preprocessing technique was found to be applying Standard Normal Variate (SNV) \cite{barnes1989standard} followed by the application of a Savitzky-Golay (SG) filter \cite{savitzky1964smoothing} with a window-size of $7$, a polynomial order of $2$ and a derivative order of $2$. Additionally, for PLS-R, the protein reference values are centered around the mean protein reference value of the training splits. The best preprocessing technique for PLS2-DA on spectral data from Dataset \#2 was applying only a spectral centering (Center).

For each dataset, the best-performing CNN is a modification of ResNet-18 \cite{he2016deep}. For Dataset \#1, the modification involves adding two initial layers, with the first one serving to learn a spectral preprocessing technique much similar to that of an SG filter and the second layer reducing the spectral dimensionality by downsampling it through a learnable weighted combination of the wavelength channels output by the first layer. For Dataset \#2, the modification involves adding only one initial layer, namely the layer that learns to downsample the spectral dimensionality. Engstrøm \etal~\cite{engstrom2023improving} explain the architectures in detail. They named the models ResNet-18 Modification \#10 for Dataset \#1 and ResNet-18 Modification \#8 for Dataset \#2. Moving forward, in this report, we will refer to these models by Modified ResNet-18 Regressor and Modified ResNet-18 Classifier, respectively.

Engstrøm \etal~\cite{engstrom2023improving} use five-fold cross-validation for calibrating their models on Dataset \#1. They show that Modified ResNet-18 Regressor performance can be improved on the test set by constructing an ensemble model that outputs an arithmetic mean of all five cross-validated models. Constructing an ensemble for the PLS-R yields neither better nor worse results on the test set than using an individual PLS-R model. For consistency, we focus our work with Dataset \#1 on the mean model predictions that are the outputs of the ensemble Modified ResNet-18 Regressor and the ensemble PLS-R while still providing SEMs on the ensembles' predictions to show the degree of disagreement present in the ensembles' constituents.

\section{Nomenclature}\label{sec:nomenclature}
The remaining parts of this report extensively use mathematical symbols and terms. This section contains a description of them.

The following notation is used for mathematical symbols:

\begin{align*}
    &_{i} && \text{Subscript integer index counter.}\\
    &_{\text{train}} && \text{Subscript used to reference training set data.}\\
    &_{\text{val}} && \text{Subscript used to reference validation set data.}\\
    &_{\text{test}} && \text{Subscript used to reference test set data.}\\
    &_{\text{ss}} && \text{Subscript used to indicate bulk subsample.}\\
    &_{\text{bm}} && \text{Subscript used to indicate bulk subsample means. I.e. averaging over all $_{\text{ss}}$ for each individual bulk sample.}\\
    &_{\text{bulk}} && \text{Subscript used to indicate bulk. Used only in context with PLS-R calibrated on mean spectra from bulk samples.}\\
    &n && \text{Number of samples. Will either be number of bulk samples or number of bulk subsamples dependent on context.}\\
    &\mathbf{\epsilon} && \text{Error terms vector of shape $n \times 1$.}\\
    &\mathbf{y} && \text{Reference value column vector of shape $n \times 1$.}\\
    &\hat{\mathbf{y}} && \text{Prediction value column vector of shape $n \times 1$.}\\
    &\mathbf{Y} && \text{Augmented reference value matrix of shape $n \times 2$. The first column is ones. The second column is $\mathbf{y}$.}\\
    &\hat{\mathbf{Y}} && \text{Augmented prediction value matrix of shape $n \times 2$. The first column is ones. The second column is $\hat{\mathbf{y}}$.}\\
\end{align*}

The following describes "ensemble model prediction," a term often used in our figures. Recall that, for Dataset \#1, we concern ourselves with ensemble models. The "ensemble model prediction" is a uniform average over the ensembles' constituents. Recall that the ensembles are trained with a 5-fold CV. Thus, when used in training sets, the mean model prediction is an average prediction over the four constituents with the data point(s) in question in their training split. For the same reason, "ensemble model prediction," when used in the context of validation splits, is the prediction of only the single constituent with the particular data point(s) in its validation split. When used in the context of the test split, "ensemble model prediction" is a uniform average over all five of the ensembles' as the test split is the same for all constituents.

For the reasons mentioned above, the contents of any prediction vectors and matrices associated with the training or test splits of Dataset \#1 are ensemble model predictions unless explicitly stated otherwise.

\section{Bulk Mean As Reference For Subsample Predictions}
{\label{sec:mean_reference}}
While the RMSE \myeqnref{eq:rmse} is a good measure of how well a regression model performs, it is a summary statistic that does not give any information about potential biases in the model. In particular, we are interested in whether calibrating our models on the bulk subsamples allows them to accurately predict the bulk mean reference value - i.e. if a model optimized to minimize RMSE$(\mathbf{y}, \hat{\mathbf{y}})$ is simultaneously a good model for minimizing RMSE$(\mathbf{y}_{\text{bm}}, \hat{\mathbf{y}}_{\text{bm}})$. Recall that we are only sure about the values of $\mathbf{y}_{\text{bm}}$ as these are the ones measured with the FOSS Infratec\texttrademark Nova \cite{InfratecNova} but that we distribute $\mathbf{y}_{\text{bm}, i}$ to all subsamples originating from bulk sample $i$ to inflate the amount of data available for calibration, resulting in $\mathbf{y}$.

\begin{equation}\label{eq:rmse}
    \text{RMSE}(\mathbf{y}, \hat{\mathbf{y}}) = \sqrt{\frac{1}{n}\sum_{i=0}^{n-1}\left(\mathbf{y}_i - \hat{\mathbf{y}}_i\right)^2}
\end{equation}

Ideally, the models would have been calibrated such that the predictions lie on the straight line $\mathbf{y}_{\text{bm}} = \hat{\mathbf{y}}_{\text{bm}}$. In practice, however, this may be difficult to achieve due to various sources of errors. Instead, aggregating the error terms in $\mathbf{\epsilon}$, the relationship can be written as \myeqnref{eq:lin_reg}.

\begin{equation}\label{eq:lin_reg}
    \mathbf{y}_{\text{bm}} = \hat{\mathbf{y}}_{\text{bm}} + \mathbf{\epsilon}
\end{equation}

Using RMSE as a loss function, as done by Engstrøm \etal~\cite{engstrom2023improving}, the task is to find $\hat{\mathbf{y}}_{\text{bm}}$ such that $\sum_{i=0}^{n-1}\mathbf{\epsilon}_{i}^2$ is minimized - this is a variant of linear regression known as least squares regression. To assess how well our models perform least squares regression, we compare their RMSE$(\mathbf{y}_{\text{bm}}, \hat{\mathbf{y}}_{\text{bm}})$ with their sYX$(\mathbf{y}_{\text{bm}}, \hat{\mathbf{y}}_{\text{bm}})$, defined in \myeqnref{eq:syx}. sYX uses OLS-R (\myeqnref{eq:ols-r}), a closed form solution to the least squares regression problem, to compute a bias$_\text{bm}$ and a scale$_\text{bm}$ parameter that minimize $\sum_{i=0}^{n-1}\mathbf{\epsilon}_{i}^2$ in \myeqnref{eq:lin_reg_adjusted}. As we estimate bias$_\text{bm}$ and scale$_\text{bm}$, we lose two degrees of freedom in the subsequent RMSE computation, hence the $n-2$ term in \myeqnref{eq:syx}.

\begin{equation}\label{eq:syx}
    \text{sYX}(\mathbf{y}, \hat{\mathbf{y}}) = \sqrt{\frac{1}{n-2}\sum_{i=0}^{n-1}\left(\mathbf{y}_i - \left(\hat{\mathbf{y}}_i \times \text{scale} + \text{bias}\right)\right)^2}
\end{equation}

\begin{equation}\label{eq:ols-r}
    \text{OLS-R}(\hat{\mathbf{Y}}, \mathbf{y})=\begin{bmatrix} \text{bias}\\ \text{scale} \end{bmatrix} = \left(\hat{\mathbf{Y}}^T\hat{\mathbf{Y}}\right)^{-1} \hat{\mathbf{Y}}^T \mathbf{y}
\end{equation}

\begin{equation}\label{eq:lin_reg_adjusted}
    \mathbf{y}_{\text{bm}} = \text{bias}+\text{scale}\times\hat{\mathbf{y}}_{\text{bm}} + \mathbf{\epsilon}
\end{equation}

\begin{figure}[htbp]
    \centering
    \begin{subfigure}[b]{0.33\textwidth}
        \centering
        \includegraphics[width=\textwidth]{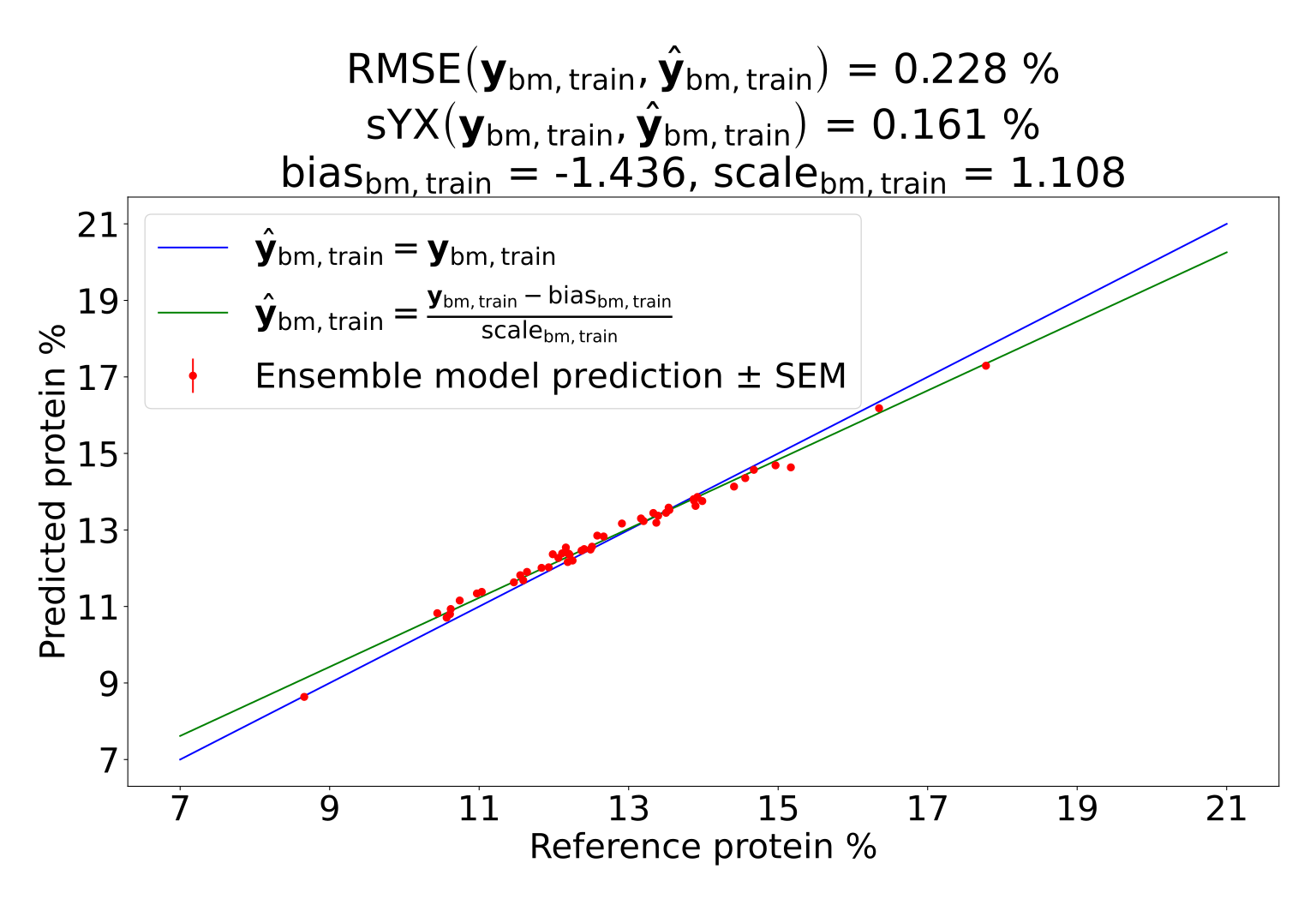}
        \caption{Modified ResNet-18 Regressor, using $\hat{\mathbf{y}}_{\text{bm,train}}$ and $\mathbf{y}_{\text{bm,train}}$.}
    \end{subfigure}
    \begin{subfigure}[b]{0.33\textwidth}
        \centering
        \includegraphics[width=\textwidth]{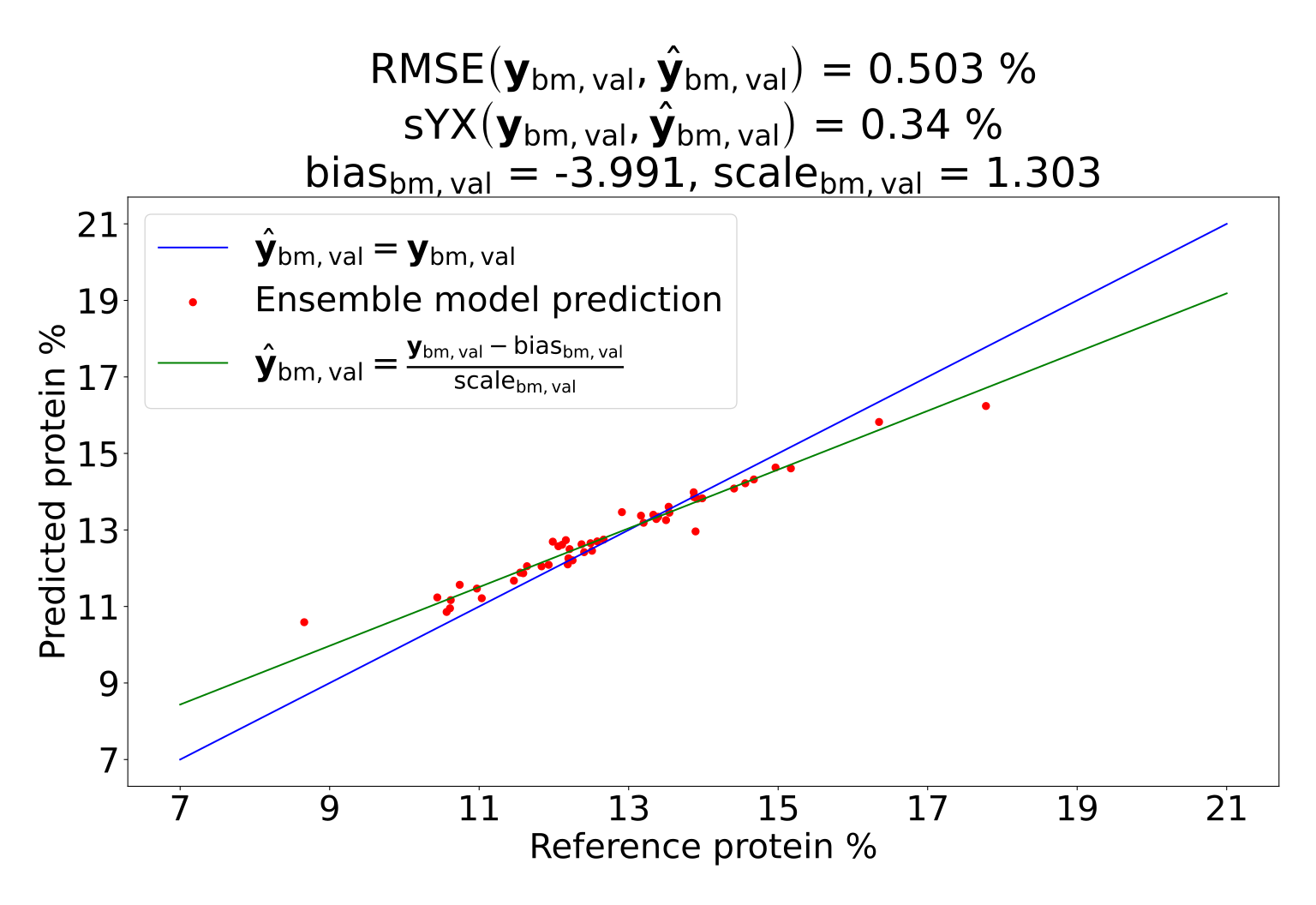}
        \caption{Modified ResNet-18 Regressor, using $\hat{\mathbf{y}}_{\text{bm,val}}$ and $\mathbf{y}_{\text{bm,val}}$.}
    \end{subfigure}
    \begin{subfigure}[b]{0.33\textwidth}
        \centering
        \includegraphics[width=\textwidth]{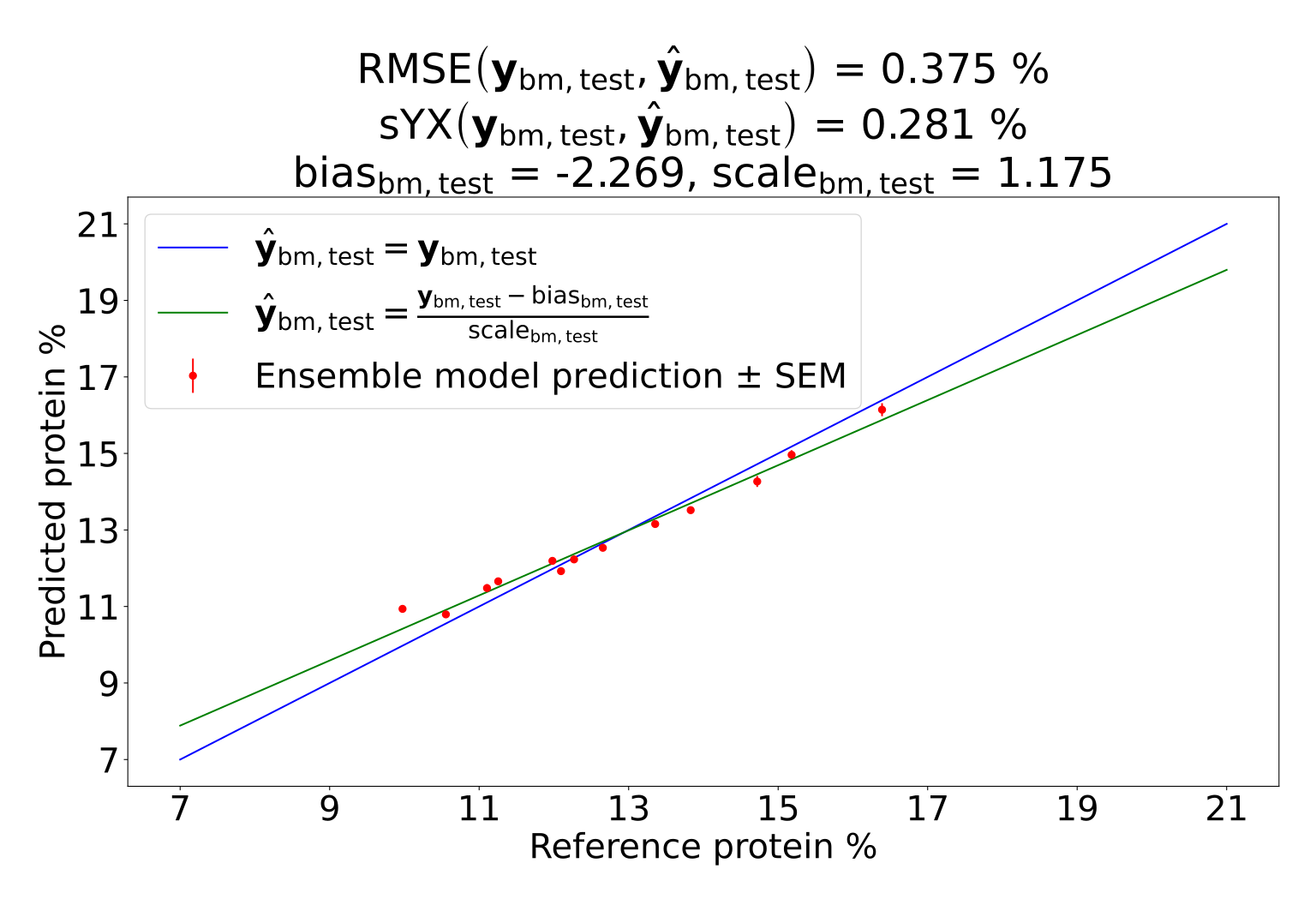}
        \caption{Modified ResNet-18 Regressor, using $\hat{\mathbf{y}}_{\text{bm,test}}$ and $\mathbf{y}_{\text{bm,test}}$.}
    \end{subfigure}
    \\
    \begin{subfigure}[b]{0.33\textwidth}
        \centering
        \includegraphics[width=\textwidth]{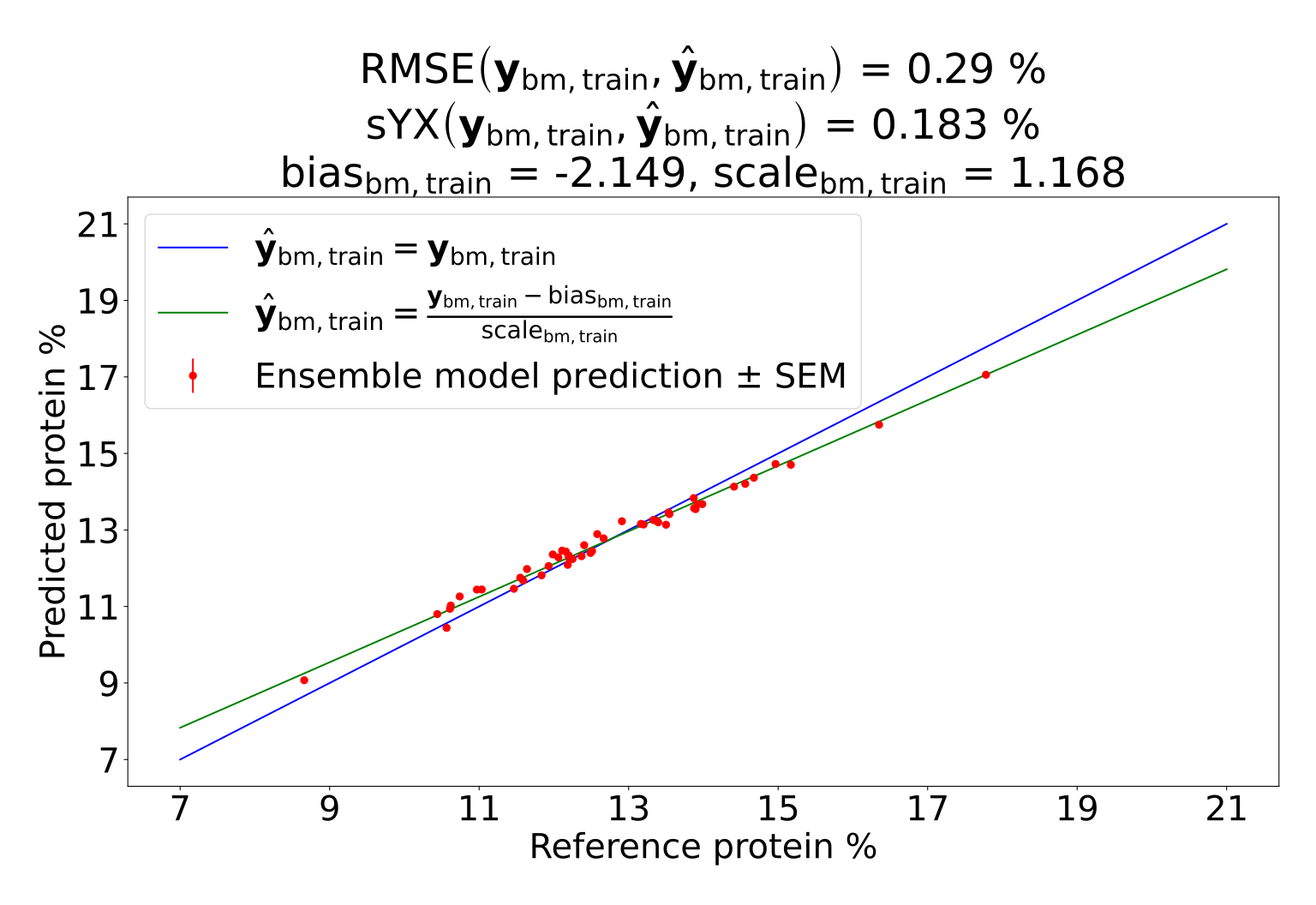}
        \caption{PLS-R, using $\hat{\mathbf{y}}_{\text{bm,train}}$ and $\mathbf{y}_{\text{bm,train}}$.}
    \end{subfigure}
    \begin{subfigure}[b]{0.33\textwidth}
        \centering
        \includegraphics[width=\textwidth]{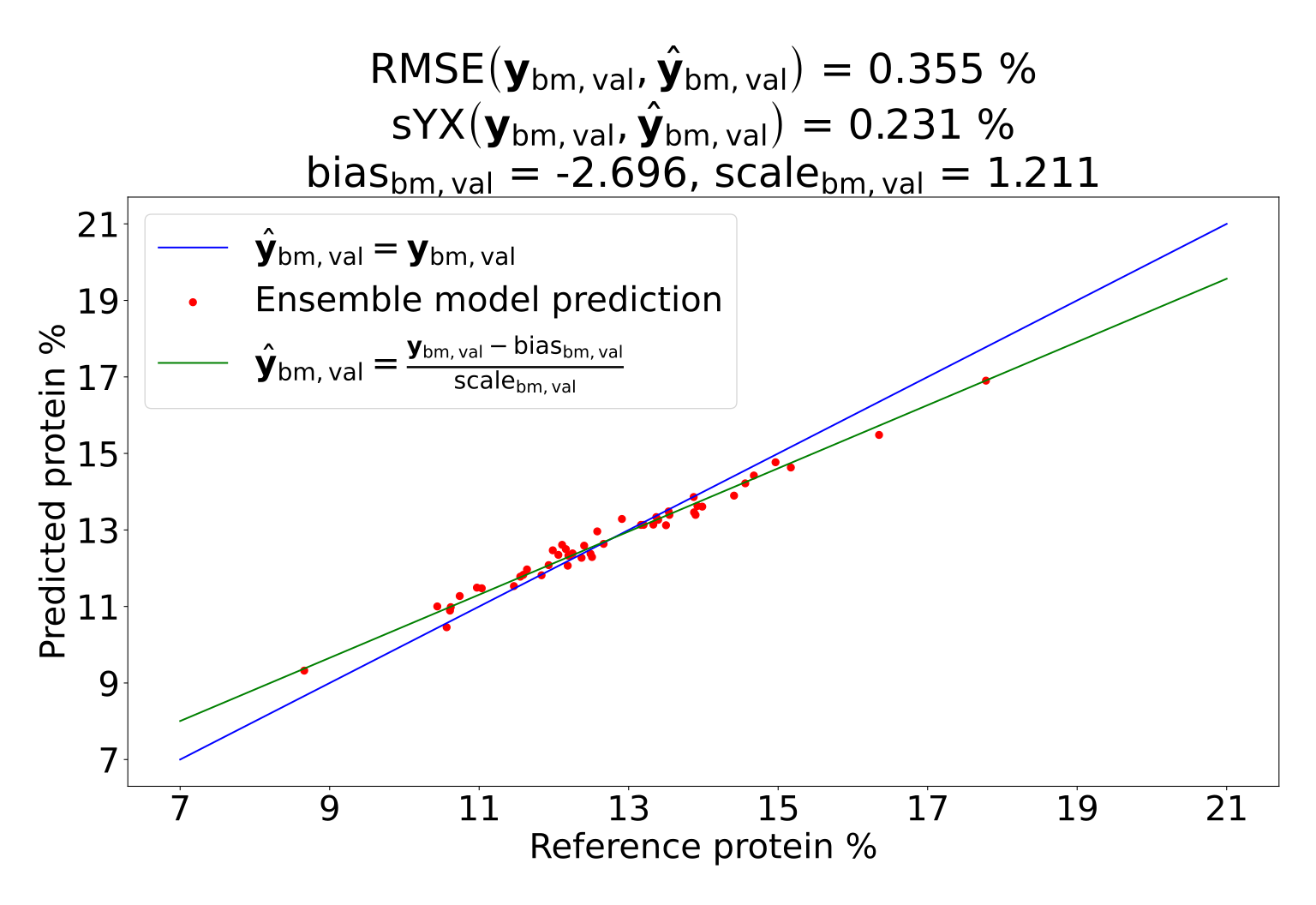}
        \caption{PLS-R, using $\hat{\mathbf{y}}_{\text{bm,val}}$ and $\mathbf{y}_{\text{bm,val}}$.}
    \end{subfigure}
    \begin{subfigure}[b]{0.33\textwidth}
        \centering
        \includegraphics[width=\textwidth]{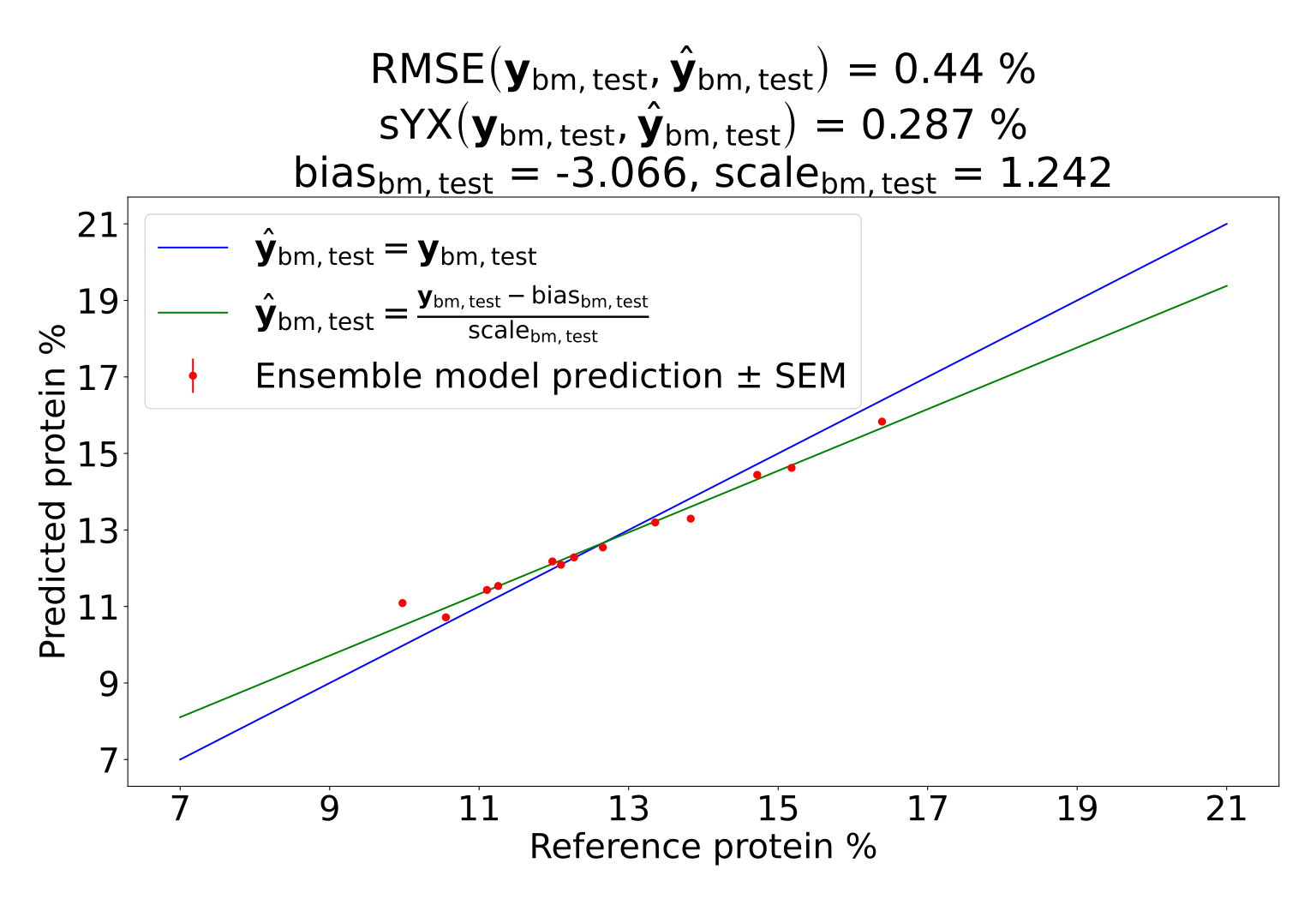}
        \caption{PLS-R, using $\hat{\mathbf{y}}_{\text{bm,test}}$ and $\mathbf{y}_{\text{bm,test}}$.}
    \end{subfigure}
    \caption{Modified ResNet-18 Regressor and PLS-R errorbar plots of $\hat{\mathbf{y}}_{\text{bm}}$ and $\mathbf{y}_{\text{bm}}$.}
    \label{fig:mean_bulk_sample_predictions_errorbar_plots}
\end{figure}

In \myfigref{fig:mean_bulk_sample_predictions_errorbar_plots} we plot $\hat{\mathbf{y}}_{\text{bm}}$ against $\mathbf{y}_{\text{bm}}$, show RMSE$(\mathbf{y}_{\text{bm}}, \hat{\mathbf{y}}_{\text{bm}})$, sYX$(\mathbf{y}_{\text{bm}}, \hat{\mathbf{y}}_{\text{bm}})$, and plot the regression line found by OLS-R$(\hat{\mathbf{Y}}_{\text{bm}}, \mathbf{y}_{\text{bm}})$. Note that bias$_\text{bm}$ and scale$_\text{bm}$ tell us how to adjust $\hat{\mathbf{y}}_{\text{bm}}$ so that it lies on a line that best matches $\mathbf{y}_{\text{bm}}$. However, for visualization purposes, it is clearer to show the equivalent, inverse relationship - i.e., how to adjust $\mathbf{y}_{\text{bm}}$ so that its line best matches $\hat{\mathbf{y}}_{\text{bm}}$ which is what we have plotted.

\begin{figure}[htbp]
    \centering
    \begin{subfigure}[b]{0.33\textwidth}
        \centering
        \includegraphics[width=\textwidth]{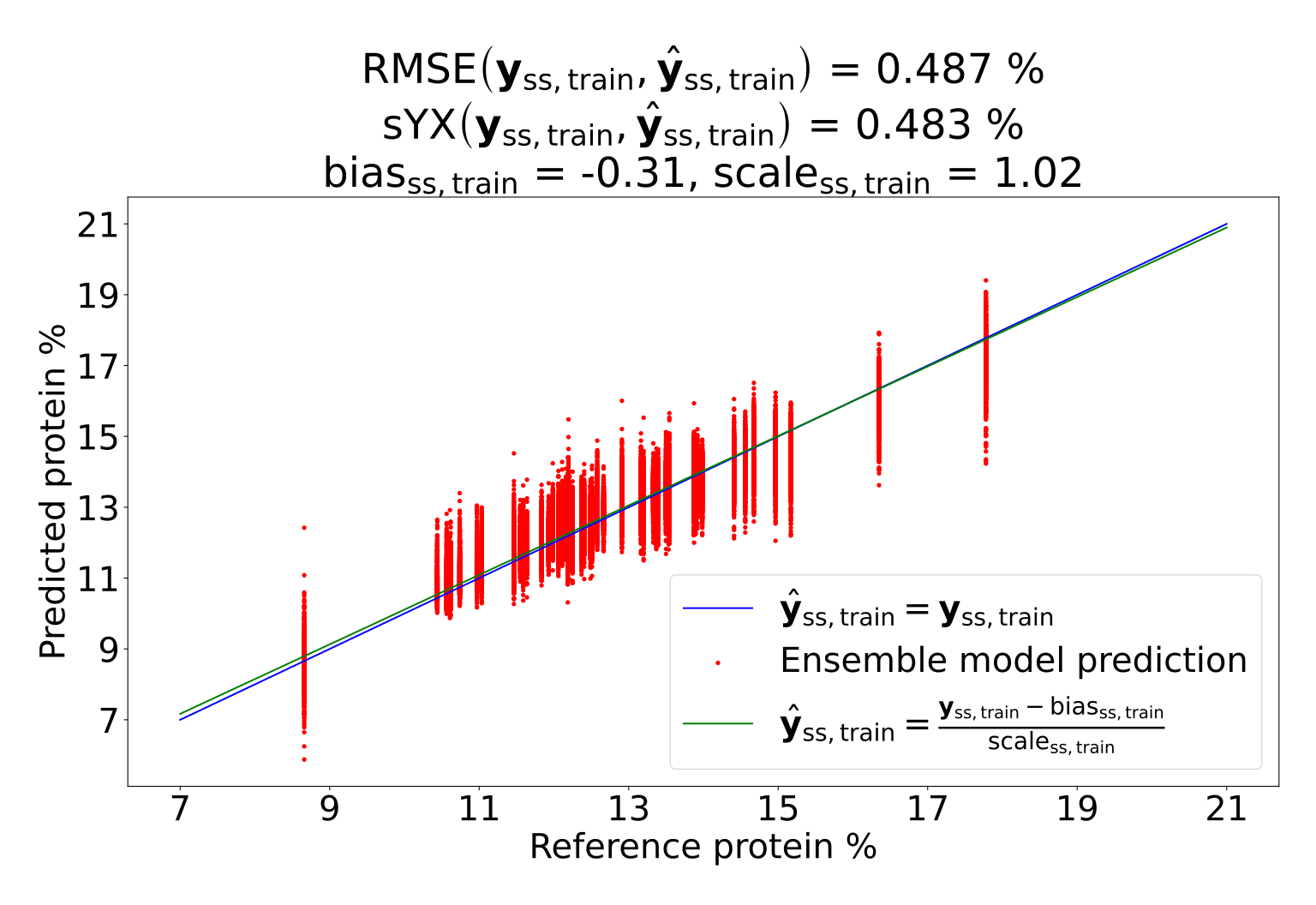}
        \caption{Modified ResNet-18 Regressor, training set.}
        \label{fig:training_individual_predictions_resnet_errorbar_plot}
    \end{subfigure}
    \begin{subfigure}[b]{0.33\textwidth}
        \centering
        \includegraphics[width=\textwidth]{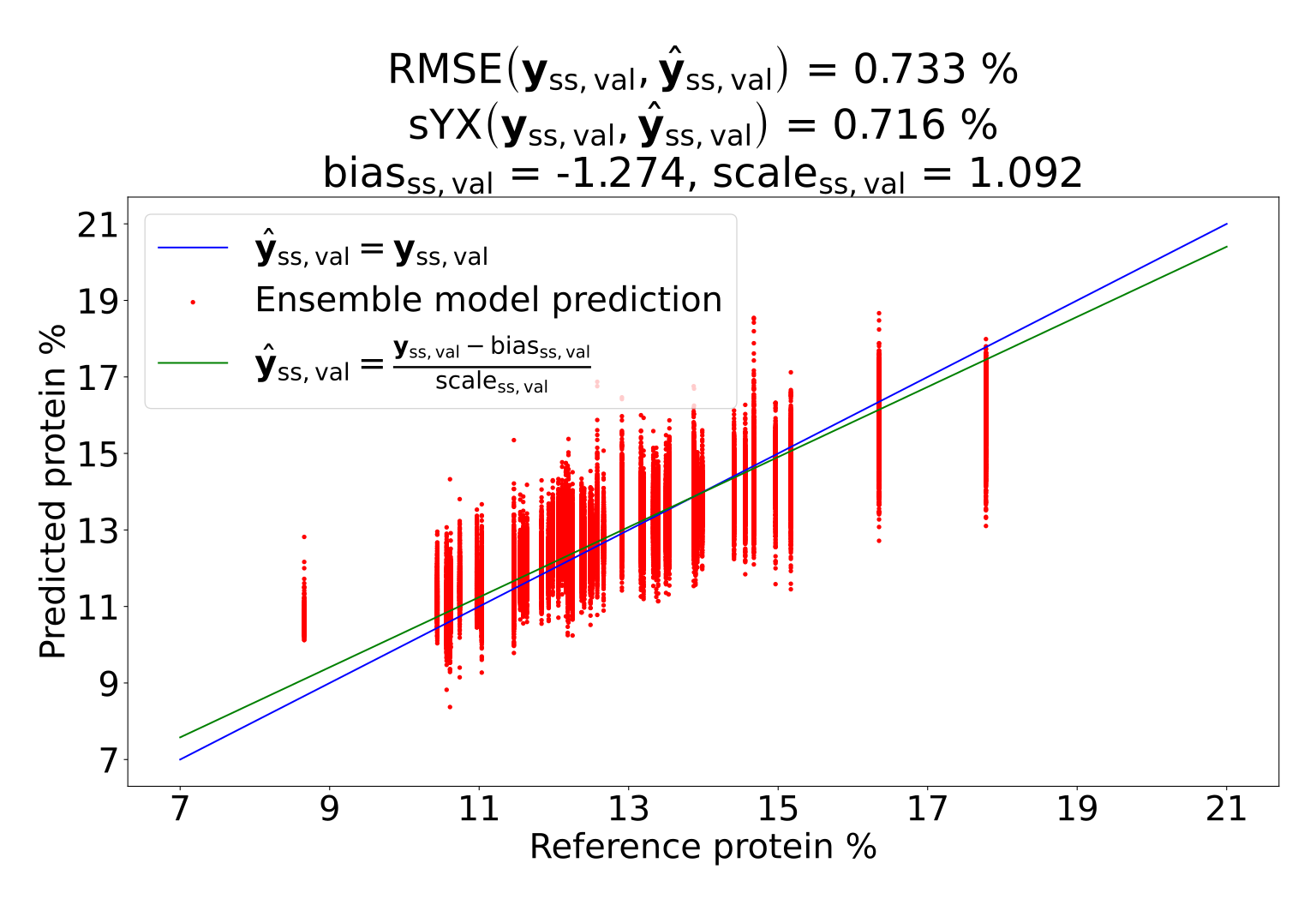}
        \caption{Modified ResNet-18 Regressor, validation set.}
    \end{subfigure}
    \begin{subfigure}[b]{0.33\textwidth}
        \centering
        \includegraphics[width=\textwidth]{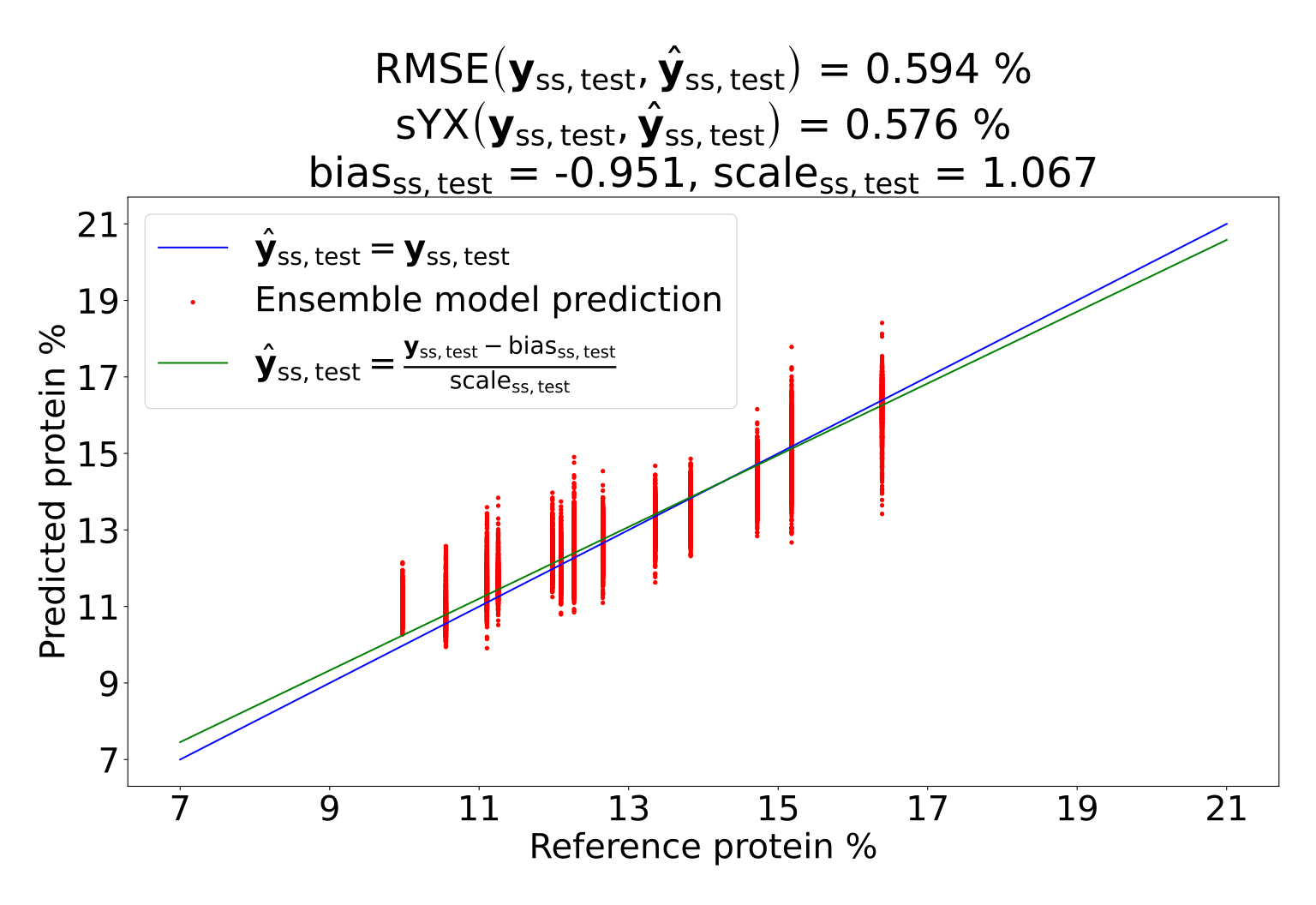}
        \caption{Modified ResNet-18 Regressor, test set.}
    \end{subfigure}
    \\
    \begin{subfigure}[b]{0.33\textwidth}
        \centering
        \includegraphics[width=\textwidth]{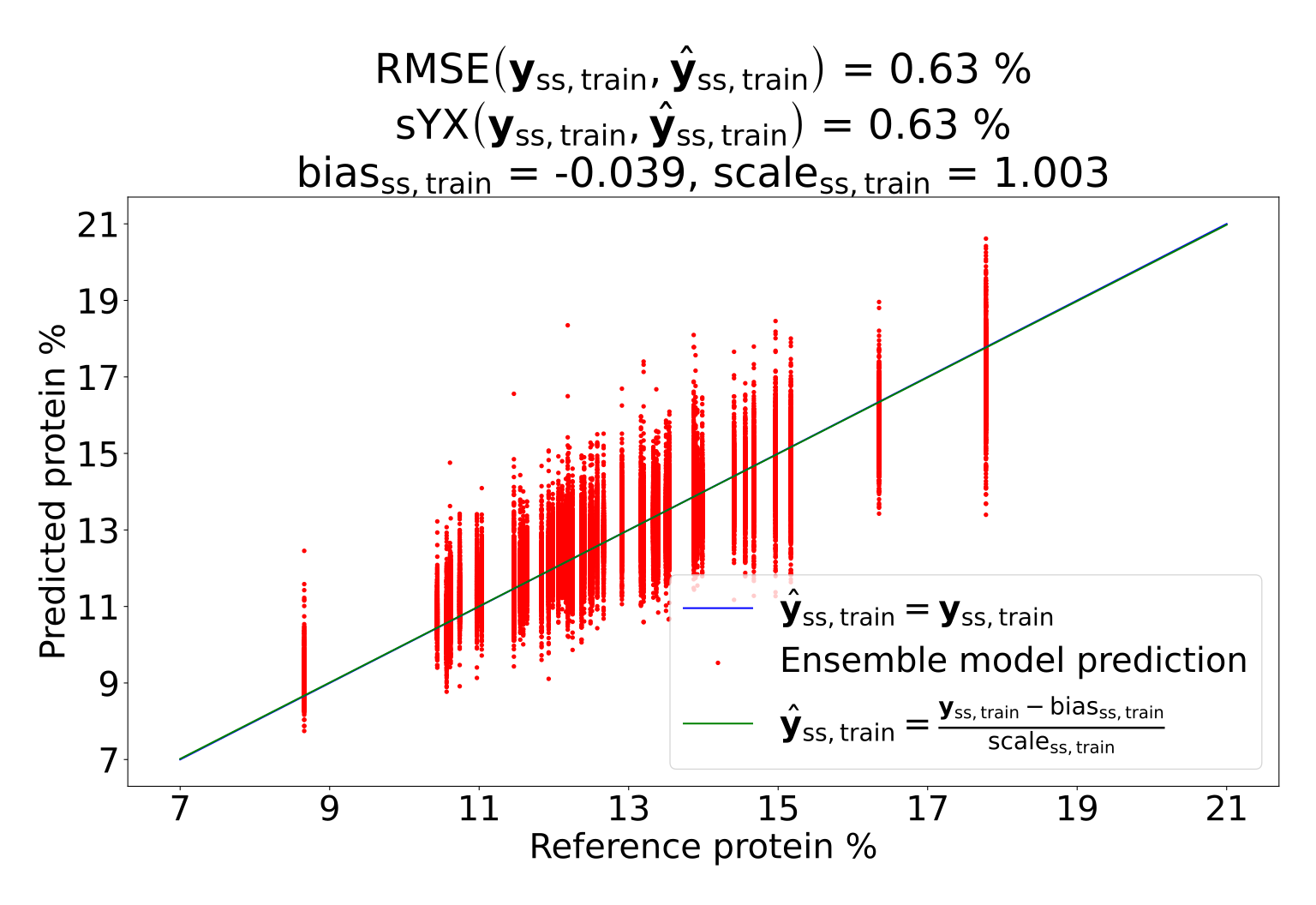}
        \caption{PLS-R, training set.}
        \label{fig:training_individual_predictions_pls_errorbar_plot}
    \end{subfigure}
    \begin{subfigure}[b]{0.33\textwidth}
        \centering
        \includegraphics[width=\textwidth]{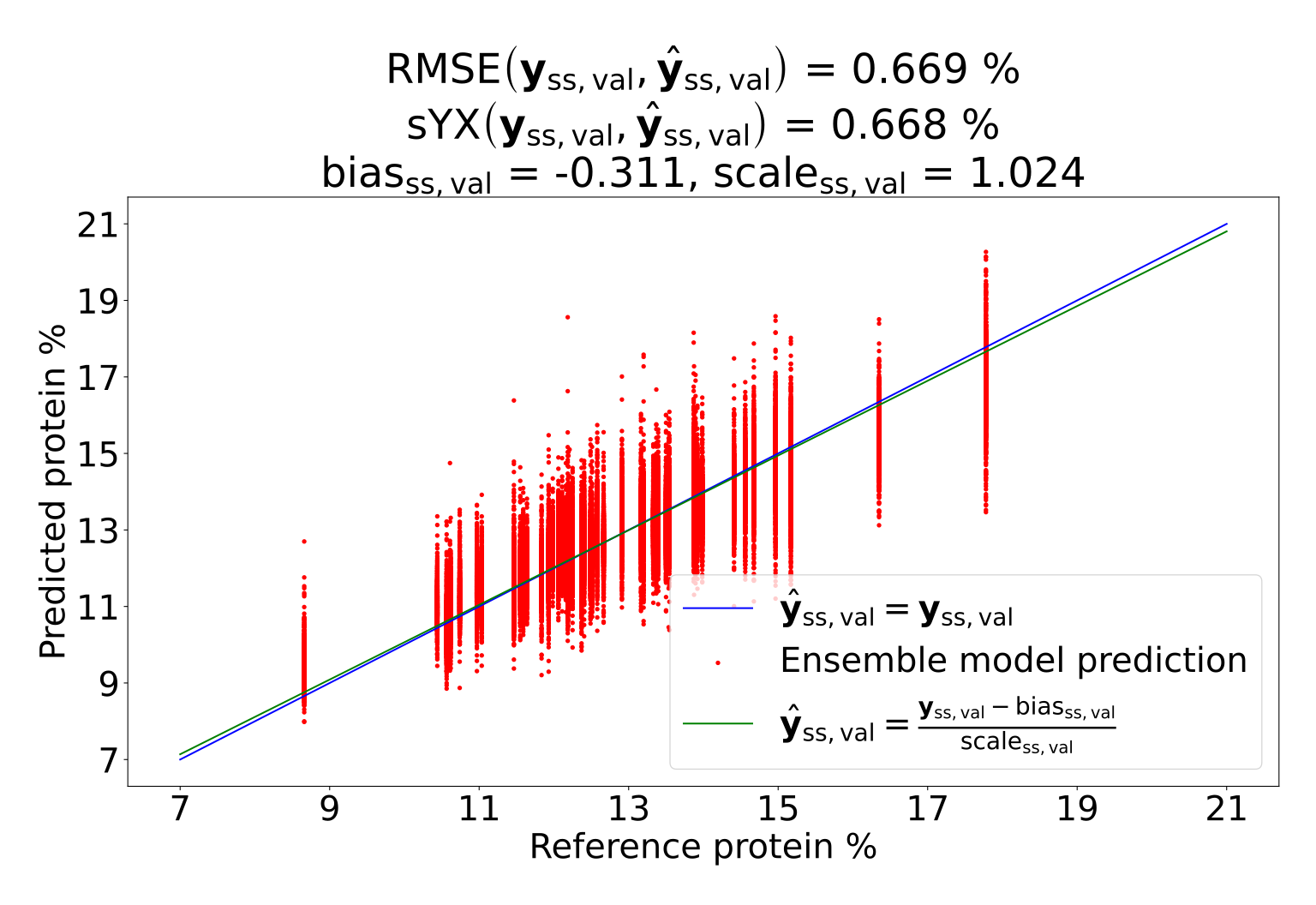}
        \caption{PLS-R, validation set.}
    \end{subfigure}
    \begin{subfigure}[b]{0.33\textwidth}
        \centering
        \includegraphics[width=\textwidth]{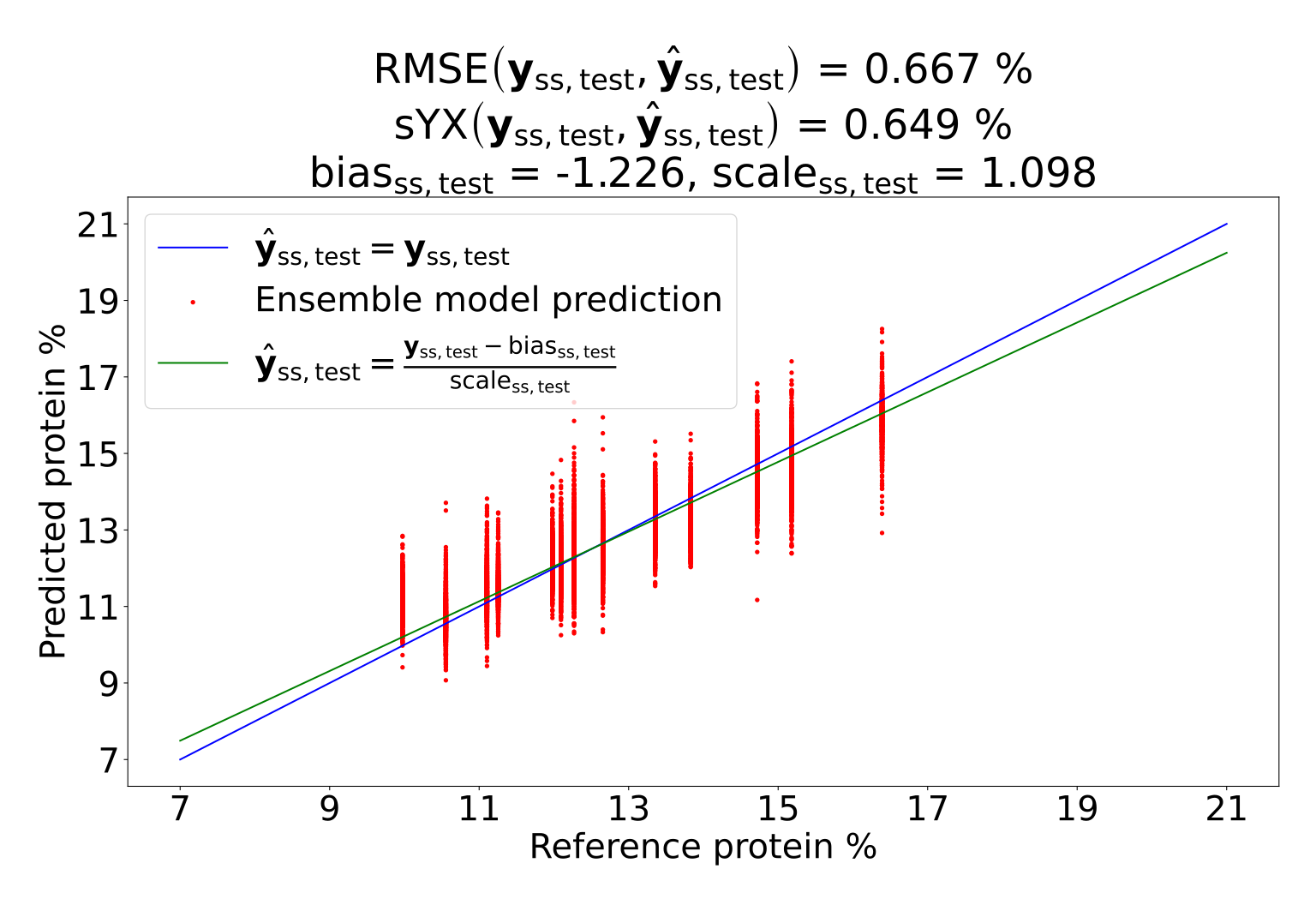}
        \caption{PLS-R, test set.}
    \end{subfigure}
    \caption{Modified ResNet-18 Regressor and PLS-R prediction plots of $\hat{\mathbf{y}}_{\text{ss}}$ and $\mathbf{y}_{\text{ss}}$.}
    \label{fig:individual_predictions_errorbar_plots}
\end{figure}

By inspecting \myfigref{fig:mean_bulk_sample_predictions_errorbar_plots}, it is evident that all plots show an sYX that is much lower than the corresponding RMSE. Likewise, all plots exhibit a bias$_\text{bm}$ of less than $0$ and a scale$_\text{bm}$ of greater than $1$. These values imply that bulk samples with a low protein reference value are systematically overpredicted and that bulk samples with a high protein reference value are systematically underpredicted. This phenomenon is persistent across both models, and all dataset splits.
Interestingly, in \myfigref{fig:individual_predictions_errorbar_plots}, where we plot $\hat{\mathbf{y}}_{\text{ss}}$ against $\mathbf{y}_{\text{ss}}$, this phenomenon is hard to notice as RMSE$(\mathbf{y}_{\text{ss}}, \hat{\mathbf{y}}_{\text{ss}})$ and sYX$(\mathbf{y}_{\text{ss}}, \hat{\mathbf{y}}_{\text{ss}})$ are very close to eachother.

\begin{figure}[htbp]
    \centering
    \begin{subfigure}[b]{0.48\textwidth}
        \centering
        \includegraphics[width=\textwidth]{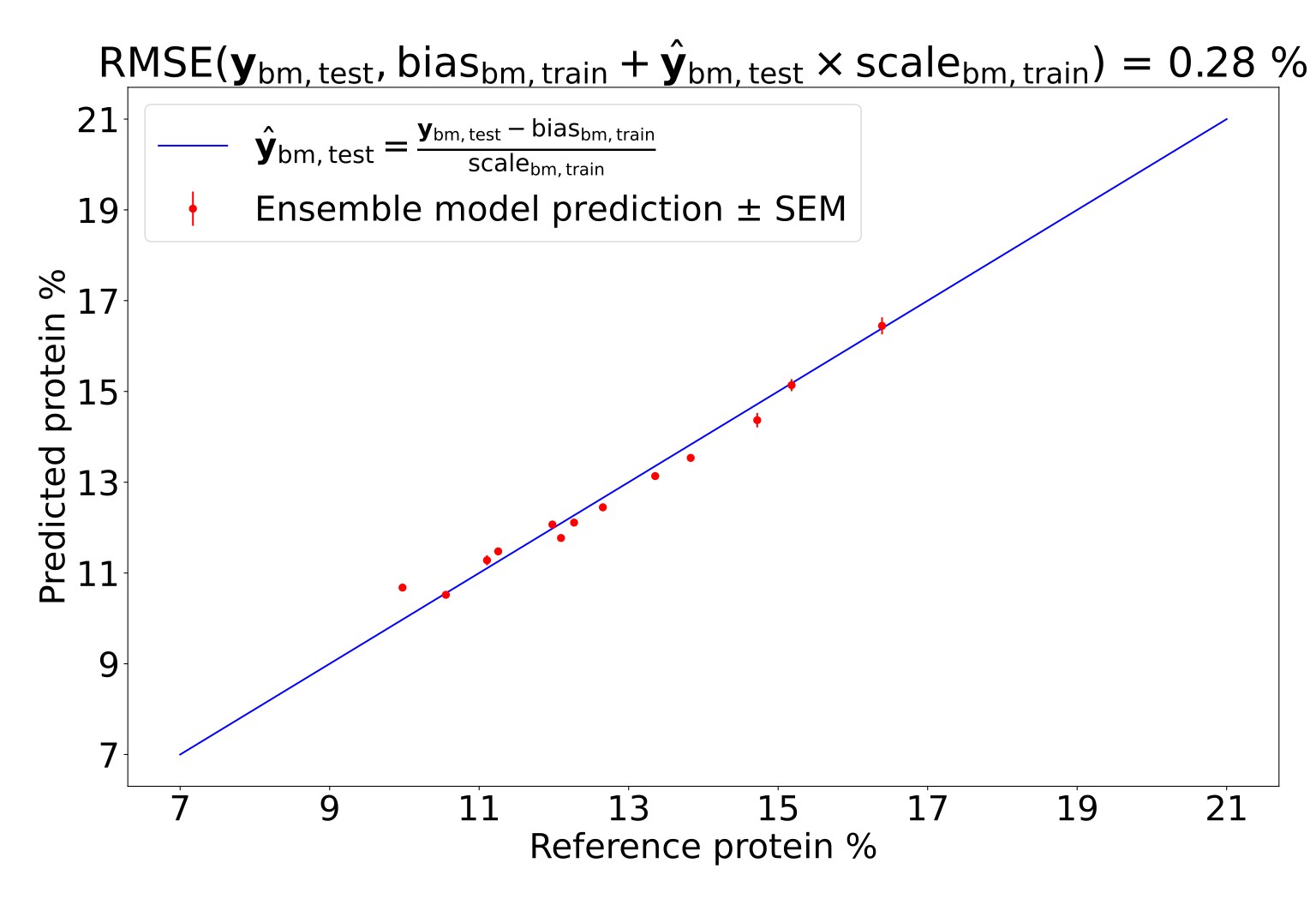}
        \caption{Modified ResNet-18 Regressor, test set using bias and scale from training set mean predictions.}
    \end{subfigure}
    \begin{subfigure}[b]{0.48\textwidth}
        \centering
        \includegraphics[width=\textwidth]{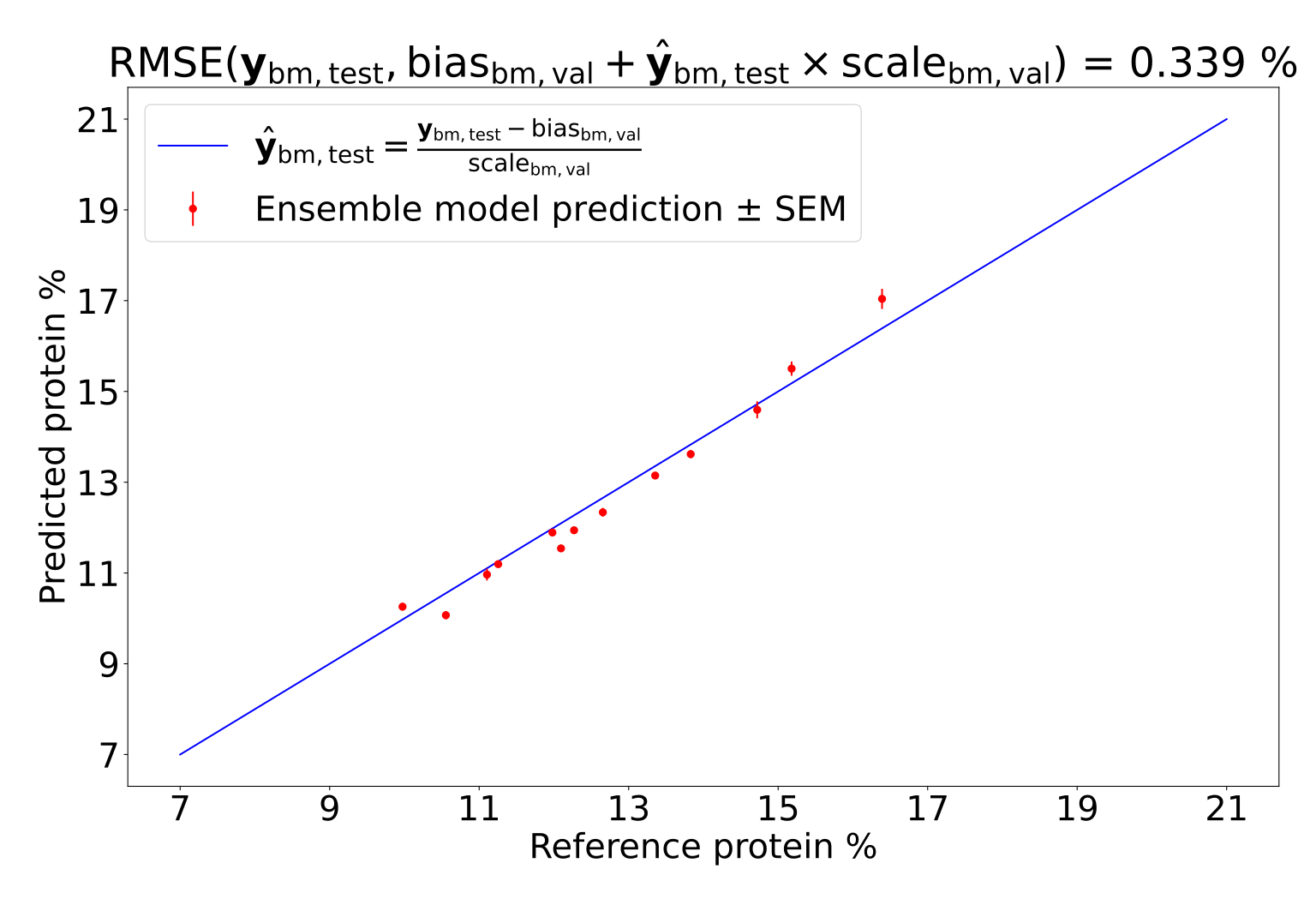}
        \caption{Modified ResNet-18 Regressor, test set using bias and scale from validation set mean predictions.}
    \end{subfigure}
    \\
    \begin{subfigure}[b]{0.48\textwidth}
        \centering
        \includegraphics[width=\textwidth]{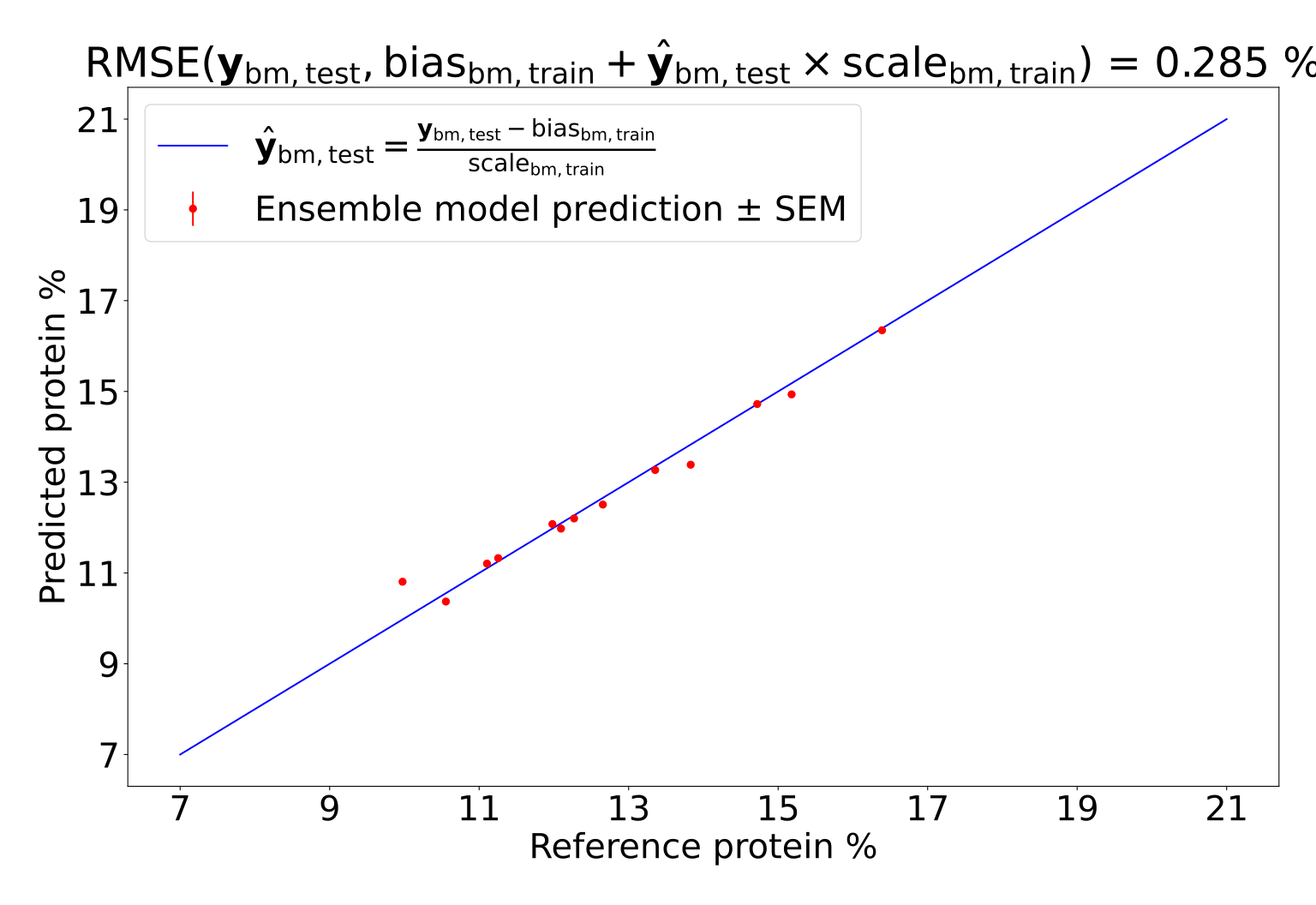}
        \caption{PLS-R, test set using bias and scale from training set mean predictions.}
    \end{subfigure}
    \begin{subfigure}[b]{0.48\textwidth}
        \centering
        \includegraphics[width=\textwidth]{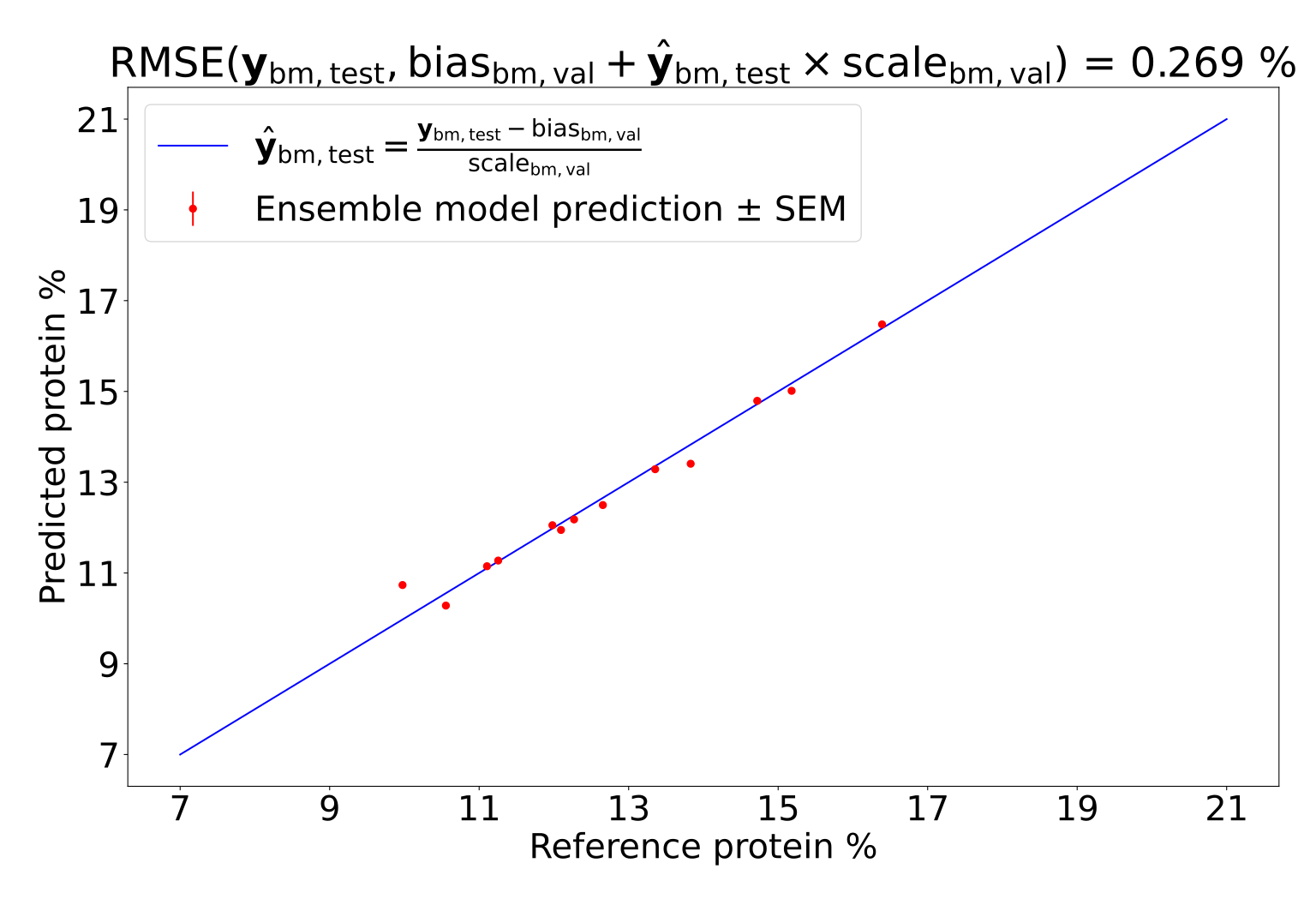}
        \caption{PLS-R, test set using bias and scale from validation set mean predictions.}
    \end{subfigure}
    \caption{Test set errorbar plots of linearly corrected predictions based on bias and scale parameters from the training and validation splits, respectively.}
    \label{fig:syx_corrected_mean_bulk_sample_predictions_errorbar_plots}
\end{figure}

These results indicate that simply taking the bulk sample-wise average over the subsample predictions, $\hat{\mathbf{y}}_{\text{ss}}$, to achieve $\hat{\mathbf{y}}_{\text{bm}}$ does not lead to a good estimate of the bulk sample reference, $\mathbf{y}_{\text{bm}}$. Instead, a linear correction must be applied to achieve the best relationship. However, while bias$_\text{test}$ and scale$_\text{test}$ tell us how to best correct our predictions, we can not use these parameters as that would make the test predictions biased. Instead, we are free to choose the bias and scale parameters from either the training or validation set. In \myfigref{fig:syx_corrected_mean_bulk_sample_predictions_errorbar_plots} we show how the test set predictions look when correcting the test set predictions using either bias$_\text{bm,training}$ and scale$_\text{bm,training}$ or bias$_\text{bm,val}$ and scale$_\text{bm,val}$. Interestingly, the best choice of parameters differs for Modified ResNet-18 Regressor and PLS-R. For Modified ResNet-18 Regressor, doing a correction based on the training set parameters is optimal, while basing the correction on the validation set parameters is optimal for PLS-R. For both models, however, using either of the two sets of parameters for a linear correction yields better results than not doing a linear correction at all as evident by comparing the RMSEs in \myfigref{fig:mean_bulk_sample_predictions_errorbar_plots} with those on the test sets in \myfigref{fig:syx_corrected_mean_bulk_sample_predictions_errorbar_plots}.

\subsection{Calibrating PLS-R directly on bulk samples}
While calibrating Modified ResNet-18 Regressor, a function with $11.2$ million parameters, requires a lot of data, PLS-R can be calibrated with much less data. Here, we want to calibrate PLS-R on the 50 mean spectra from the bulk samples in the CV dataset splits and assess its performance on the 13 mean spectra in the test dataset split. Following the precedence set with the previous models, we construct an ensemble of five PLS-R models to predict $\hat{\mathbf{y}}_{\text{bulk}}$ and minimize RMSE$(\mathbf{y}_{\text{bulk}}, \hat{\mathbf{y}}_{\text{bulk}})$. We will call this ensemble model PLS-R$_\text{bulk}$. Note that, by definition, $\mathbf{y}_{\text{bulk}}=\mathbf{y}_{\text{bm}}$. We use the same preprocessing for PLS-R$_\text{bulk}$ that we used for PLS-R.

\begin{figure}[htbp]
    \centering
    \begin{subfigure}[b]{0.33\textwidth}
        \centering
        \includegraphics[width=\textwidth]{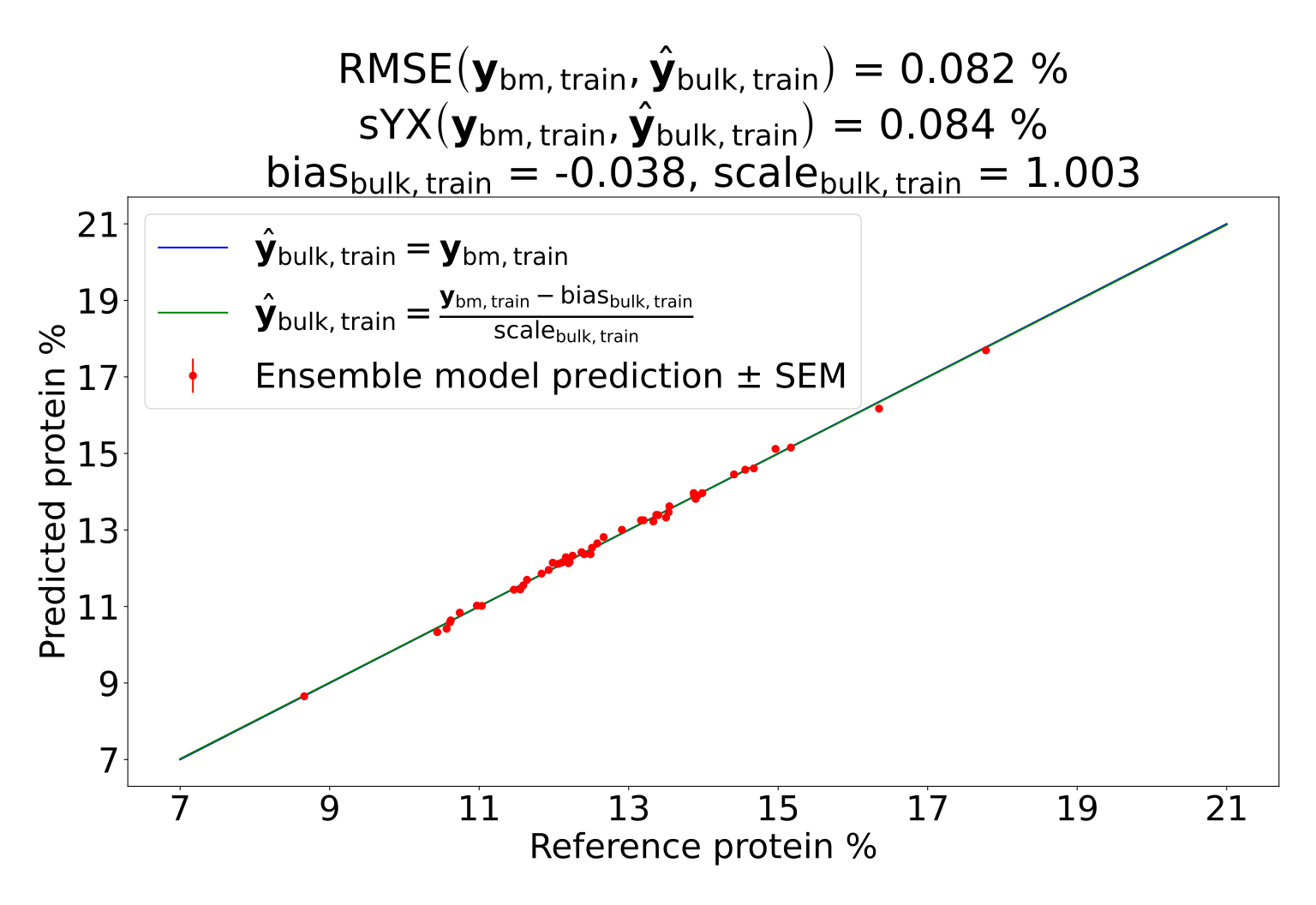}
        \caption{PLS-R$_\text{bulk}$ predicting on bulk sample mean spectra from the training set.}
    \end{subfigure}
    \begin{subfigure}[b]{0.33\textwidth}
        \centering
        \includegraphics[width=\textwidth]{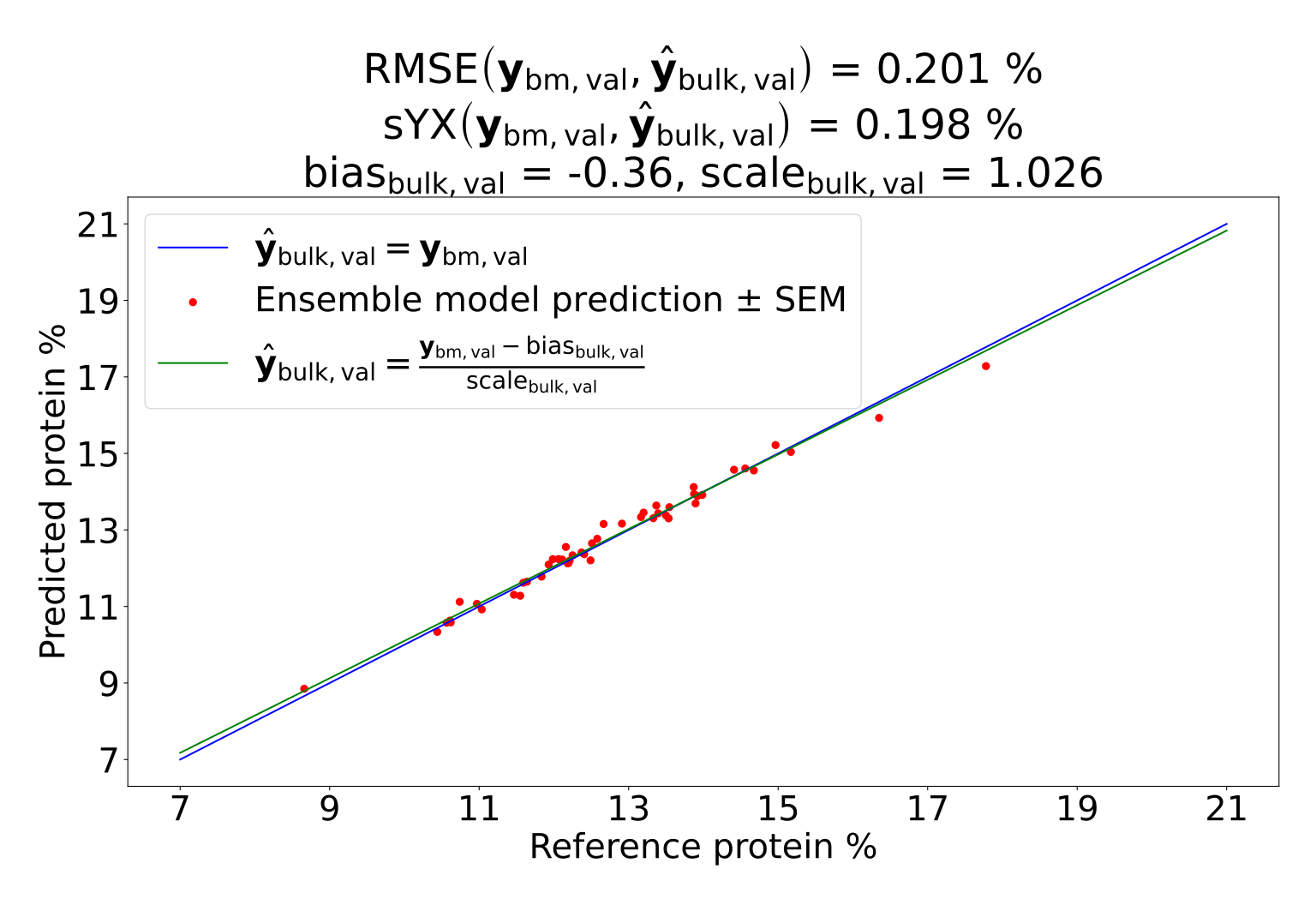}
        \caption{PLS-R$_\text{bulk}$, predicting on bulk sample mean spectra from the validation set.}
    \end{subfigure}
    \begin{subfigure}[b]{0.33\textwidth}
        \centering
        \includegraphics[width=\textwidth]{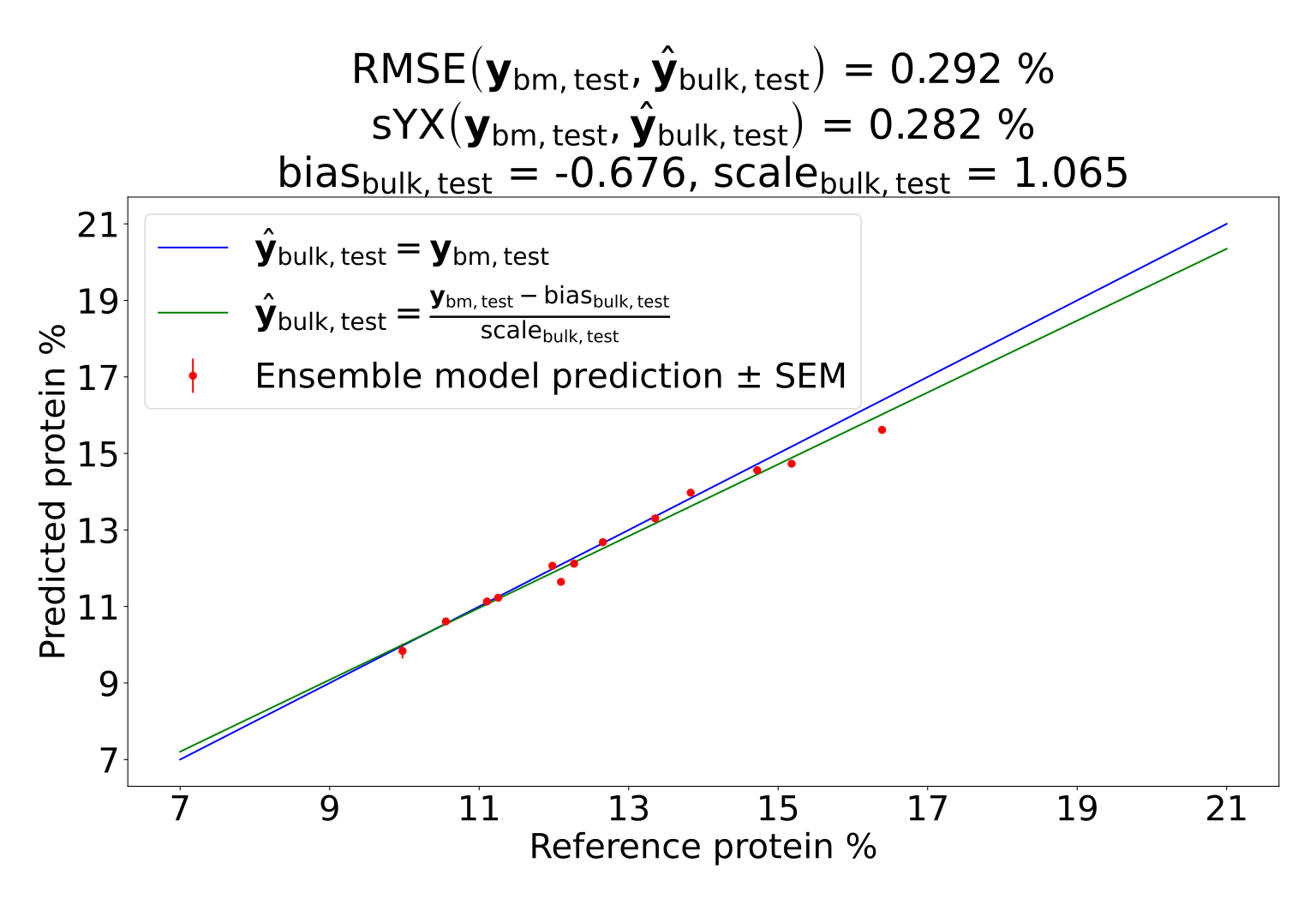}
        \caption{PLS-R$_\text{bulk}$, predicting on bulk sample mean spectra from the test set.}
    \end{subfigure}
    \\
    \begin{subfigure}[b]{0.33\textwidth}
        \centering
        \includegraphics[width=\textwidth]{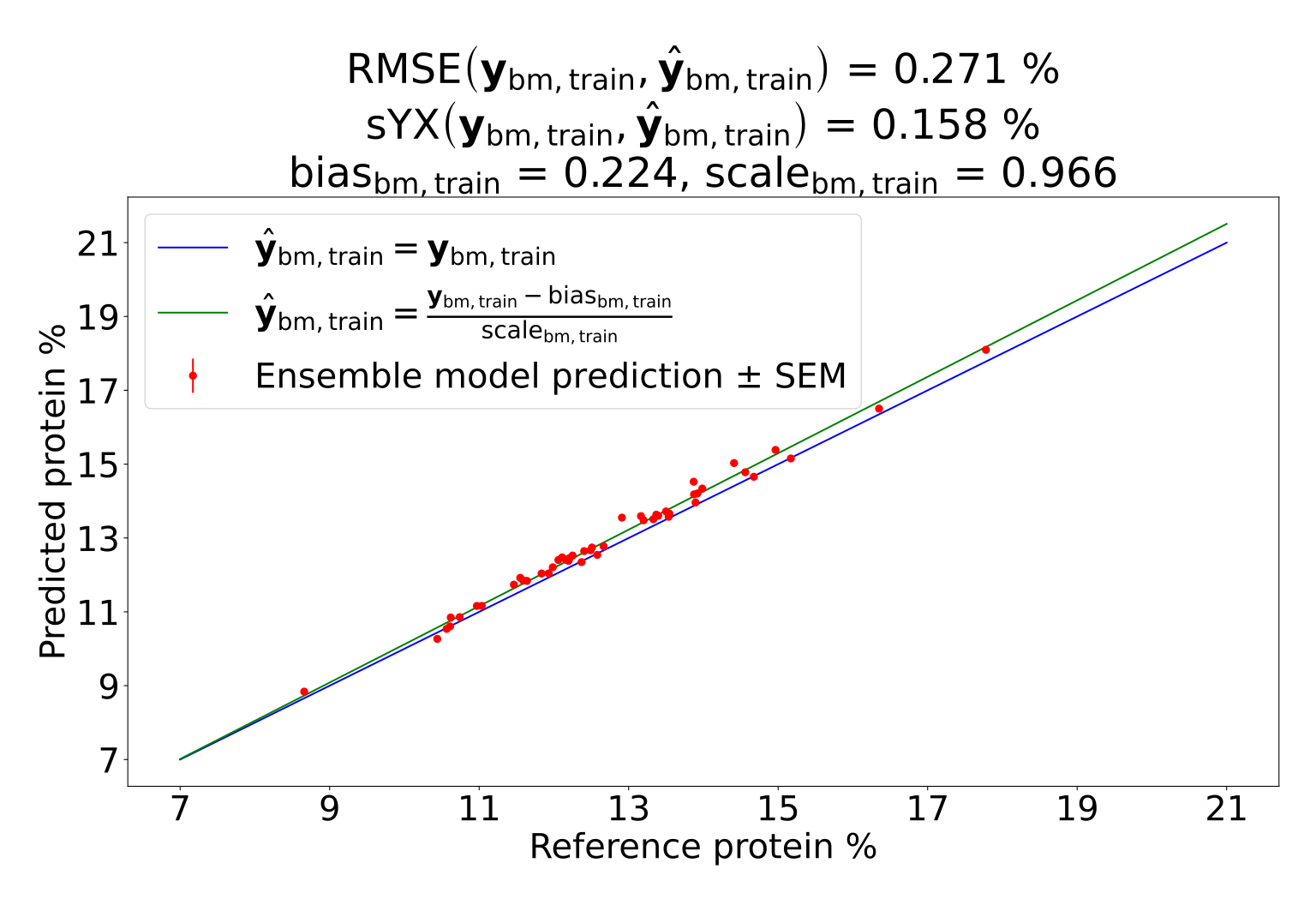}
        \caption{PLS-R$_\text{bulk}$ predicting on bulk subsample mean spectra from the training set.}
    \end{subfigure}
    \begin{subfigure}[b]{0.33\textwidth}
        \centering
        \includegraphics[width=\textwidth]{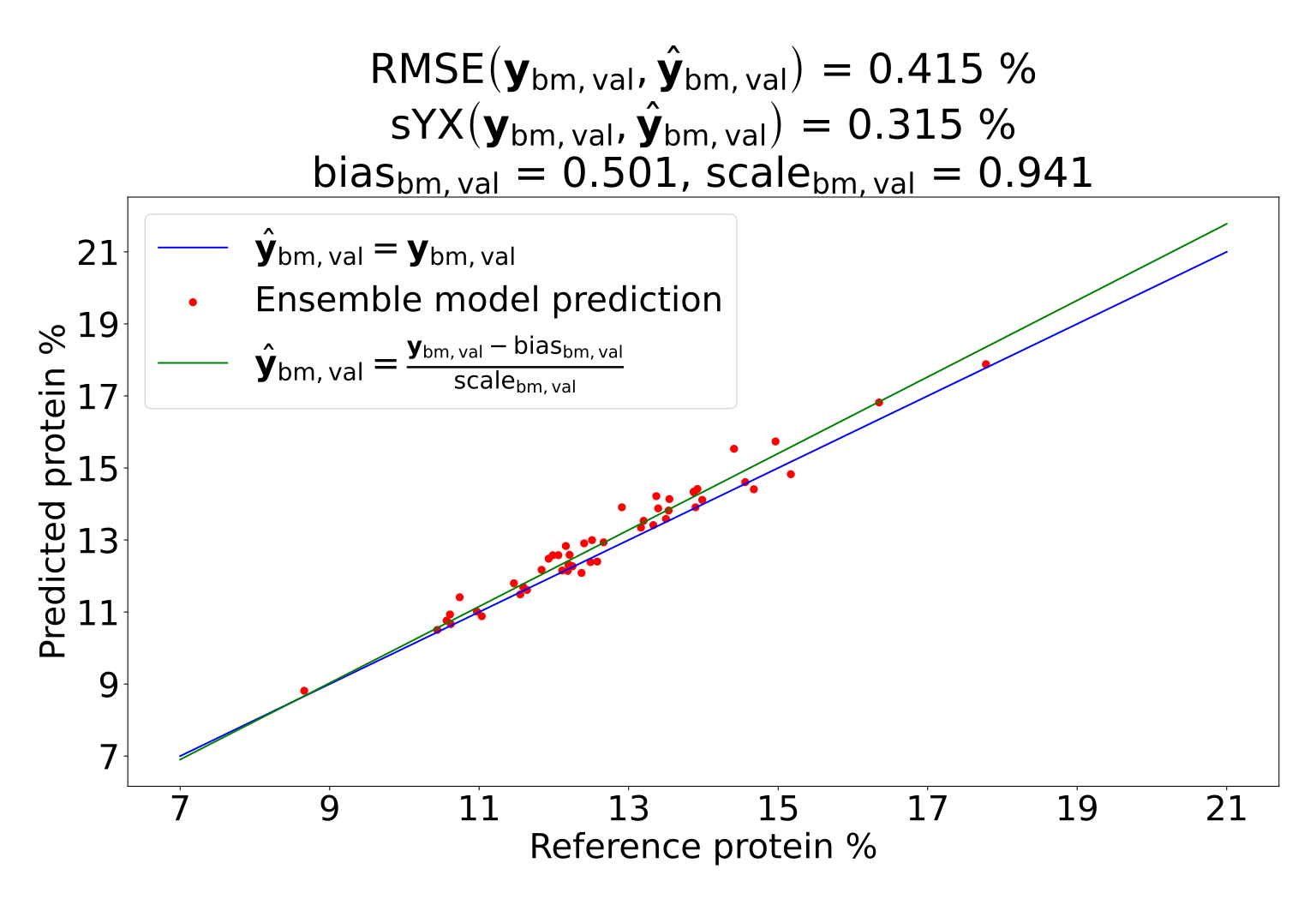}
        \caption{PLS-R$_\text{bulk}$ predicting on bulk subsample mean spectra from the training set.}
    \end{subfigure}
    \begin{subfigure}[b]{0.33\textwidth}
        \centering
        \includegraphics[width=\textwidth]{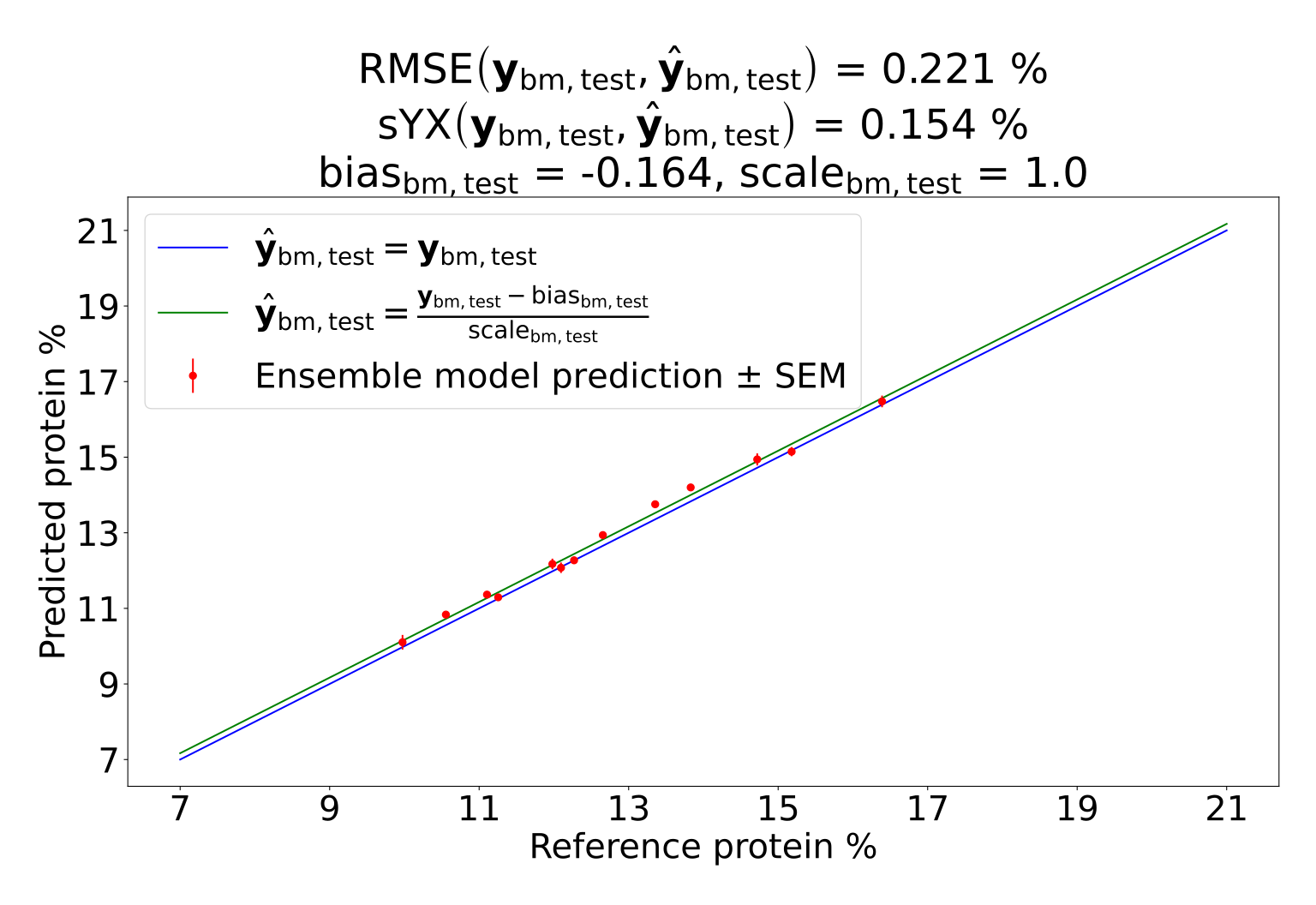}
        \caption{PLS-R$_\text{bulk}$ predicting on bulk subsample mean spectra from the validation set.}
    \end{subfigure}
    \caption{Errorbar plots for PLS-R$_\text{bulk}$ mean predictions. While the test set RMSE of PLS-R$_\text{bulk}$ given by RMSE$(\mathbf{y}_{\text{bm,test}}, \hat{\mathbf{y}}_{\text{bm,test}})$ is lower than all other test set RMSEs, we hypothesize that this is insignificant, as the corresponding training and validation RMSEs, shown in the bottom row of this figure, are not equally better than their counterparts from the other models as evident by comparing them to those shown in \myfigref{fig:mean_bulk_sample_predictions_errorbar_plots}.}
    \label{fig:bulk_plsr_mean_predictions}
\end{figure}

\begin{figure}[htbp]
    \centering
    \begin{subfigure}[b]{0.33\textwidth}
        \centering
        \includegraphics[width=\textwidth]{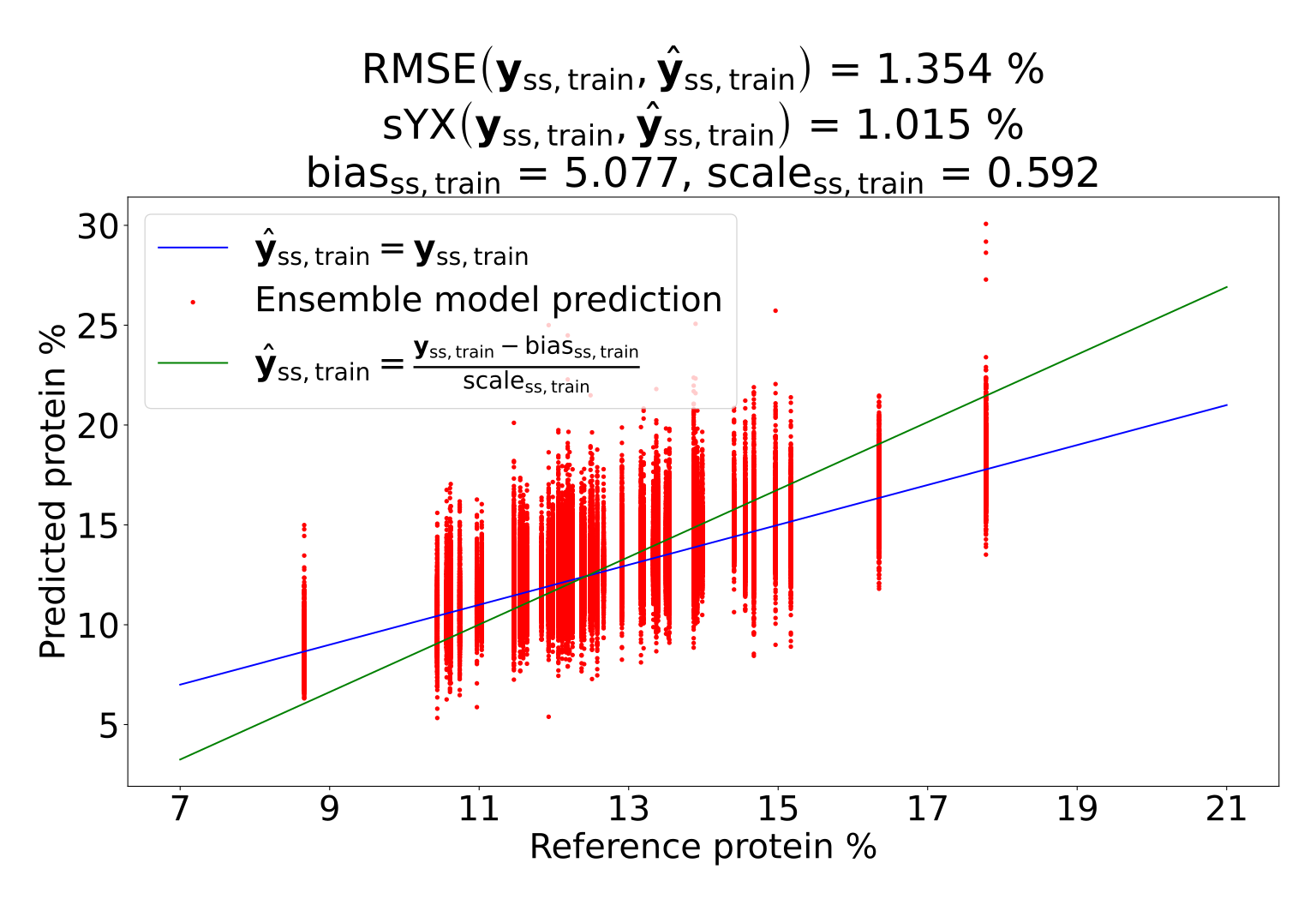}
        \caption{PLS-R$_\text{bulk}$ predicting on bulk subsamples from the training set.}
    \end{subfigure}
    \begin{subfigure}[b]{0.33\textwidth}
        \centering
        \includegraphics[width=\textwidth]{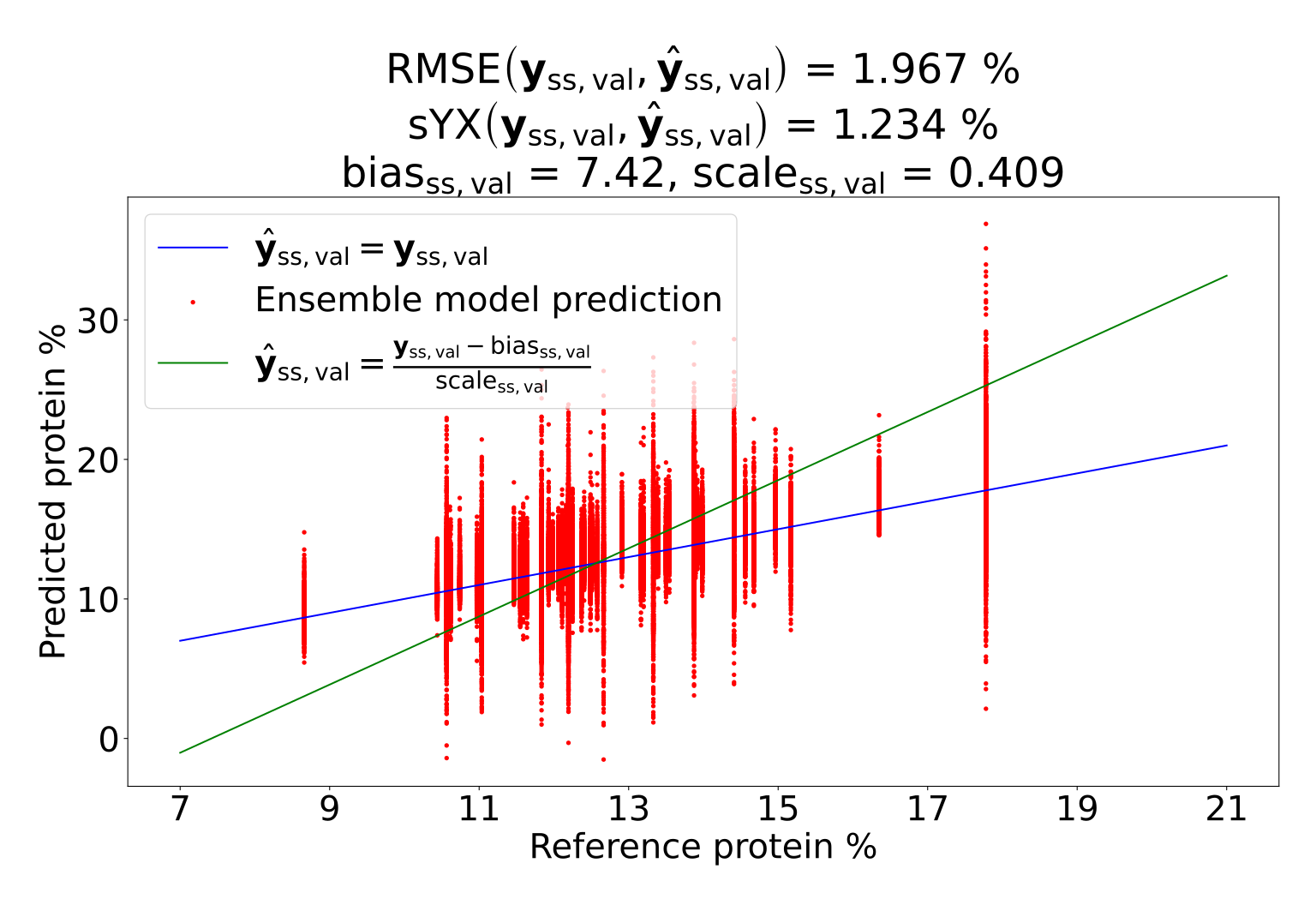}
        \caption{PLS-R$_\text{bulk}$, predicting on bulk subsamples from the validation set.}
    \end{subfigure}
    \begin{subfigure}[b]{0.33\textwidth}
        \centering
        \includegraphics[width=\textwidth]{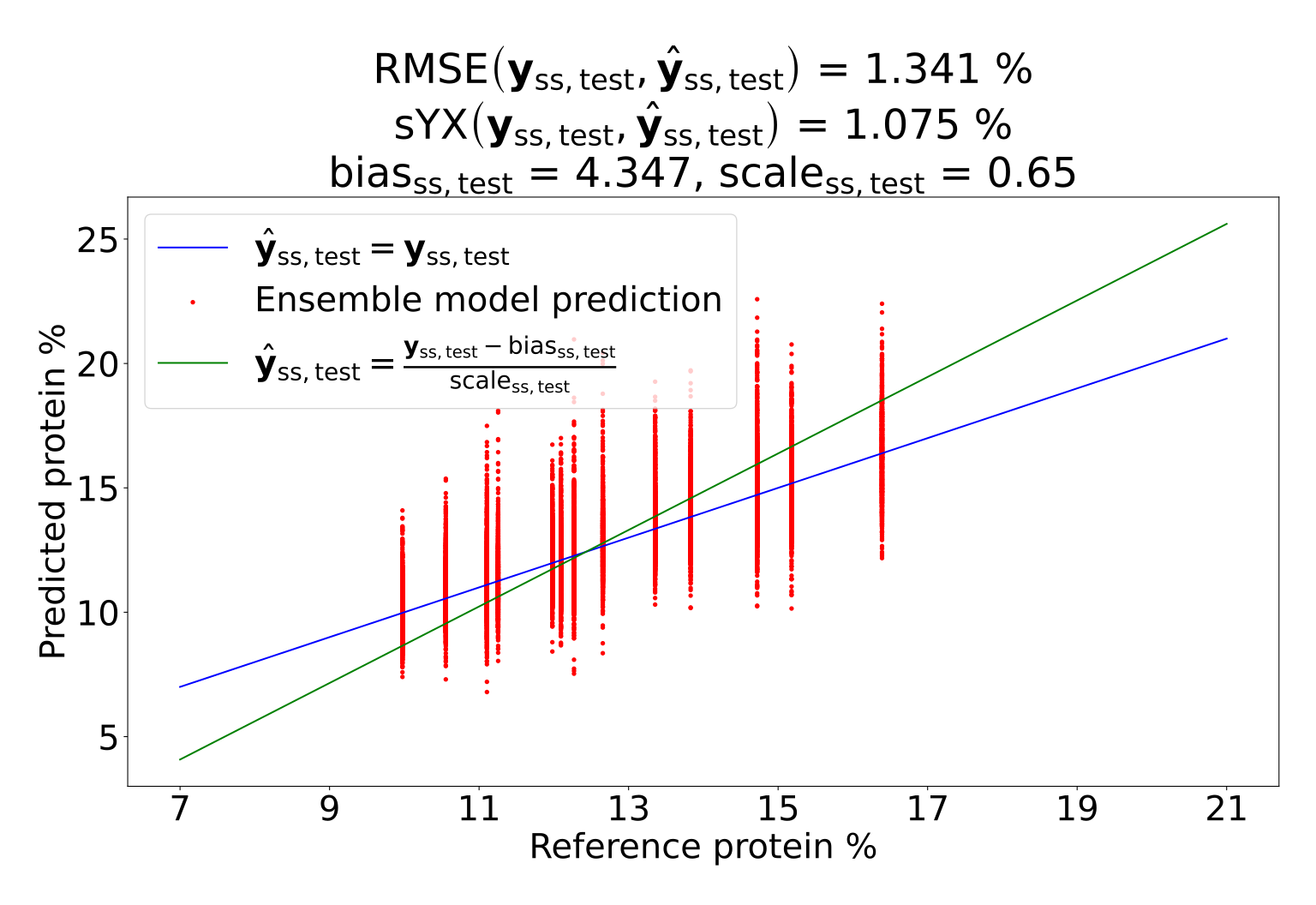}
        \caption{PLS-R$_\text{bulk}$, predicting on bulk subsamples from the test set.}
    \end{subfigure}
    \caption{Errorbar plots for PLS-R$_\text{bulk}$ bulk subsample predictions.}
    \label{fig:bulk_plsr_subsample_predictions}
\end{figure}

In \myfigref{fig:bulk_plsr_mean_predictions}, we show errorbar plots for PLS-R$_\text{bulk}$ on all three dataset splits. We show the errorbar plots both for $\hat{\mathbf{y}}_{\text{bulk}}$ and for $\hat{\mathbf{y}}_{\text{bm}}$. As can be seen on the top row in \myfigref{fig:bulk_plsr_mean_predictions}, the RMSE$(\mathbf{y}_{\text{bulk}}, \hat{\mathbf{y}}_{\text{bulk}})$ and sYX$(\mathbf{y}_{\text{bulk}}, \hat{\mathbf{y}}_{\text{bulk}})$ are very similar, indicating that the model has solved the task for which it was calibrated. The bottom row also reveals values for $\text{bias}_{\text{bm}}$ and $\text{scale}_{\text{bm}}$ much closer to $0$ and $1$, respectively, than those in the bottom row of \myfigref{fig:mean_bulk_sample_predictions_errorbar_plots}, indicating that calibrating a model to minimize RMSE$(\mathbf{y}_{\text{bulk}}, \hat{\mathbf{y}}_{\text{bulk}})$ will enable it to minimize RMSE$(\mathbf{y}_{\text{bm}}, \hat{\mathbf{y}}_{\text{bm}})$ much better than a model trained to minimize RMSE$(\mathbf{y}_{\text{ss}}, \hat{\mathbf{y}}_{\text{ss}})$. When we compare RMSE$(\mathbf{y}_{\text{bulk}}, \hat{\mathbf{y}}_{\text{bulk}})$ with those linearly corrected test set predictions of Modified ResNet-18 Regressor and PLS-R in \myfigref{fig:syx_corrected_mean_bulk_sample_predictions_errorbar_plots}, they are approximately identical, indicating that either calibration approach is equally viable for predicting the mean protein content. For the sake of completeness, we show errorbar plots and corresponding RMSE and sYX of the bulk subsample predictions using PLS-R$_\text{bulk}$ in \myfigref{fig:bulk_plsr_subsample_predictions}. The results are much worse than those of Modified ResNet-18 Regressor and PLS-R in \myfigref{fig:individual_predictions_errorbar_plots}. The corresponding bias and scale parameters are also quite extreme, indicating that calibrating a model to minimize RMSE$(\mathbf{y}_{\text{bulk}}, \hat{\mathbf{y}}_{\text{bulk}})$ will not simultaneously let it minimize RMSE$(\mathbf{y}_{\text{ss}}, \hat{\mathbf{y}}_{\text{ss}})$.

\section{Statistical analysis of prediction distributions}\label{sec:distributions}
The phenomena causing the discrepancy between minimizing RMSE$(\mathbf{y}_{\text{ss}}, \hat{\mathbf{y}}_{\text{ss}})$ or RMSE$(\mathbf{y}_{\text{bm}}, \hat{\mathbf{y}}_{\text{bm}})$ and RMSE$(\mathbf{y}_{\text{bulk}}, \hat{\mathbf{y}}_{\text{bulk}})$ can be found by inspecting the distributions of $\hat{\mathbf{y}}_{\text{ss}}$, grouped by bulk sample. If the distribution of $\hat{\mathbf{y}}_{\text{ss}}$ is symmetric, then OLS-R$\left(\hat{\mathbf{Y}}_{\text{bm}}, \mathbf{y}_{\text{bm}}\right)$ $=$ OLS-R$\left(\hat{\mathbf{Y}}_{\text{ss}}, \mathbf{y}_{\text{ss}}\right)$. While this is not the case, we wish to analyze whether the discrepancy is statistically significant or if it can be attributed to randomness. Two common symmetric distributions are the uniform distribution and the normal distribution. It is clear from inspecting the plots that the distribution of $\hat{\mathbf{y}}_{\text{ss}}$ is not uniform for any of the models on any of the dataset splits. Instead, we focus our attention on analyzing if, with high confidence, $\hat{\mathbf{y}}_{\text{ss}}$ is not normally distributed.

While inspections of the skewnesses and kurtoses are suitable for visual interpretation, they can hardly prove that each bulk sample's prediction distribution is not normal. Instead, statistical tests based on skewness and excess kurtosis are suggested by The American Statistical Association to detect deviations from normality \cite{d1990suggestion}. The skewness test \cite{d1970transformation} and kurtosis test \cite{anscombe1983distribution} are both constructed as two-sided $Z$-tests where the $Z$-scores will be approximately normally distributed under the null hypothesis that the underlying prediction distribution is also normally distributed. The related $p$-value tells us the probability of observing a $Z$-score at least as extreme as the one calculated if the underlying prediction distribution is normally distributed. Additionally, D'Agostino's $K^2$-test, an omnibus test combining the tests for skewness and kurtosis \cite{d1971omnibus, d1973tests}, summarizes the previously mentioned $Z$-tests to compute a $K^2$-score that is the sum of the squared $Z$-scores and which will be $\chi^2$-distributed with two degrees of freedom if the two $Z$-scores are normally distributed. 

Neither of the previously mentioned tests provides evidence for normality - i.e., they can not confirm the null hypothesis. Instead, they can provide evidence for deviation from normality. However, if the evidence for deviation from normality is strong enough, it is safe to assume that the predictions are not normally distributed. All statistical moments and tests are computed using software from scikit-learn \cite{scikit-learn}.

\begin{figure}[htbp]
    \centering
    \begin{subfigure}[b]{0.33\textwidth}
        \centering
        \includegraphics[width=\textwidth]{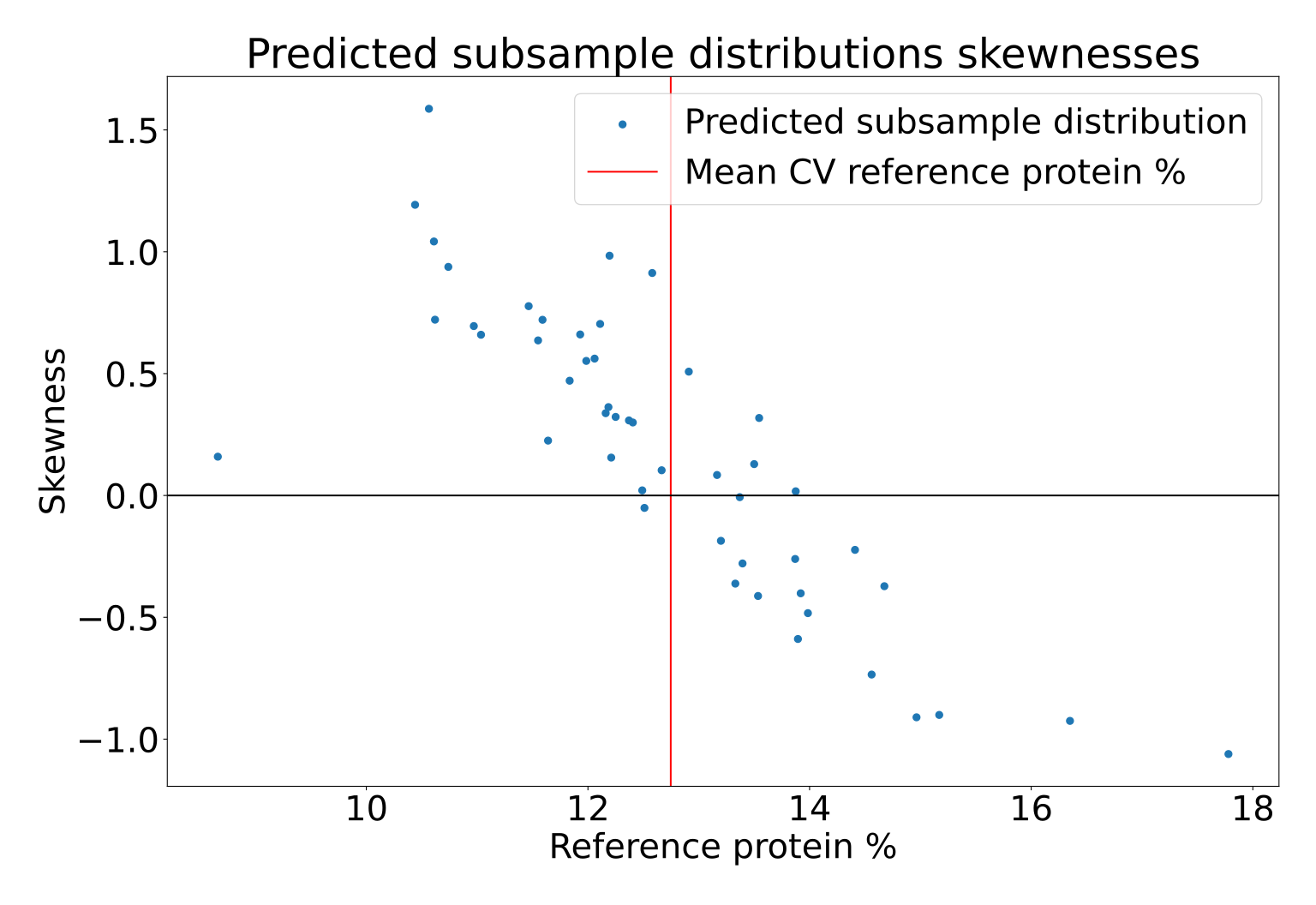}
        \caption{Training skewnesses.}
    \end{subfigure}
    \hfill
    \begin{subfigure}[b]{0.33\textwidth}
        \centering
        \includegraphics[width=\textwidth]{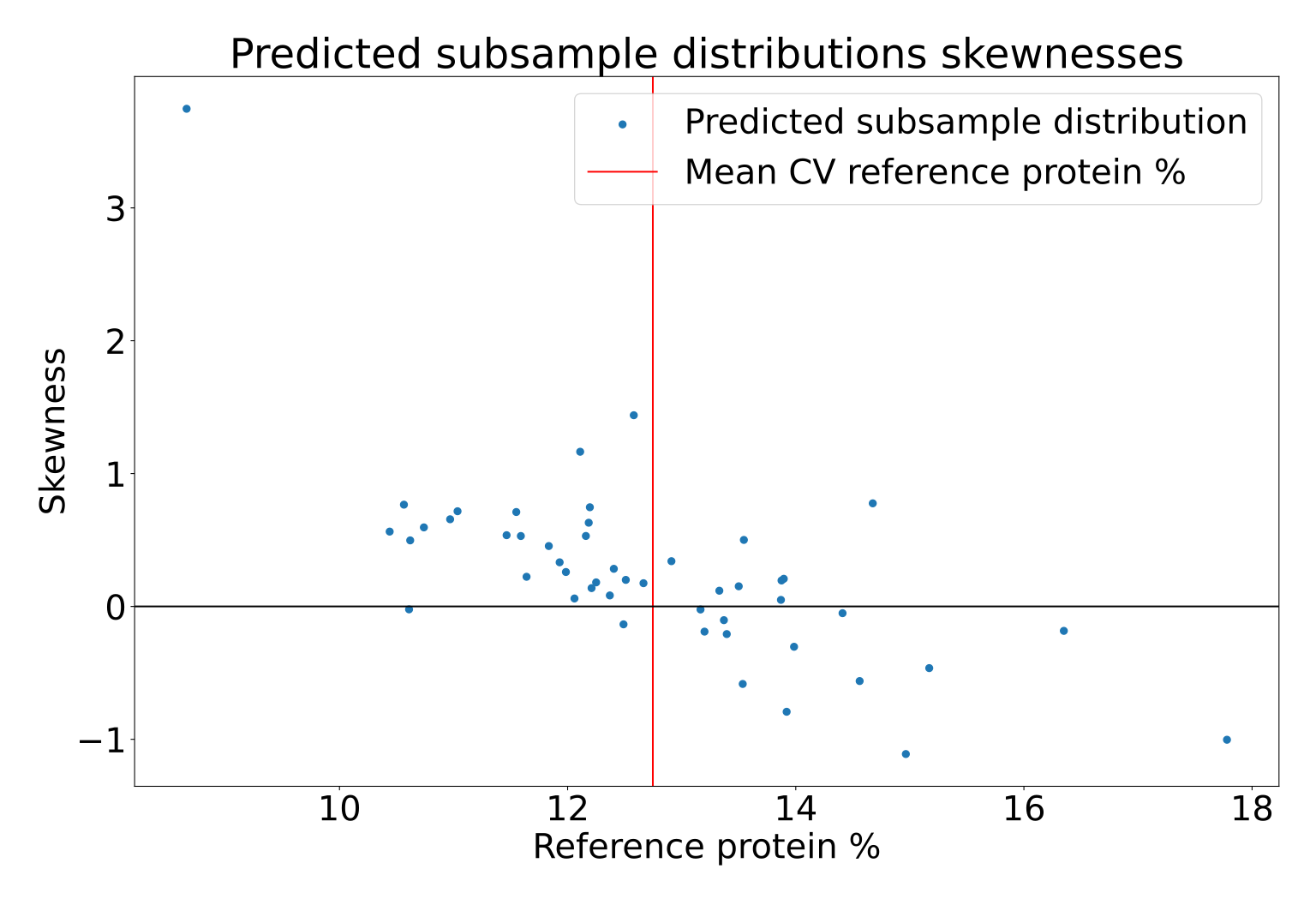}
        \caption{Validation skewnesses.}
    \end{subfigure}
    \begin{subfigure}[b]{0.33\textwidth}
        \centering
        \includegraphics[width=\textwidth]{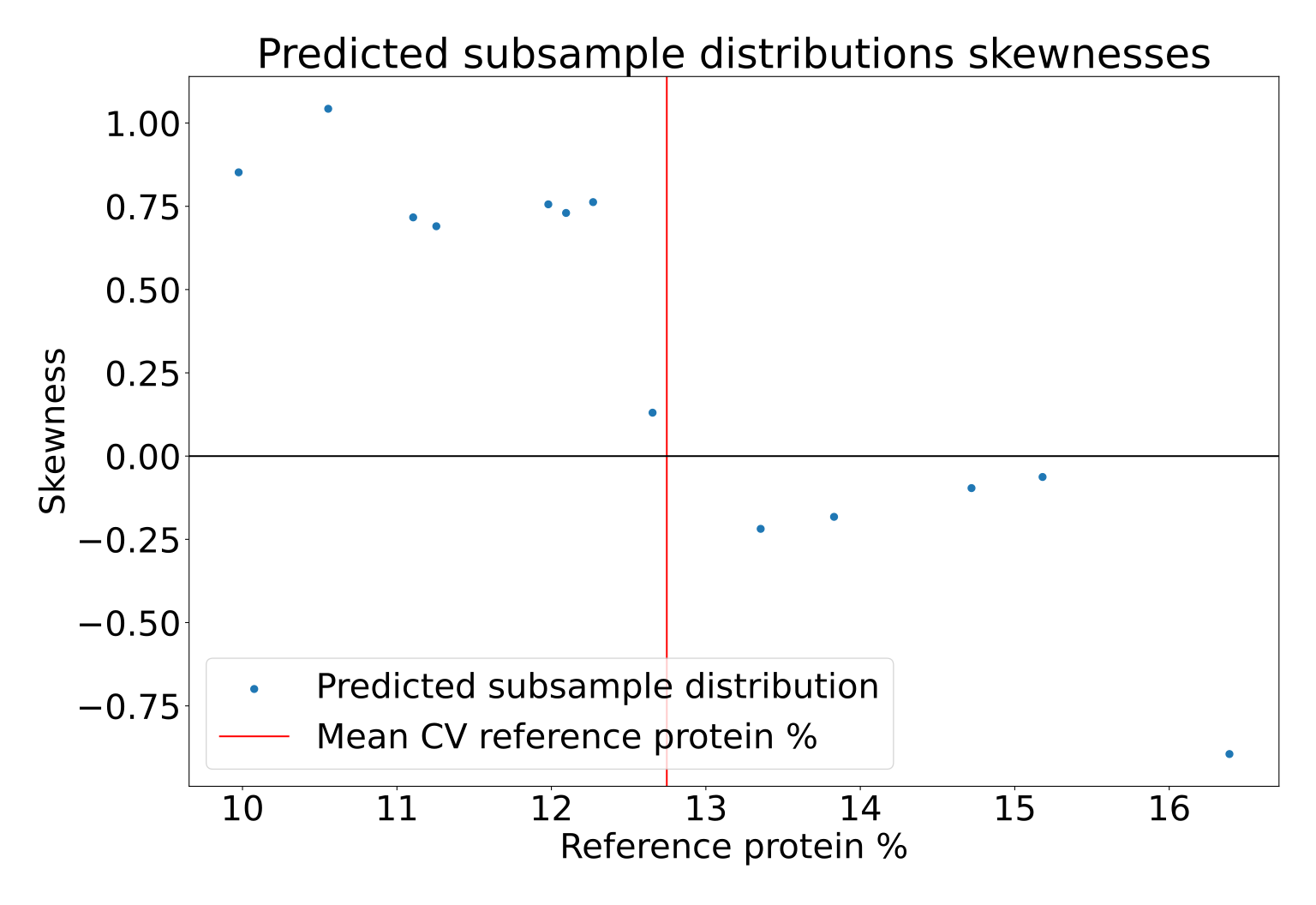}
        \caption{Test skewnesses.}
    \end{subfigure}
    \\
    \centering
    \begin{subfigure}[b]{0.33\textwidth}
        \centering
        \includegraphics[width=\textwidth]{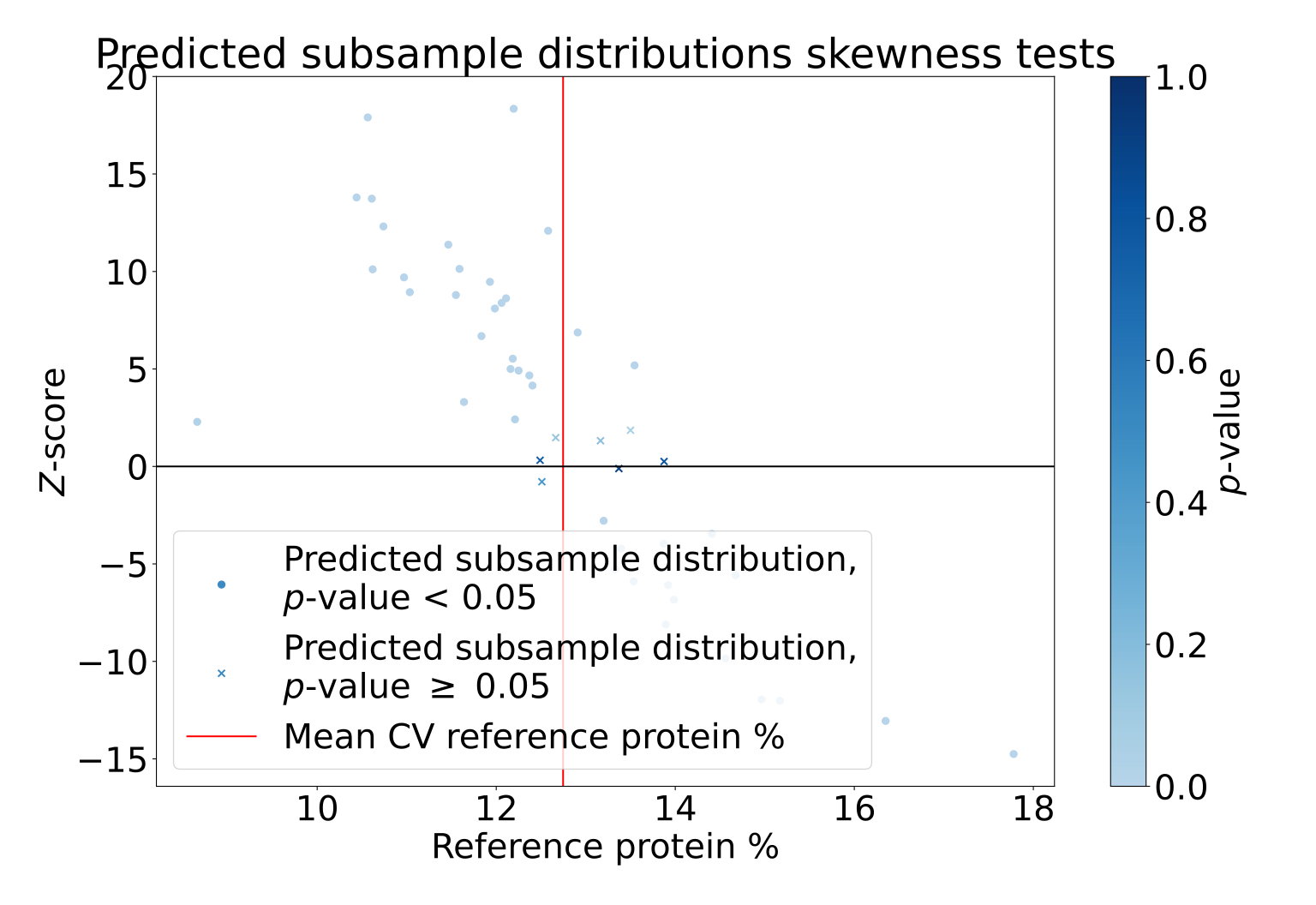}
        \caption{Training skewness $Z$-tests.}
    \end{subfigure}
    \hfill
    \begin{subfigure}[b]{0.33\textwidth}
        \centering
        \includegraphics[width=\textwidth]{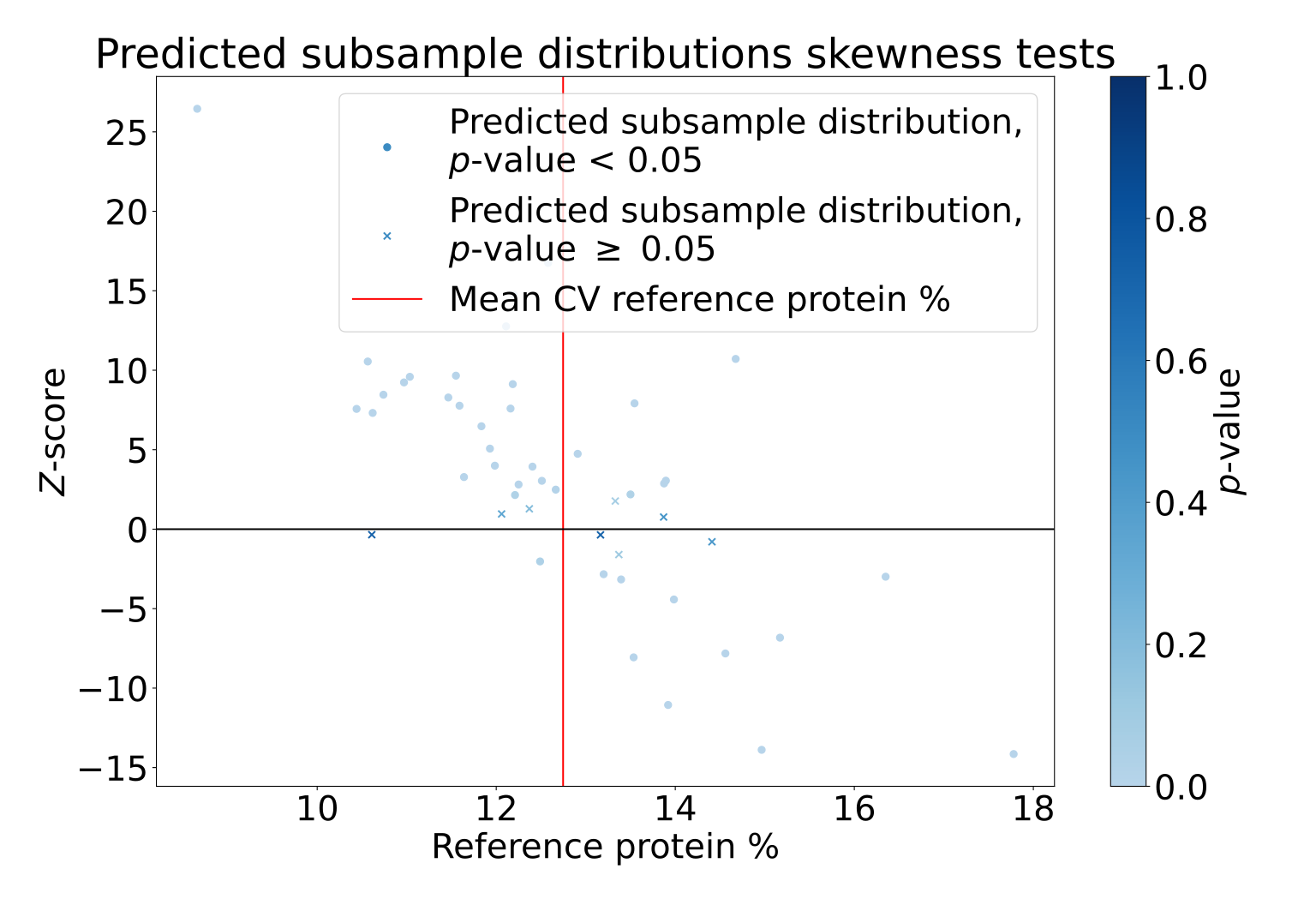}
        \caption{Validation skewness $Z$-tests.}
    \end{subfigure}
    \begin{subfigure}[b]{0.33\textwidth}
        \centering
        \includegraphics[width=\textwidth]{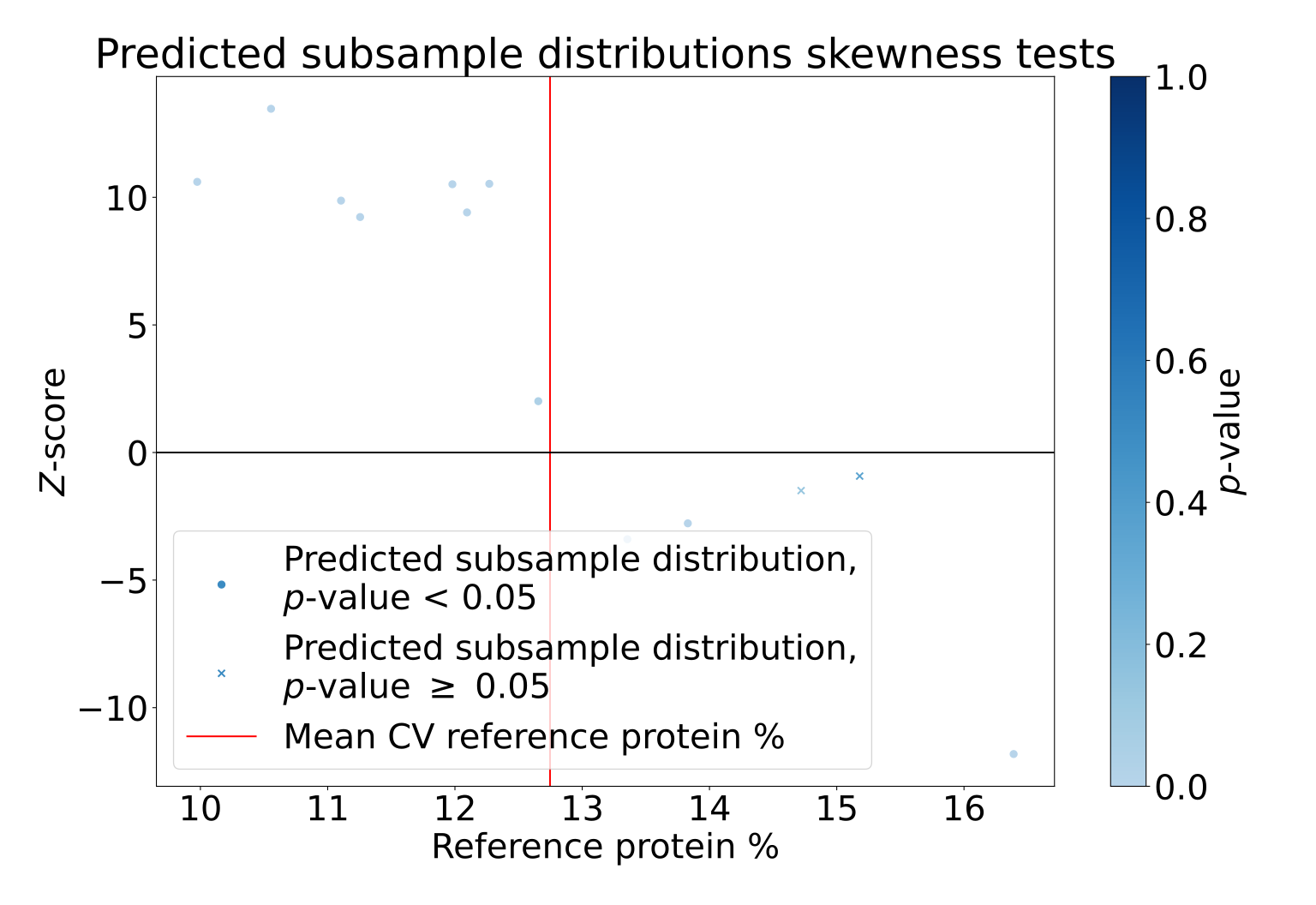}
        \caption{Test skewness $Z$-tests.}
    \end{subfigure}
    \caption{Modified ResNet-18 Regressor subsample prediction distribution skewnesses and skewness tests for each bulk sample as a function of the reference protein of the bulk sample.}
    \label{fig:prediction_skewness_resnet}
\end{figure}

\begin{figure}[htbp]
    \centering
    \begin{subfigure}[b]{0.33\textwidth}
        \centering
        \includegraphics[width=\textwidth]{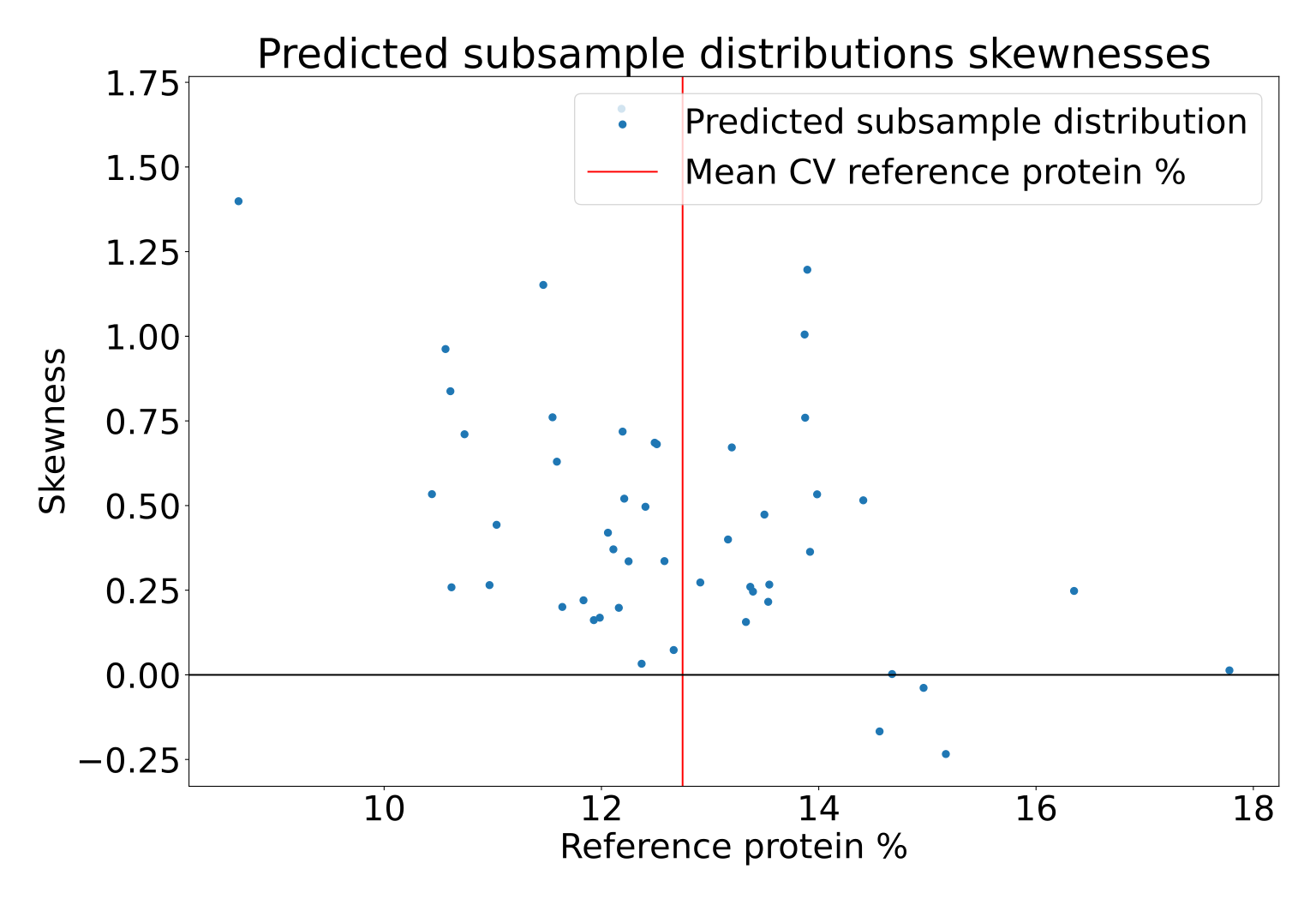}
        \caption{Training skewnesses.}
    \end{subfigure}
    \hfill
    \begin{subfigure}[b]{0.33\textwidth}
        \centering
        \includegraphics[width=\textwidth]{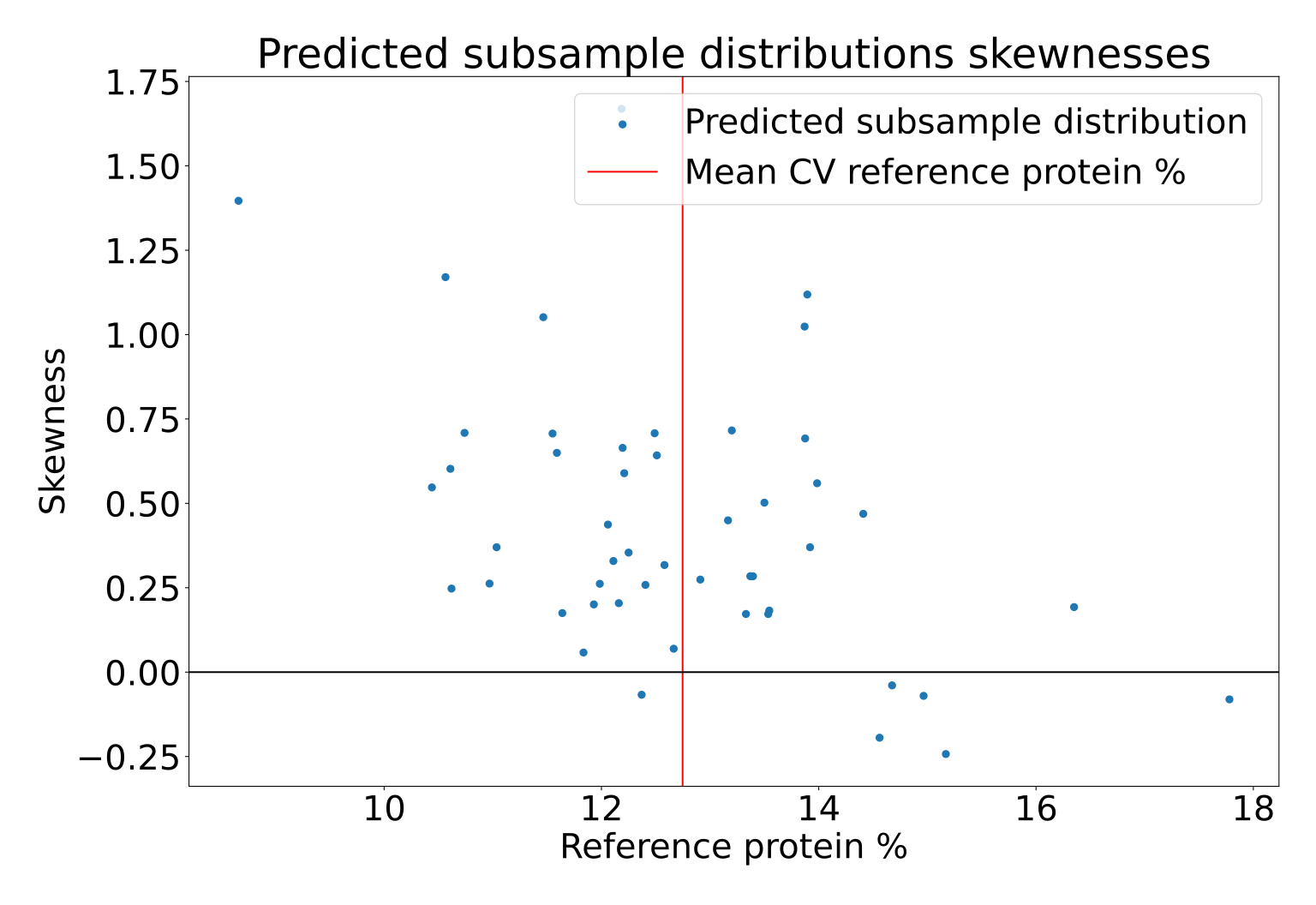}
        \caption{Validation skewnesses.}
    \end{subfigure}
    \begin{subfigure}[b]{0.33\textwidth}
        \centering
        \includegraphics[width=\textwidth]{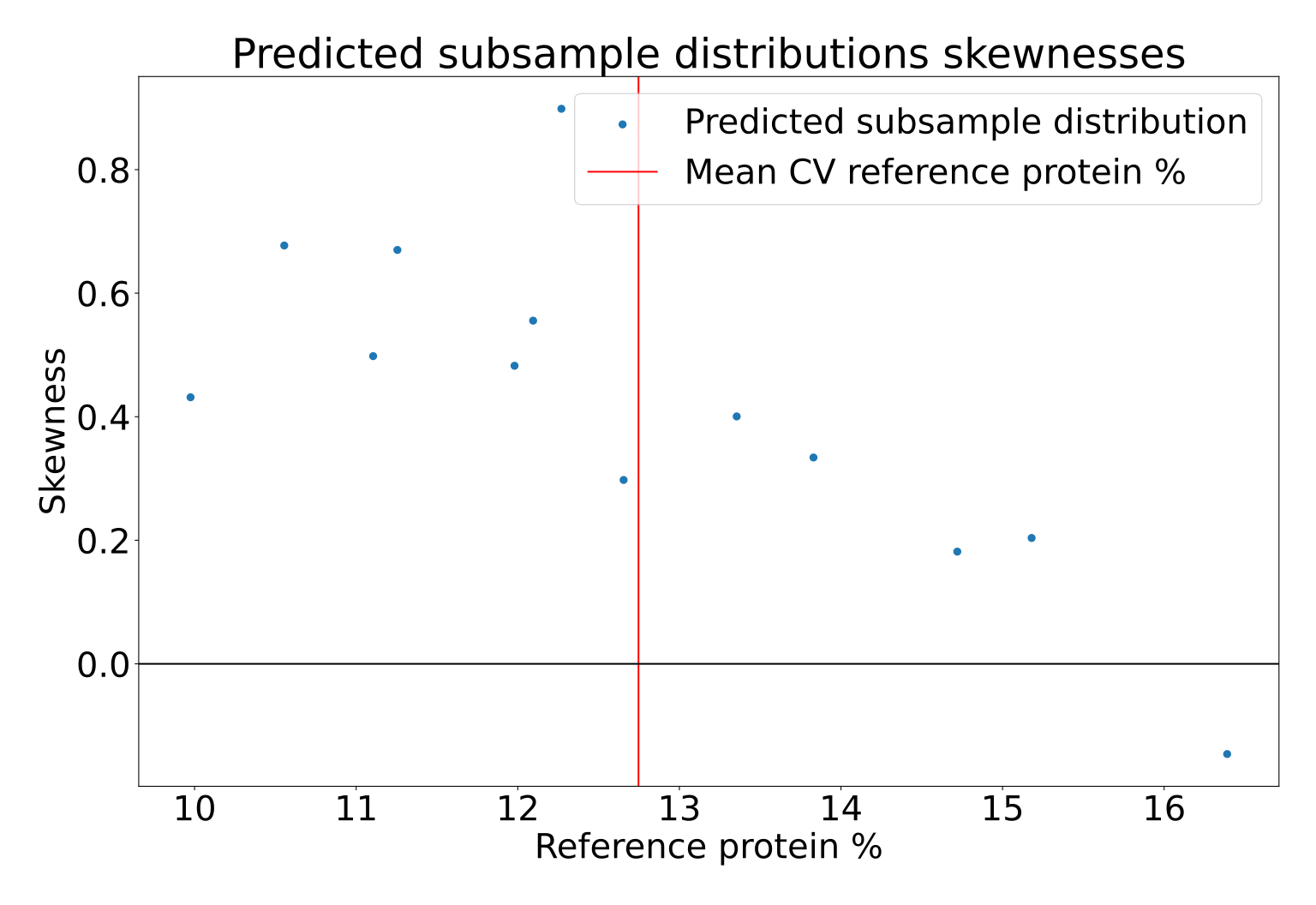}
        \caption{Test skewnesses.}
    \end{subfigure}
    \\
    \centering
    \begin{subfigure}[b]{0.33\textwidth}
        \centering
        \includegraphics[width=\textwidth]{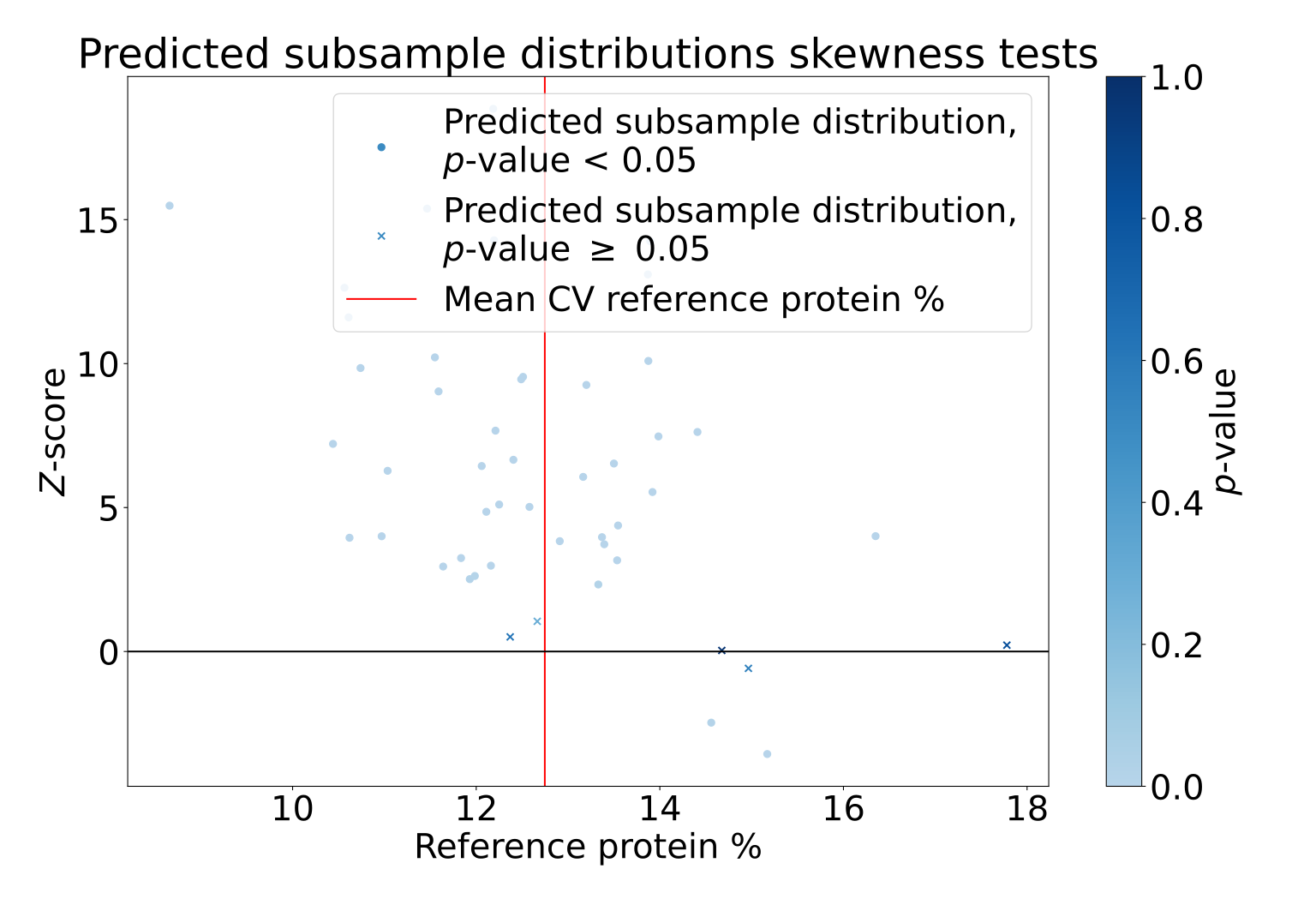}
        \caption{Training skewness $Z$-tests.}
    \end{subfigure}
    \hfill
    \begin{subfigure}[b]{0.33\textwidth}
        \centering
        \includegraphics[width=\textwidth]{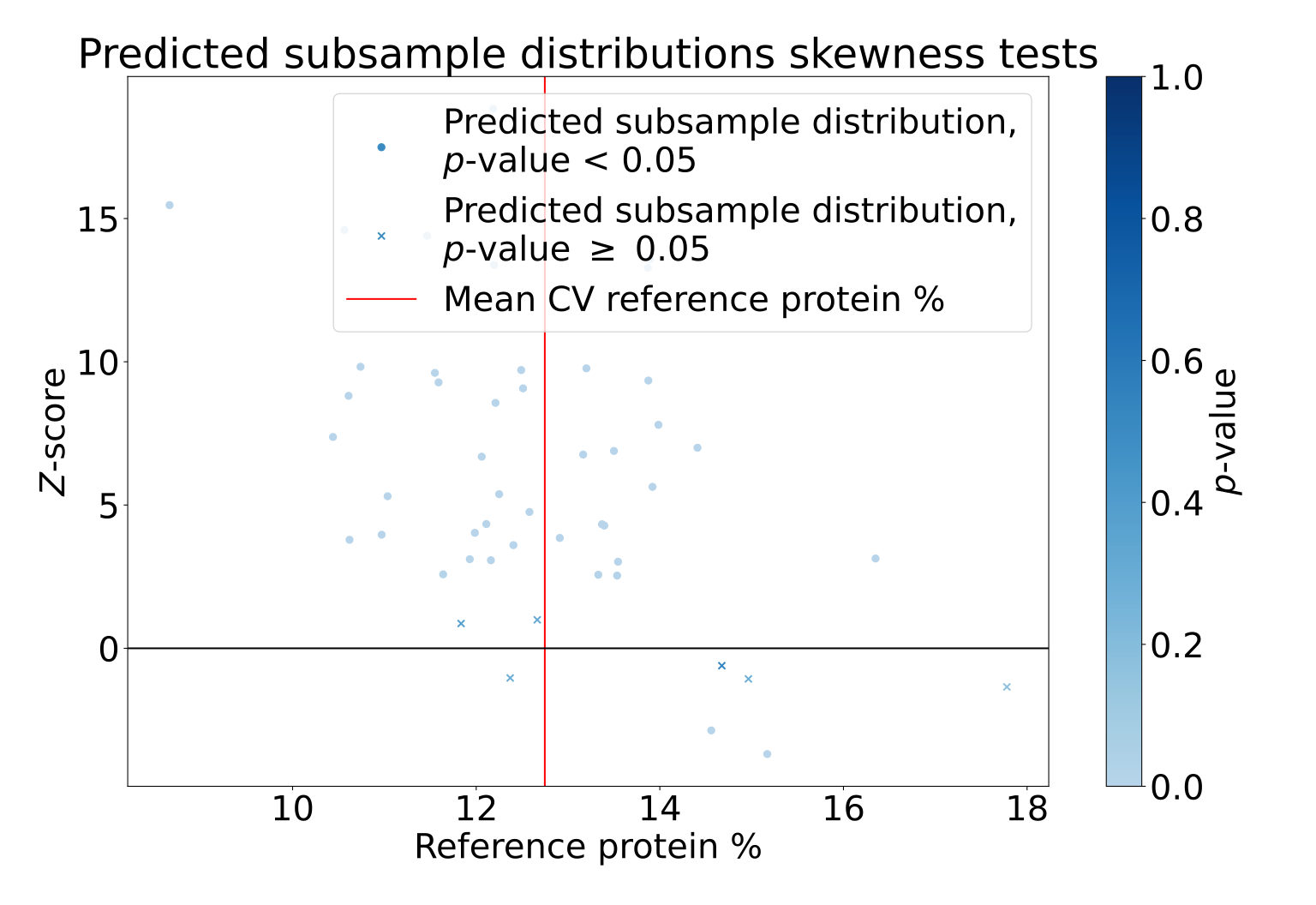}
        \caption{Validation skewness $Z$-tests.}
    \end{subfigure}
    \begin{subfigure}[b]{0.33\textwidth}
        \centering
        \includegraphics[width=\textwidth]{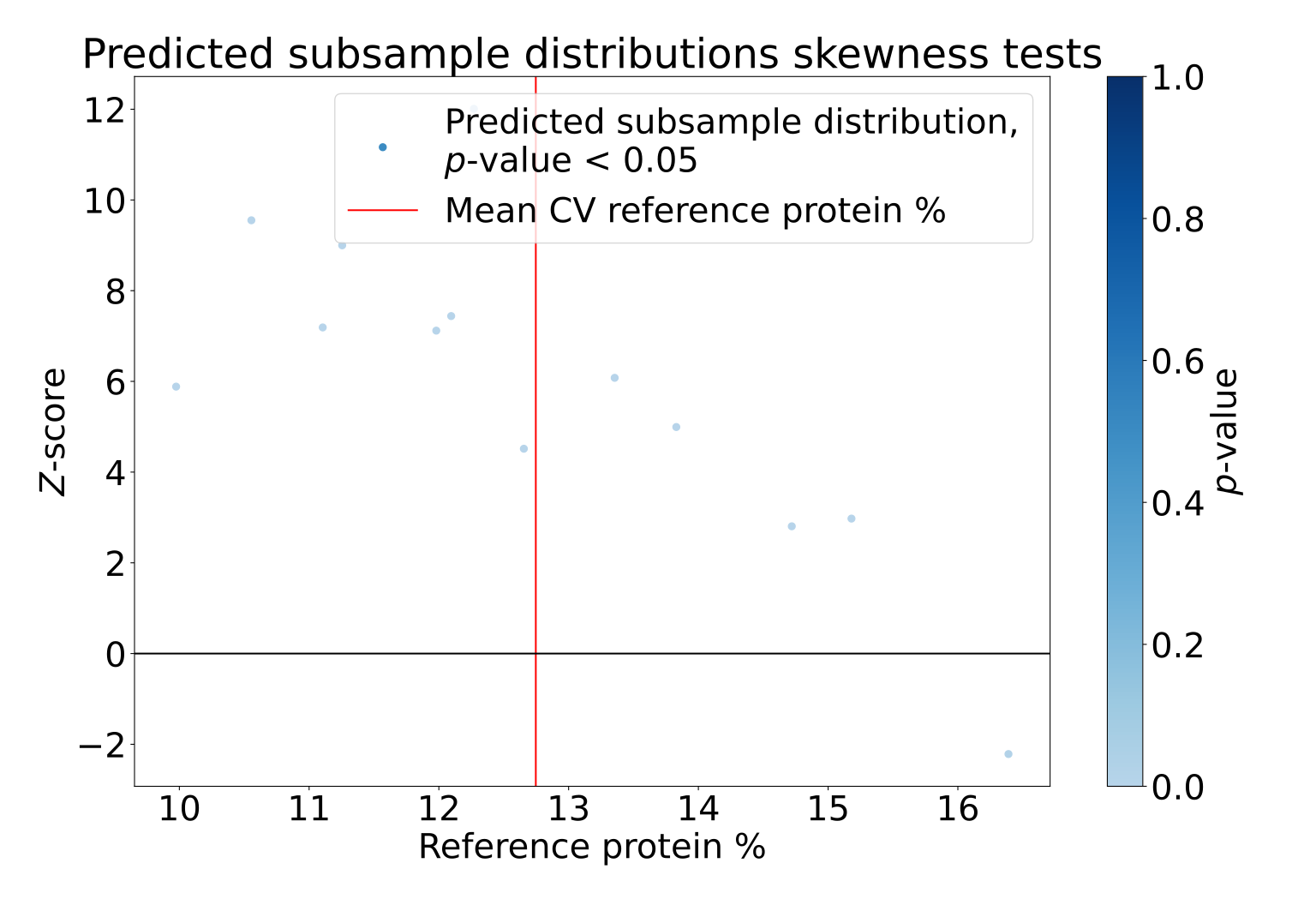}
        \caption{Test skewness $Z$-tests.}
    \end{subfigure}
    \caption{PLS-R subsample prediction distribution skewnesses and skewness tests for each bulk sample as a function of the reference protein of the bulk sample.}
    \label{fig:prediction_skewness_plsr}
\end{figure}

\begin{figure}[htbp]
    \centering
    \begin{subfigure}[b]{0.33\textwidth}
        \centering
        \includegraphics[width=\textwidth]{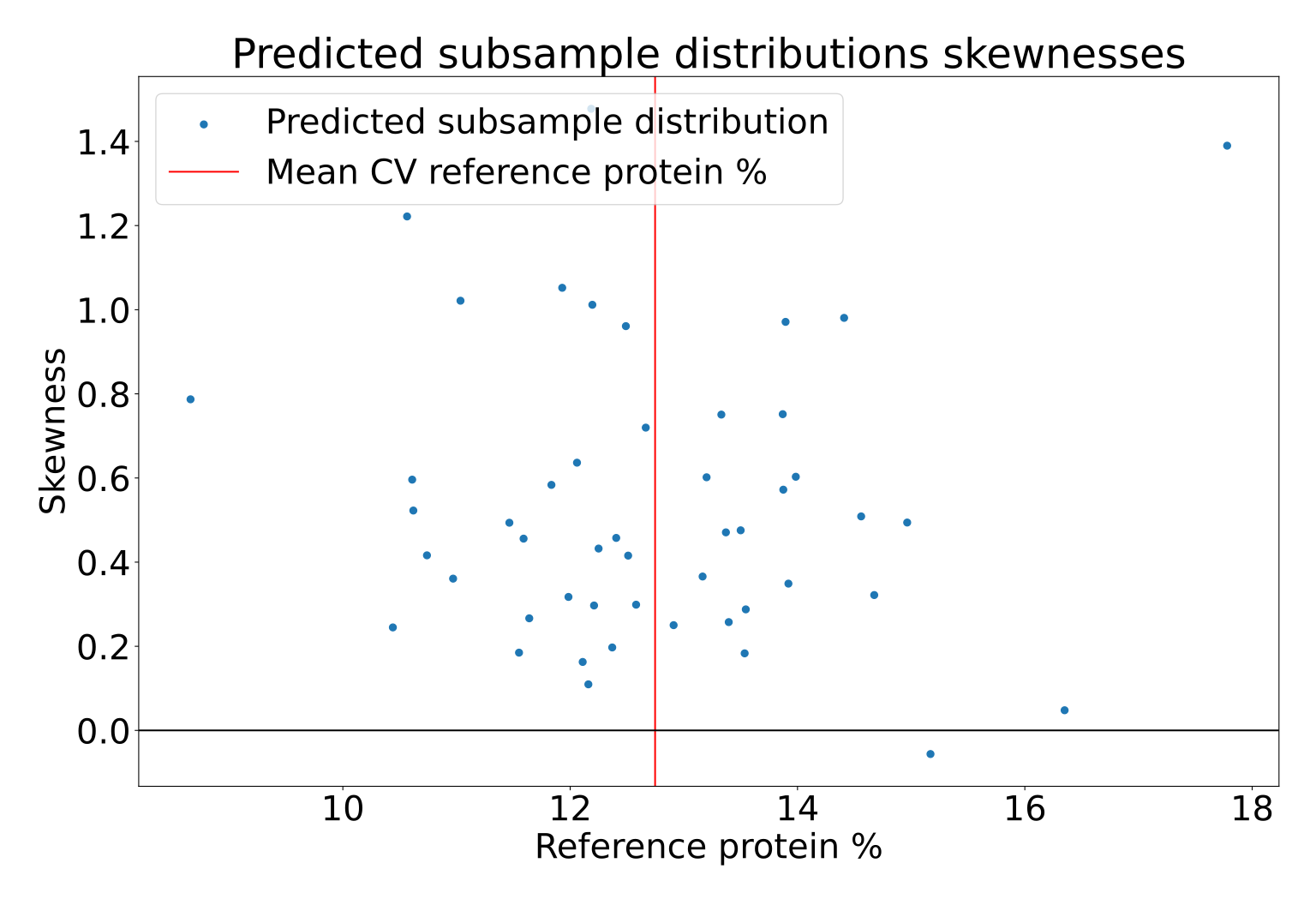}
        \caption{Training skewnesses.}
    \end{subfigure}
    \hfill
    \begin{subfigure}[b]{0.33\textwidth}
        \centering
        \includegraphics[width=\textwidth]{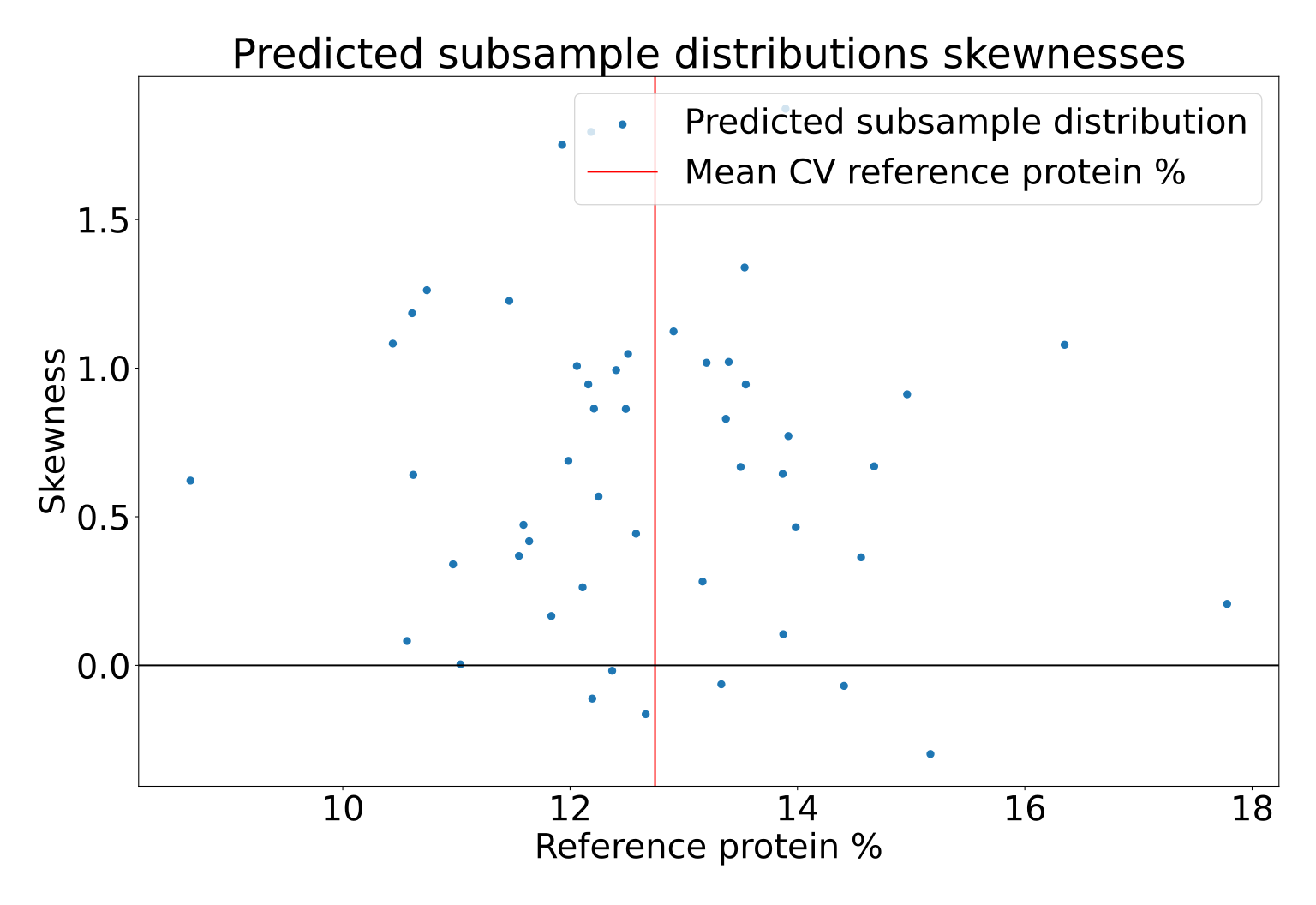}
        \caption{Validation skewnesses.}
    \end{subfigure}
    \begin{subfigure}[b]{0.33\textwidth}
        \centering
        \includegraphics[width=\textwidth]{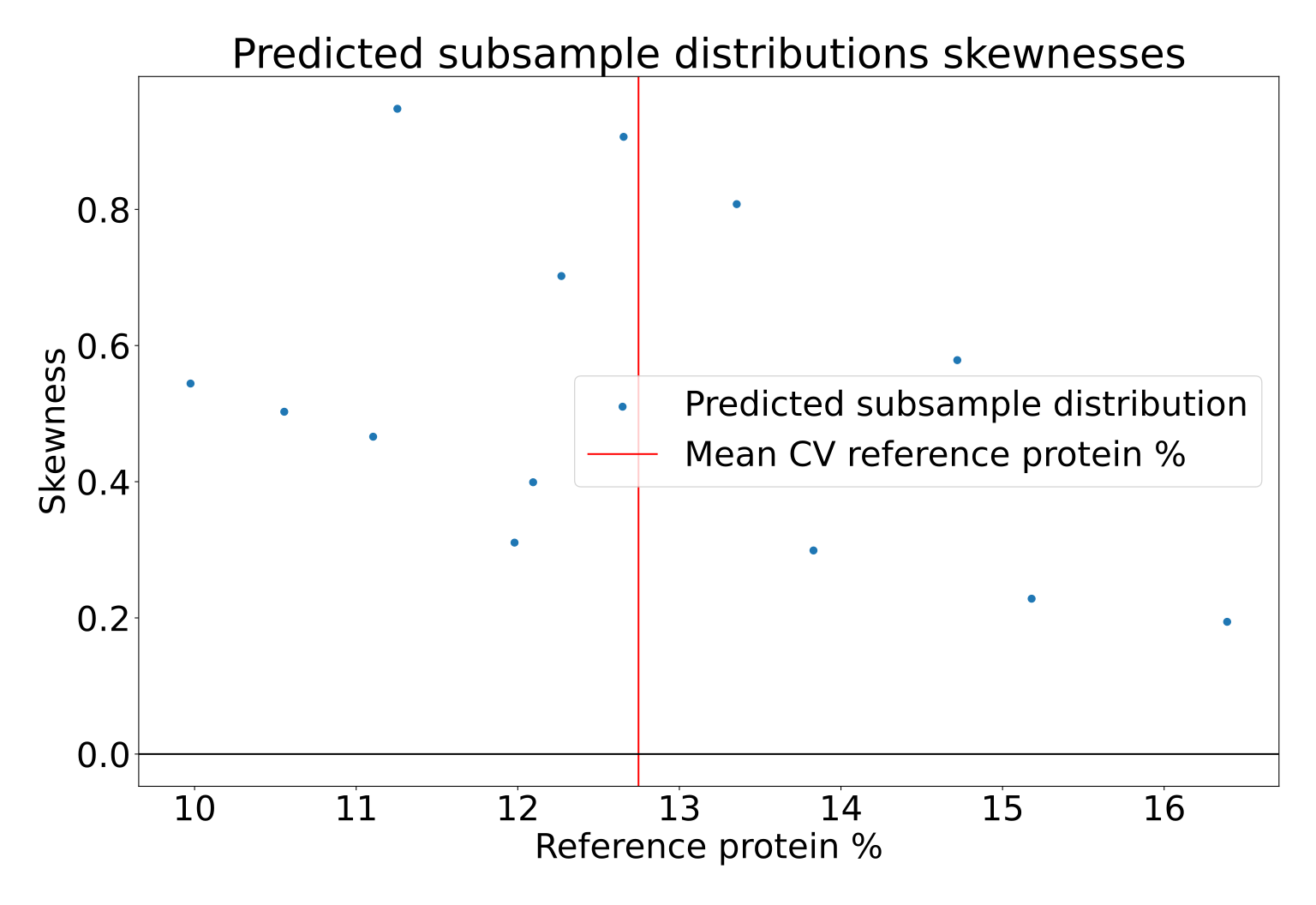}
        \caption{Test skewnesses.}
    \end{subfigure}
    \\
    \centering
    \begin{subfigure}[b]{0.33\textwidth}
        \centering
        \includegraphics[width=\textwidth]{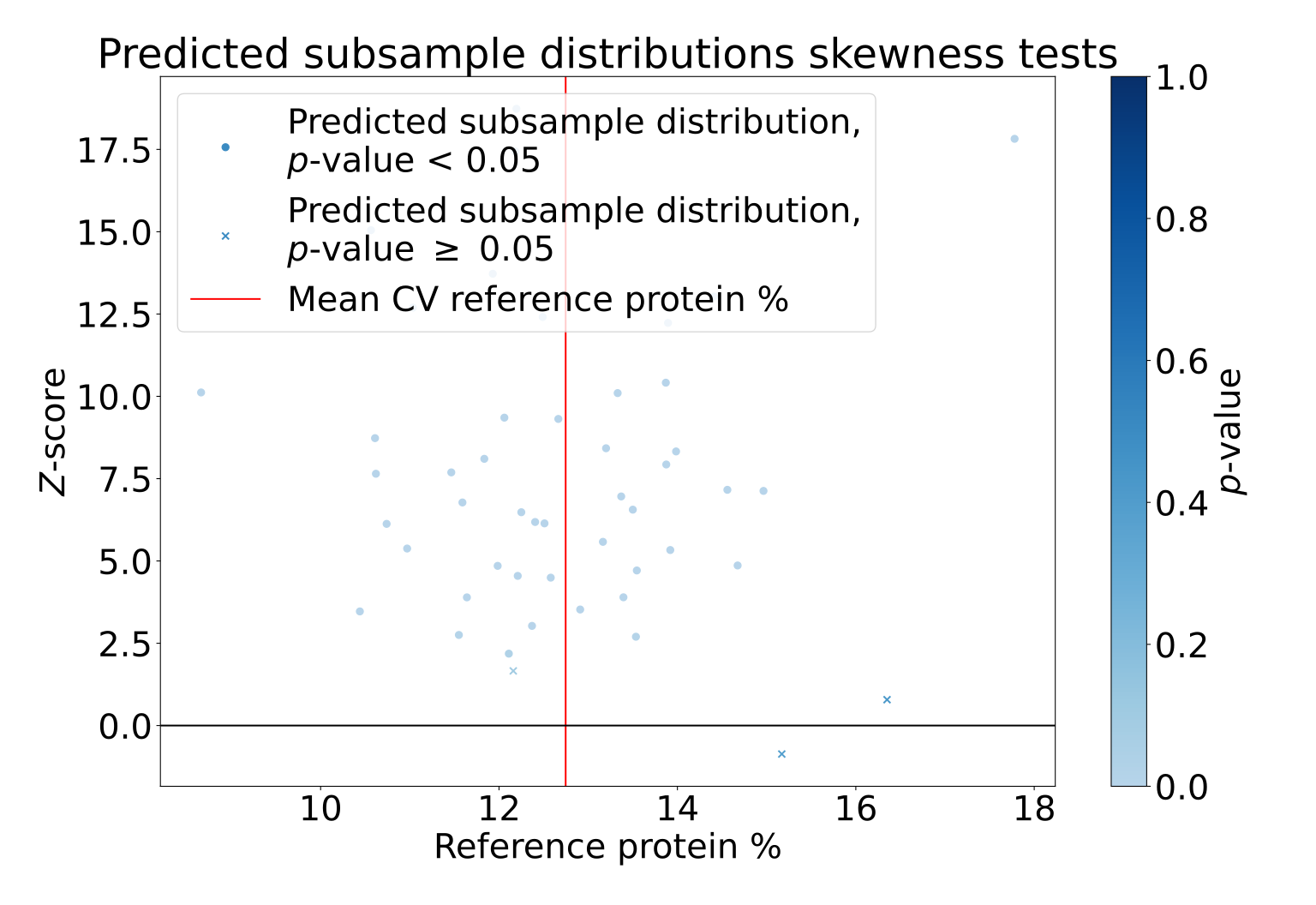}
        \caption{Training skewness $Z$-tests.}
    \end{subfigure}
    \hfill
    \begin{subfigure}[b]{0.33\textwidth}
        \centering
        \includegraphics[width=\textwidth]{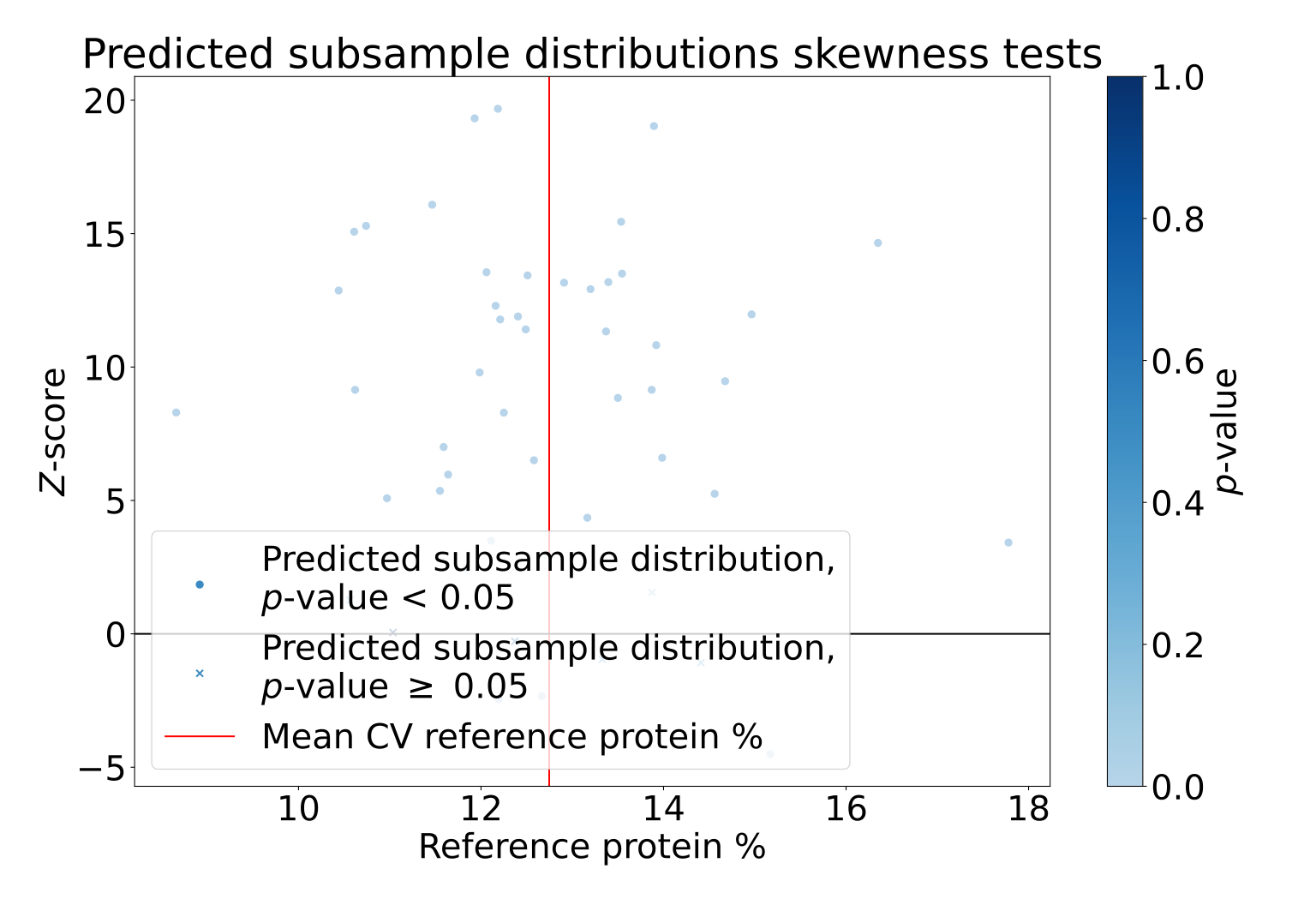}
        \caption{Validation skewness $Z$-tests.}
    \end{subfigure}
    \begin{subfigure}[b]{0.33\textwidth}
        \centering
        \includegraphics[width=\textwidth]{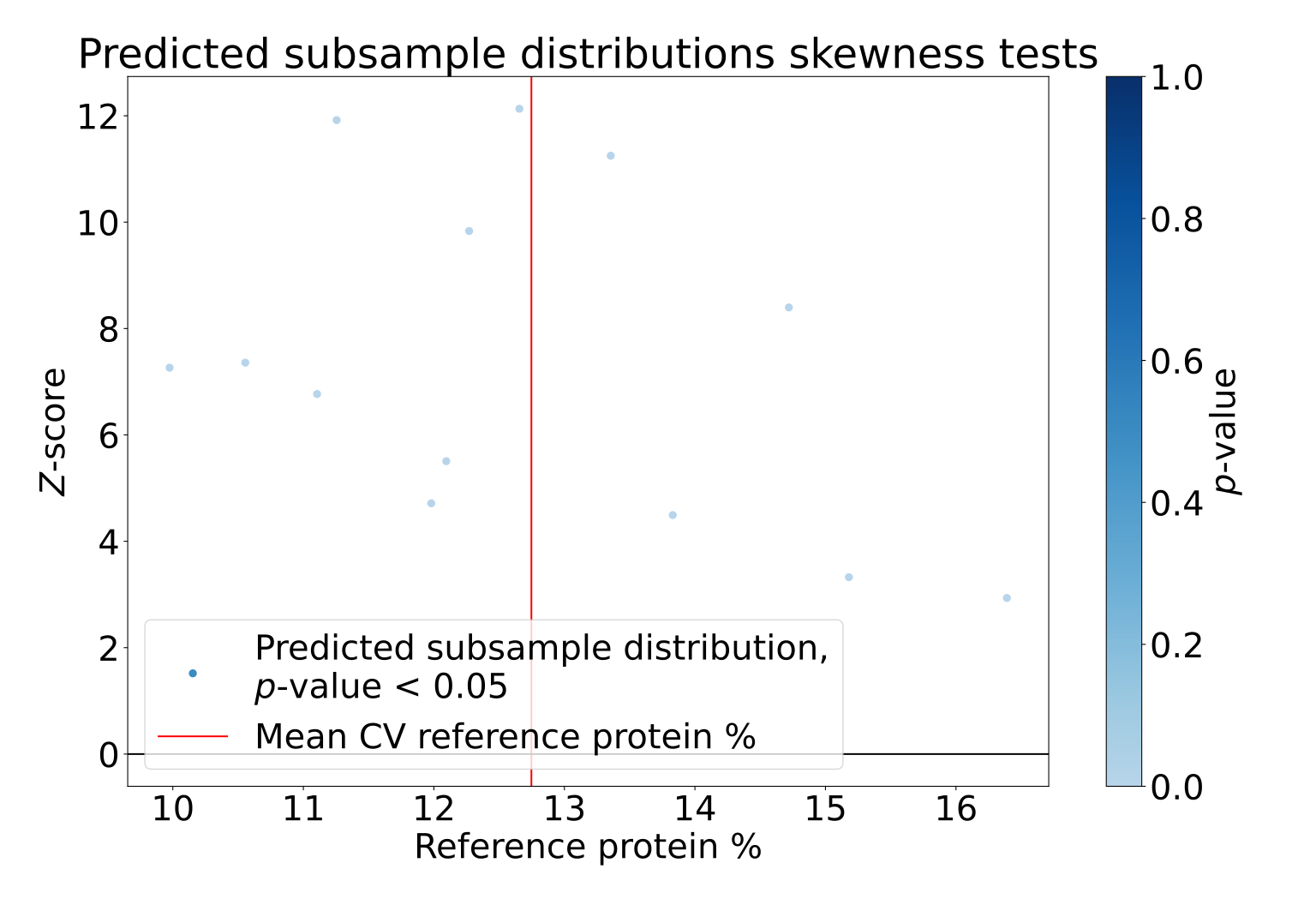}
        \caption{Test skewness $Z$-tests.}
    \end{subfigure}
    \caption{PLS-R$_{\text{bulk}}$ subsample prediction distribution skewnesses and skewness tests for each bulk sample as a function of the reference protein of the bulk sample.}
    \label{fig:prediction_skewness_bulk_plsr}
\end{figure}

Initially, we focus on asymmetry, which can be measured by skewness. If our prediction distributions are symmetric, they will have a skewness of $0$. We note that the reverse relationship does not necessarily hold - i.e., $0$ skewness does not imply a symmetric distribution. However, if a distribution has non-zero skewness, it is not symmetric. We plot the skewnesses, $Z$-scores, and related $p$-values for Modified ResNet-18 Regressor, PLS-R, and PLS-R$_{\text{bulk}}$ in \myfigref{fig:prediction_skewness_resnet}, \myfigref{fig:prediction_skewness_plsr}, and \myfigref{fig:prediction_skewness_bulk_plsr}, respectively. These tests imply that, beyond any reasonable doubt, most of the prediction distributions are too skewed to be normally distributed and are, in fact, asymmetric.

\begin{figure}[htbp]
    \centering
    \begin{subfigure}[b]{0.33\textwidth}
        \centering
        \includegraphics[width=\textwidth]{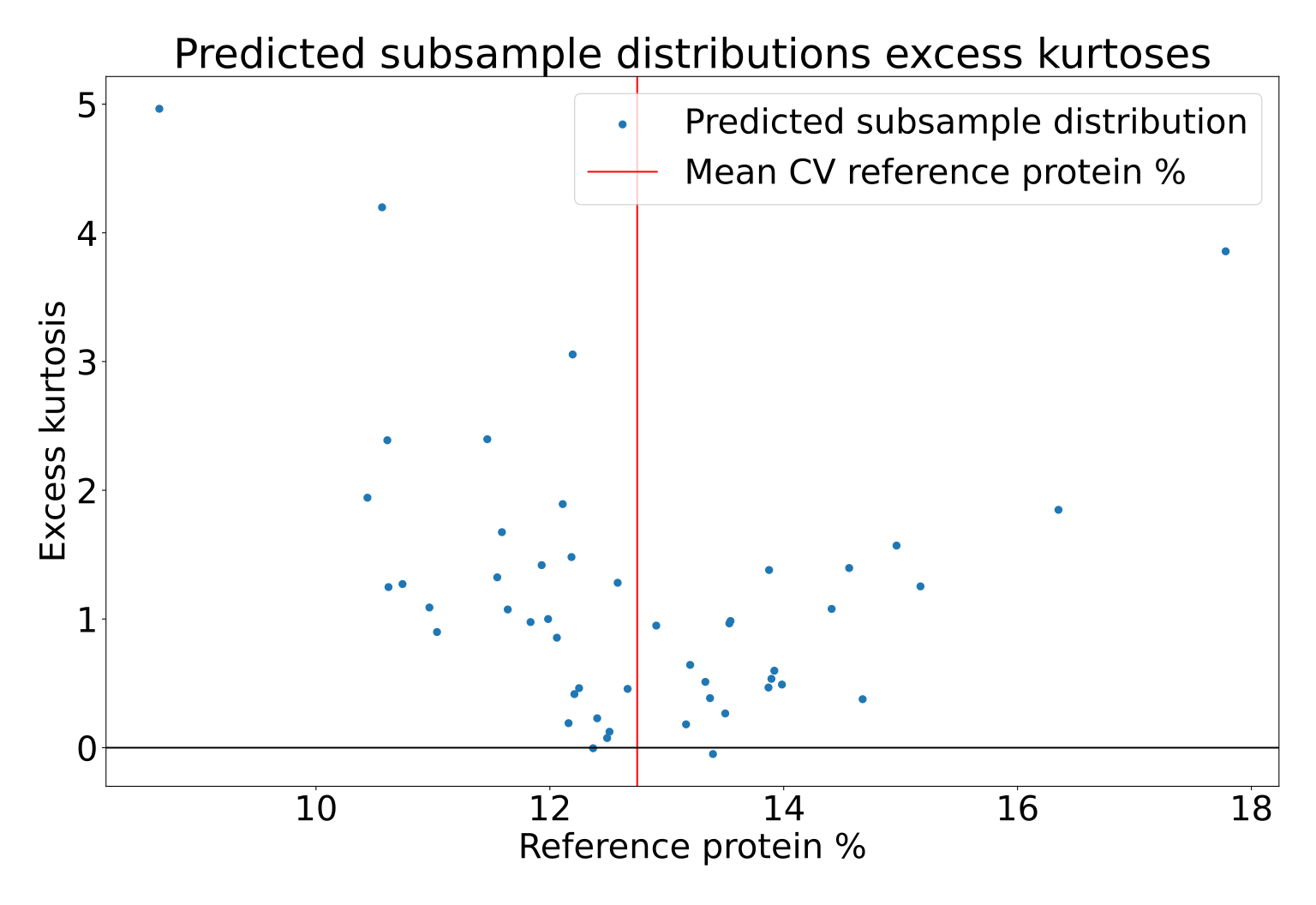}
        \caption{Training kurtoses.}
    \end{subfigure}
    \hfill
    \begin{subfigure}[b]{0.33\textwidth}
        \centering
        \includegraphics[width=\textwidth]{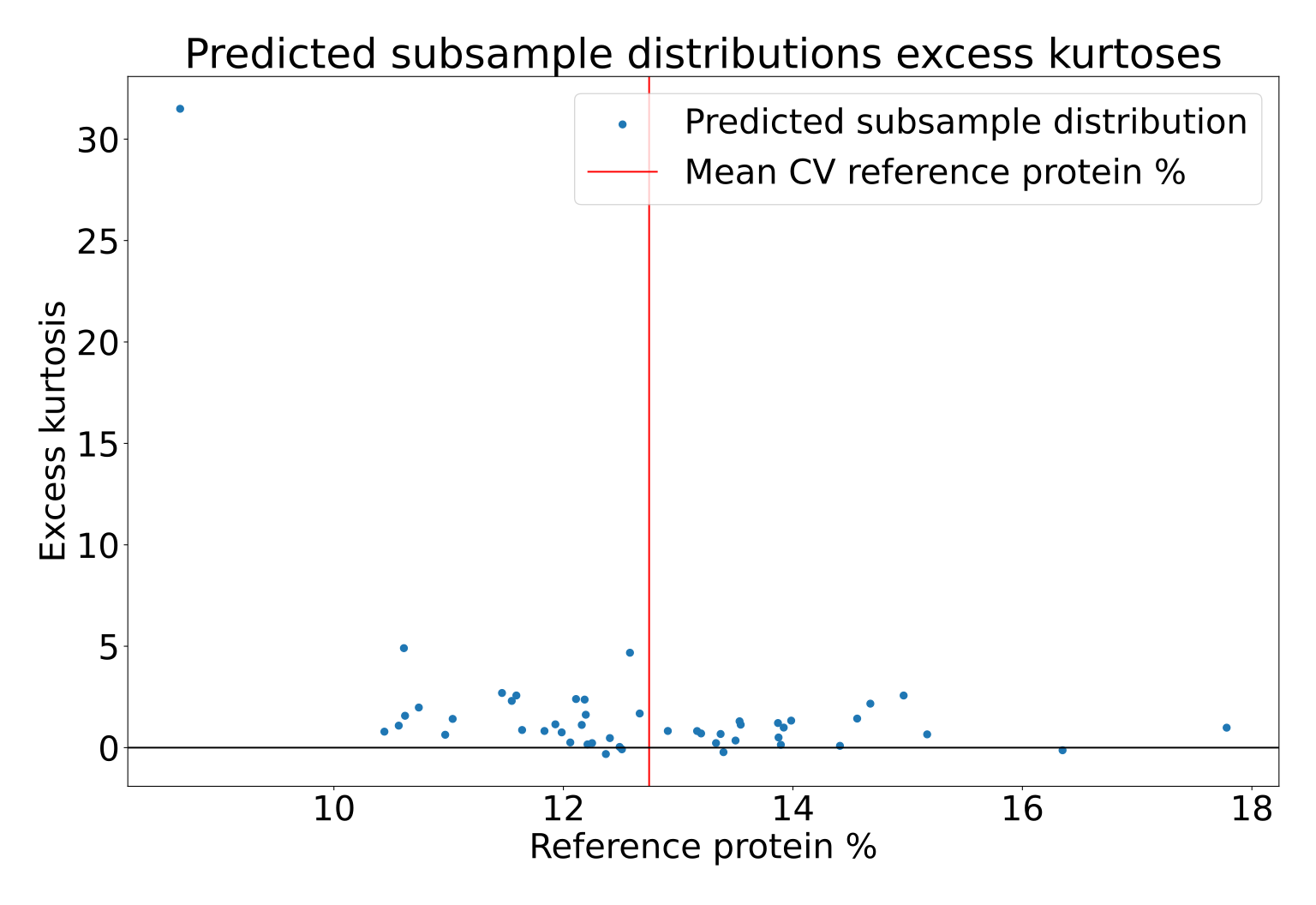}
        \caption{Validation kurtoses.}
    \end{subfigure}
    \begin{subfigure}[b]{0.33\textwidth}
        \centering
        \includegraphics[width=\textwidth]{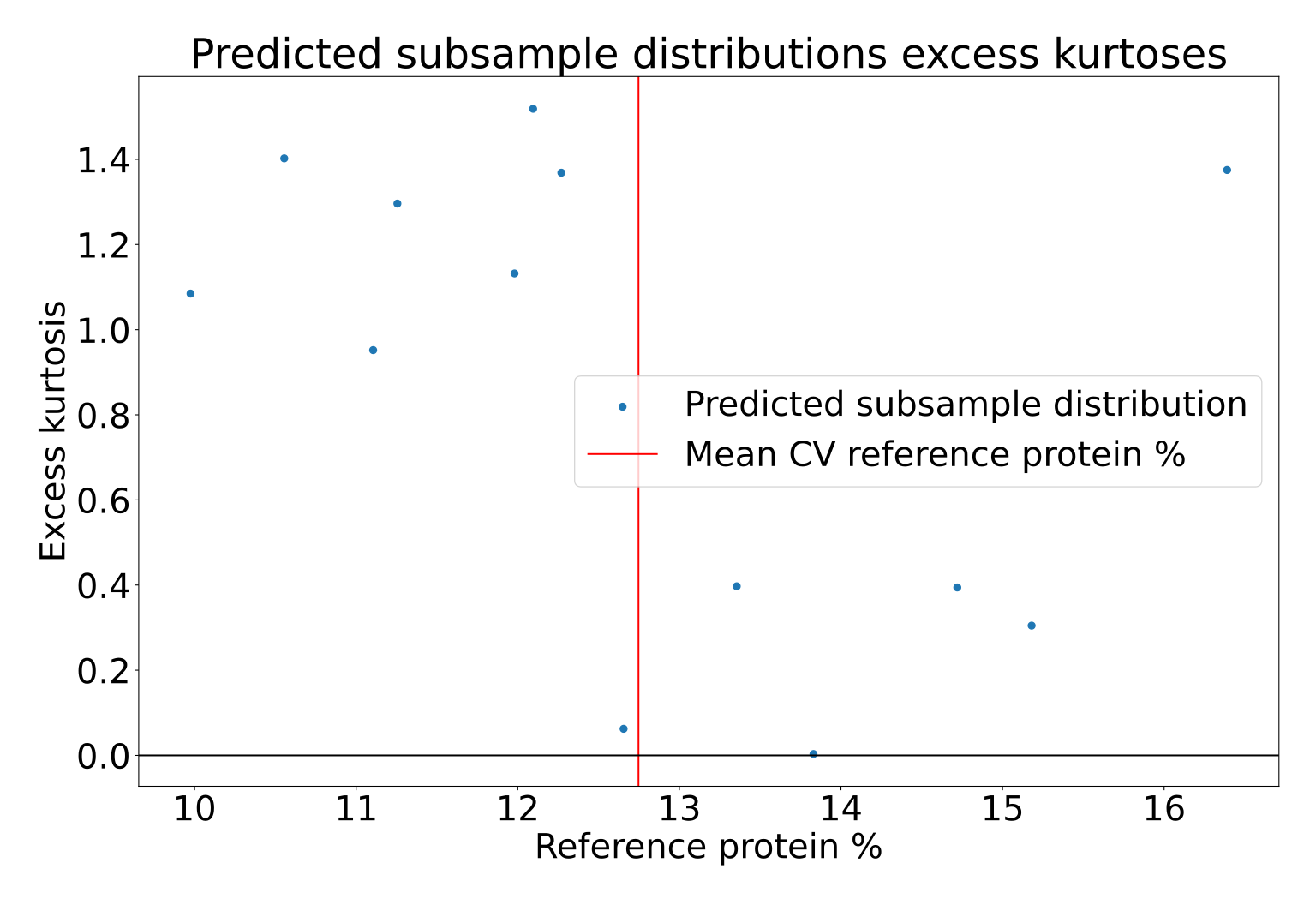}
        \caption{Test kurtoses.}
    \end{subfigure}
    \\
    \centering
    \begin{subfigure}[b]{0.33\textwidth}
        \centering
        \includegraphics[width=\textwidth]{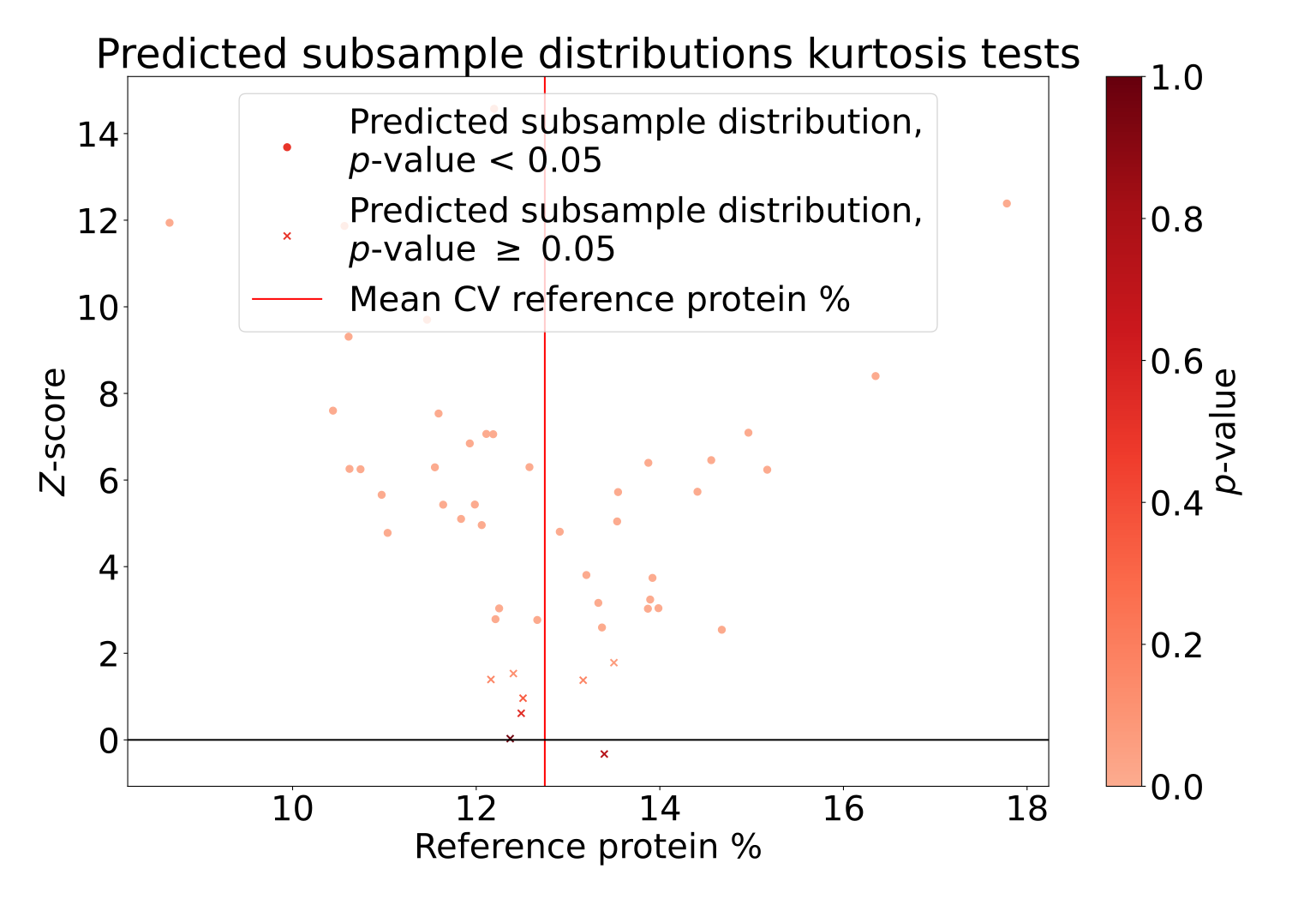}
        \caption{Training kurtosis $Z$-tests.}
    \end{subfigure}
    \hfill
    \begin{subfigure}[b]{0.33\textwidth}
        \centering
        \includegraphics[width=\textwidth]{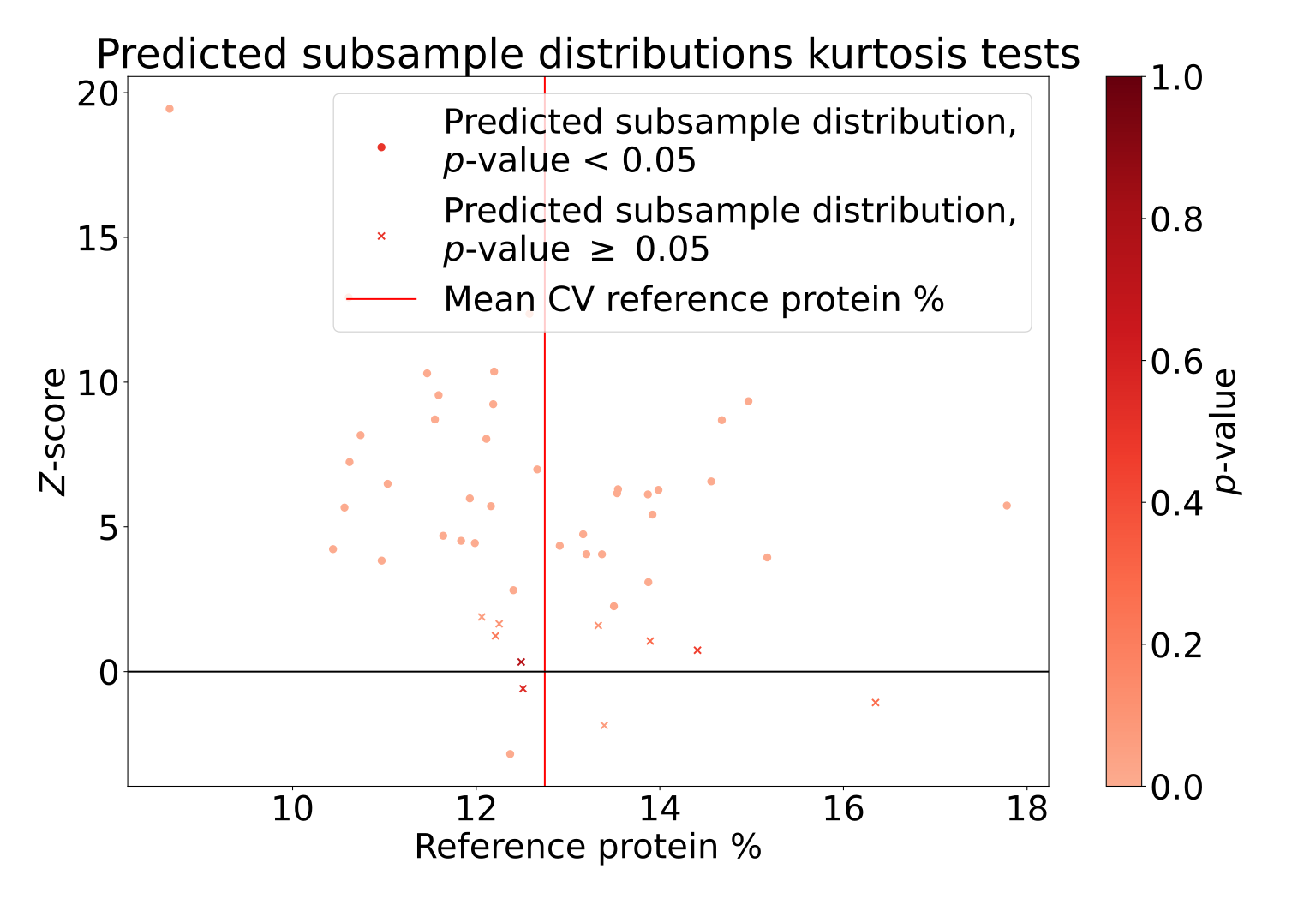}
        \caption{Validation kurtosis $Z$-tests.}
    \end{subfigure}
    \begin{subfigure}[b]{0.33\textwidth}
        \centering
        \includegraphics[width=\textwidth]{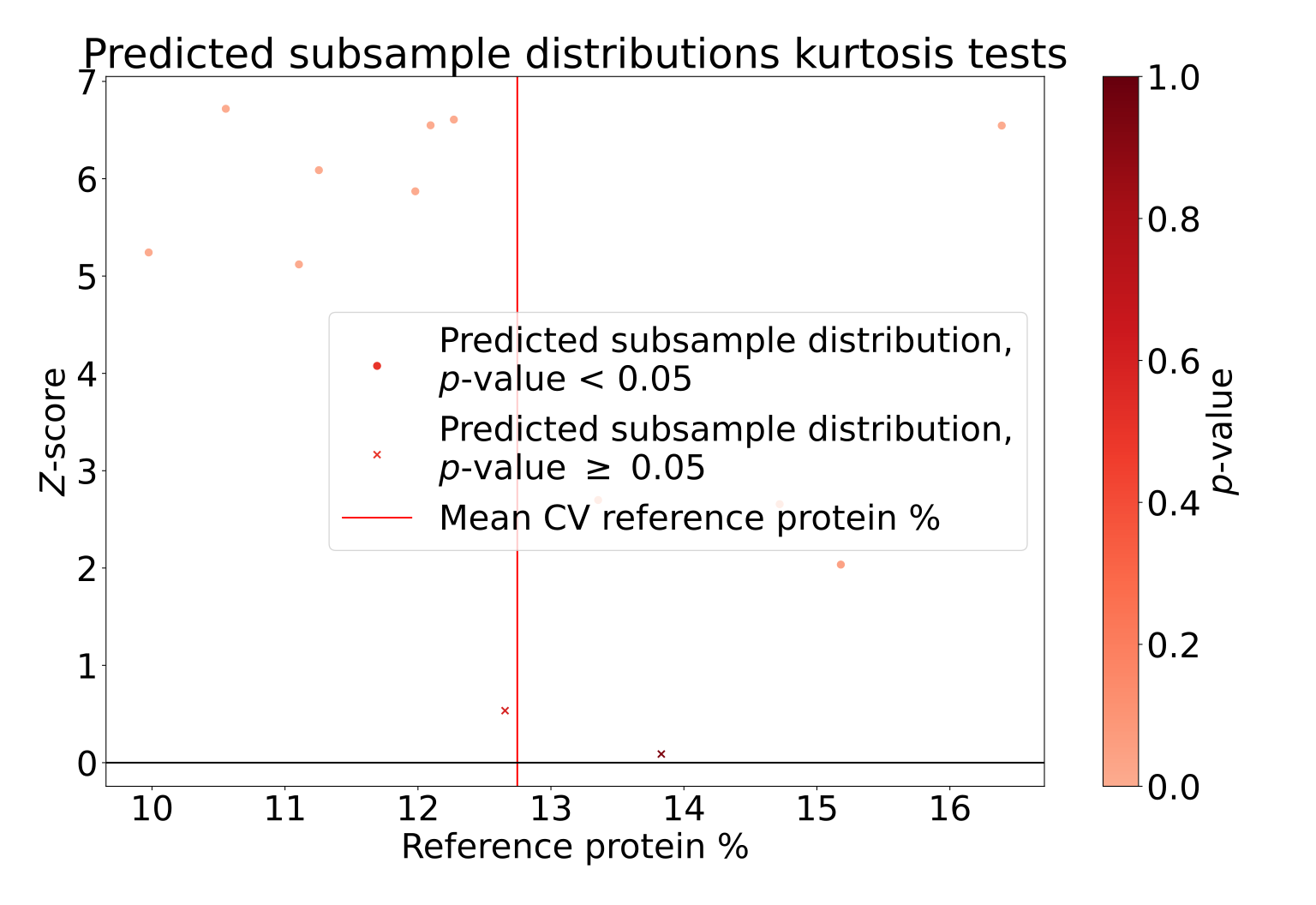}
        \caption{Test kurtosis $Z$-tests.}
    \end{subfigure}
    \caption{Modified ResNet-18 Regressor subsample prediction distribution kurtoses and kurtosis tests for each bulk sample as a function of the reference protein of the bulk sample.}
    \label{fig:prediction_excess_kurtosis_resnet}
\end{figure}

\begin{figure}[htbp]
    \centering
    \begin{subfigure}[b]{0.33\textwidth}
        \centering
        \includegraphics[width=\textwidth]{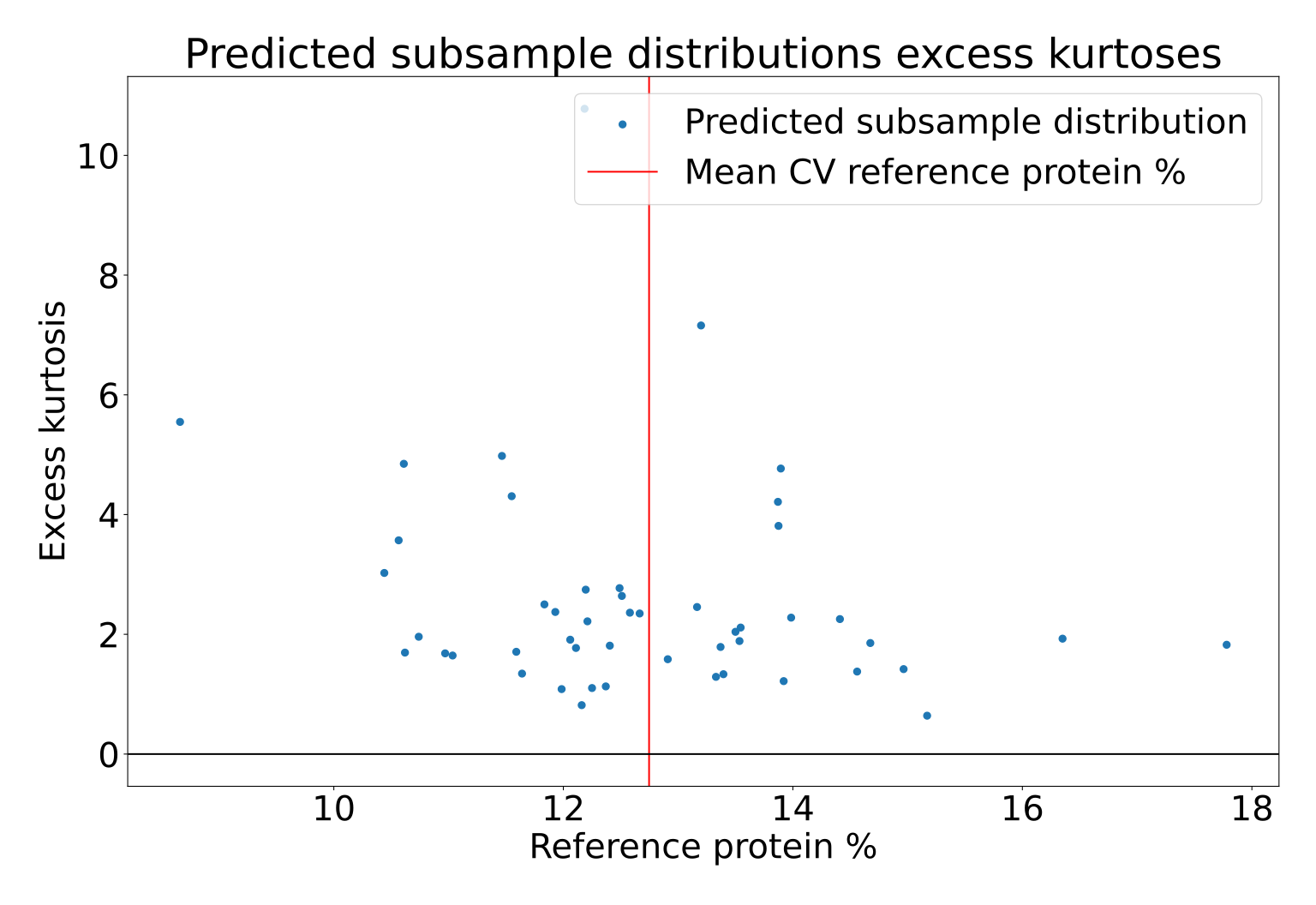}
        \caption{Training kurtoses.}
    \end{subfigure}
    \hfill
    \begin{subfigure}[b]{0.33\textwidth}
        \centering
        \includegraphics[width=\textwidth]{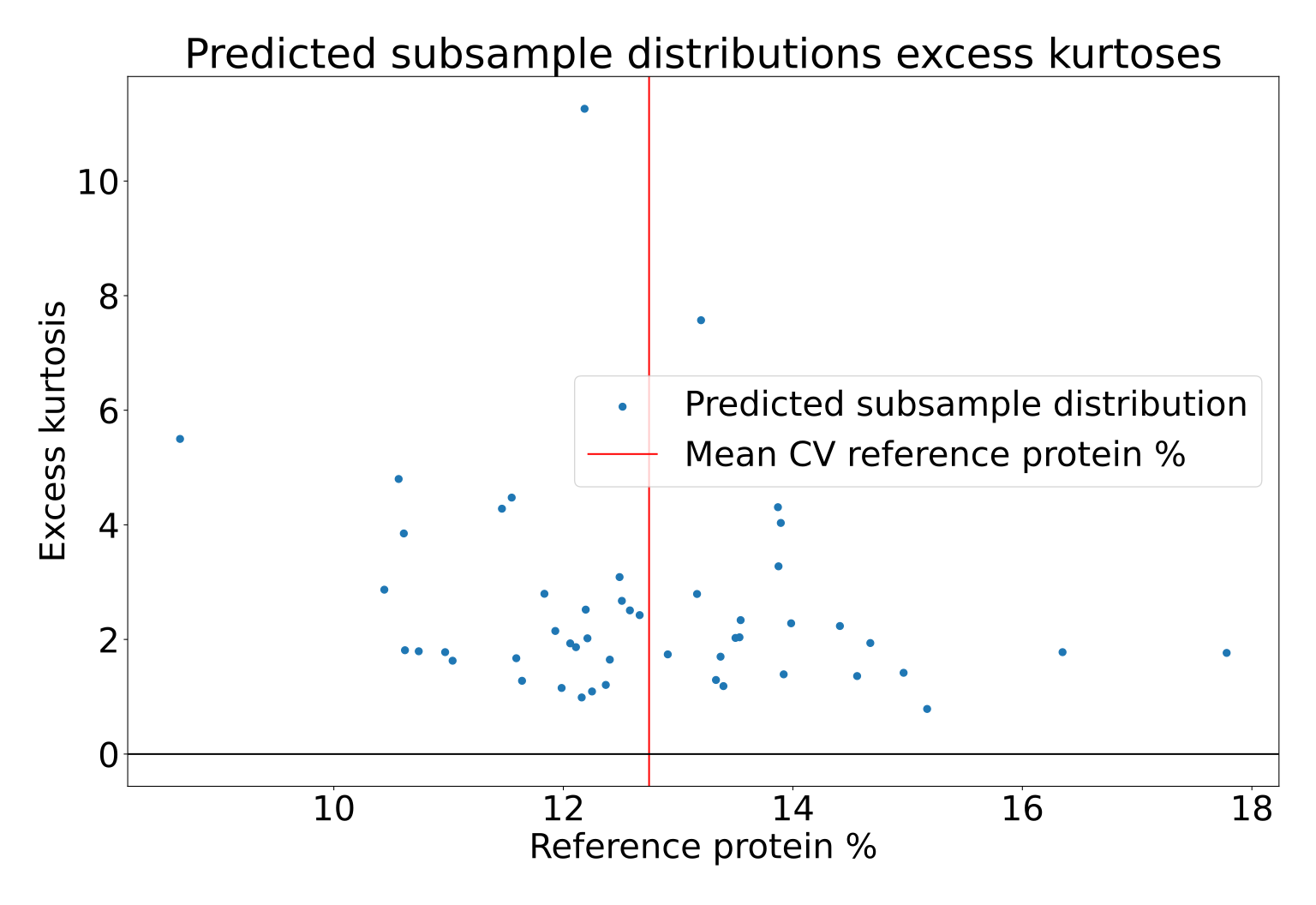}
        \caption{Validation kurtoses.}
    \end{subfigure}
    \begin{subfigure}[b]{0.33\textwidth}
        \centering
        \includegraphics[width=\textwidth]{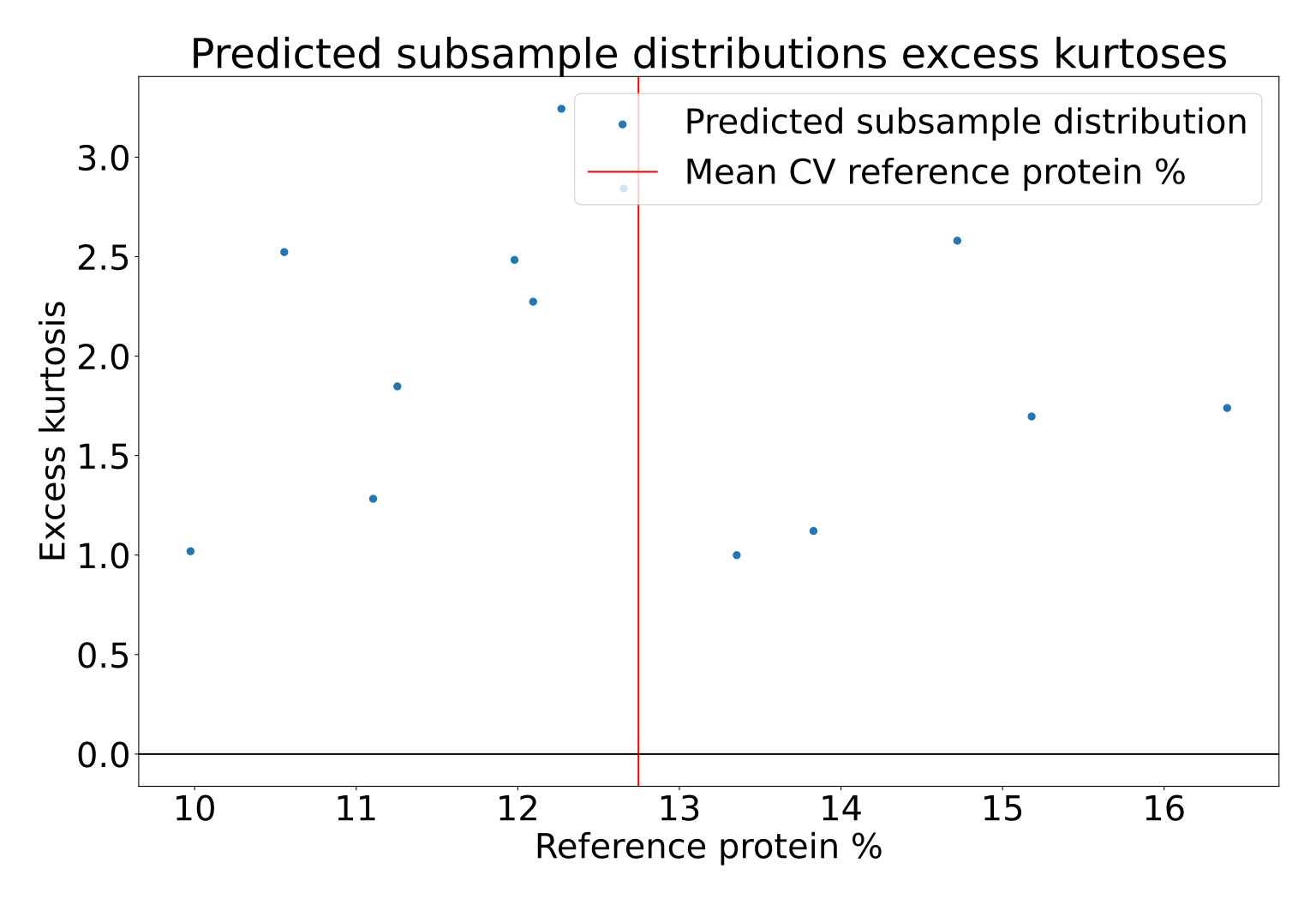}
        \caption{Test kurtoses.}
    \end{subfigure}
    \\
    \centering
    \begin{subfigure}[b]{0.33\textwidth}
        \centering
        \includegraphics[width=\textwidth]{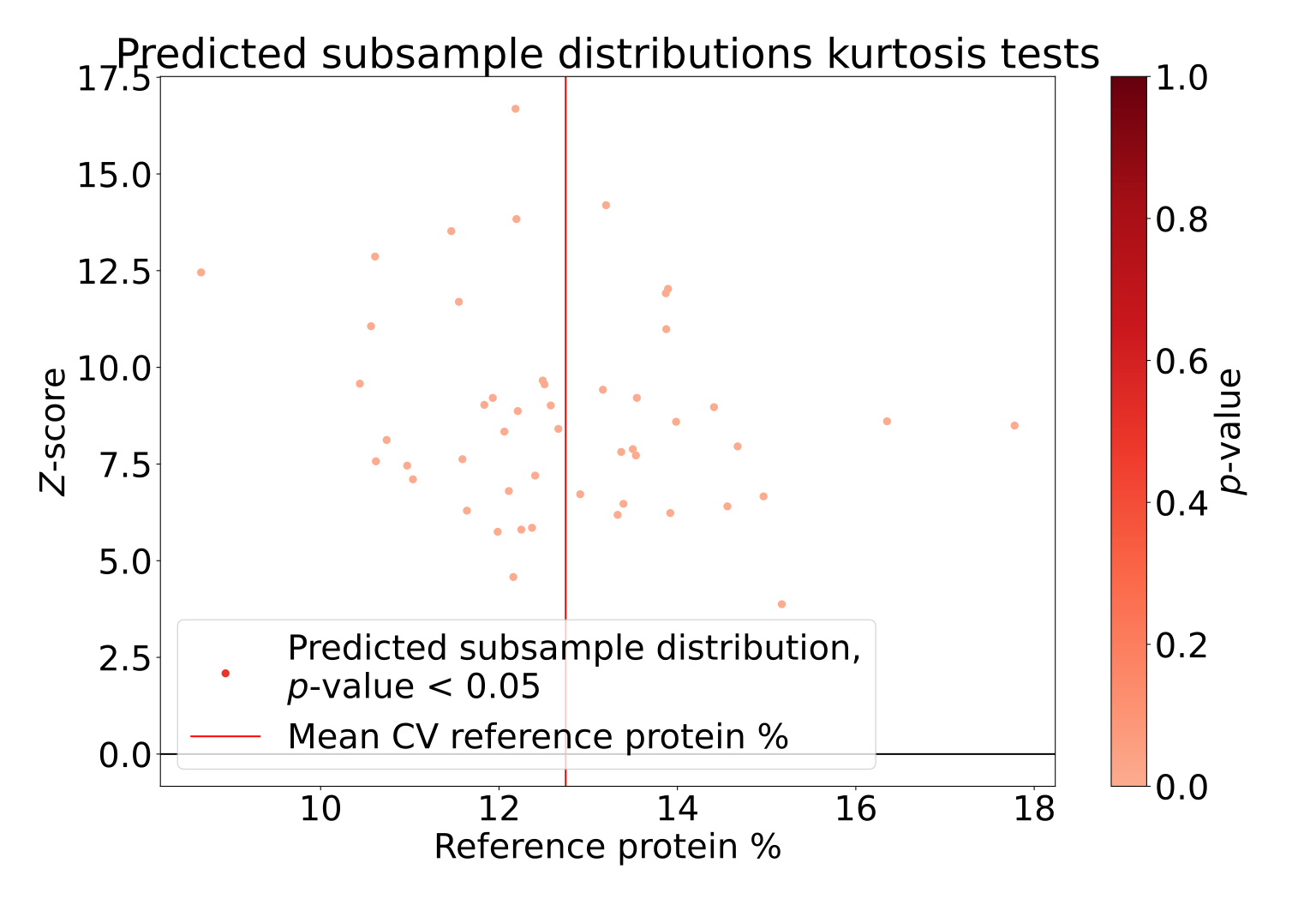}
        \caption{Training kurtosis $Z$-tests.}
    \end{subfigure}
    \hfill
    \begin{subfigure}[b]{0.33\textwidth}
        \centering
        \includegraphics[width=\textwidth]{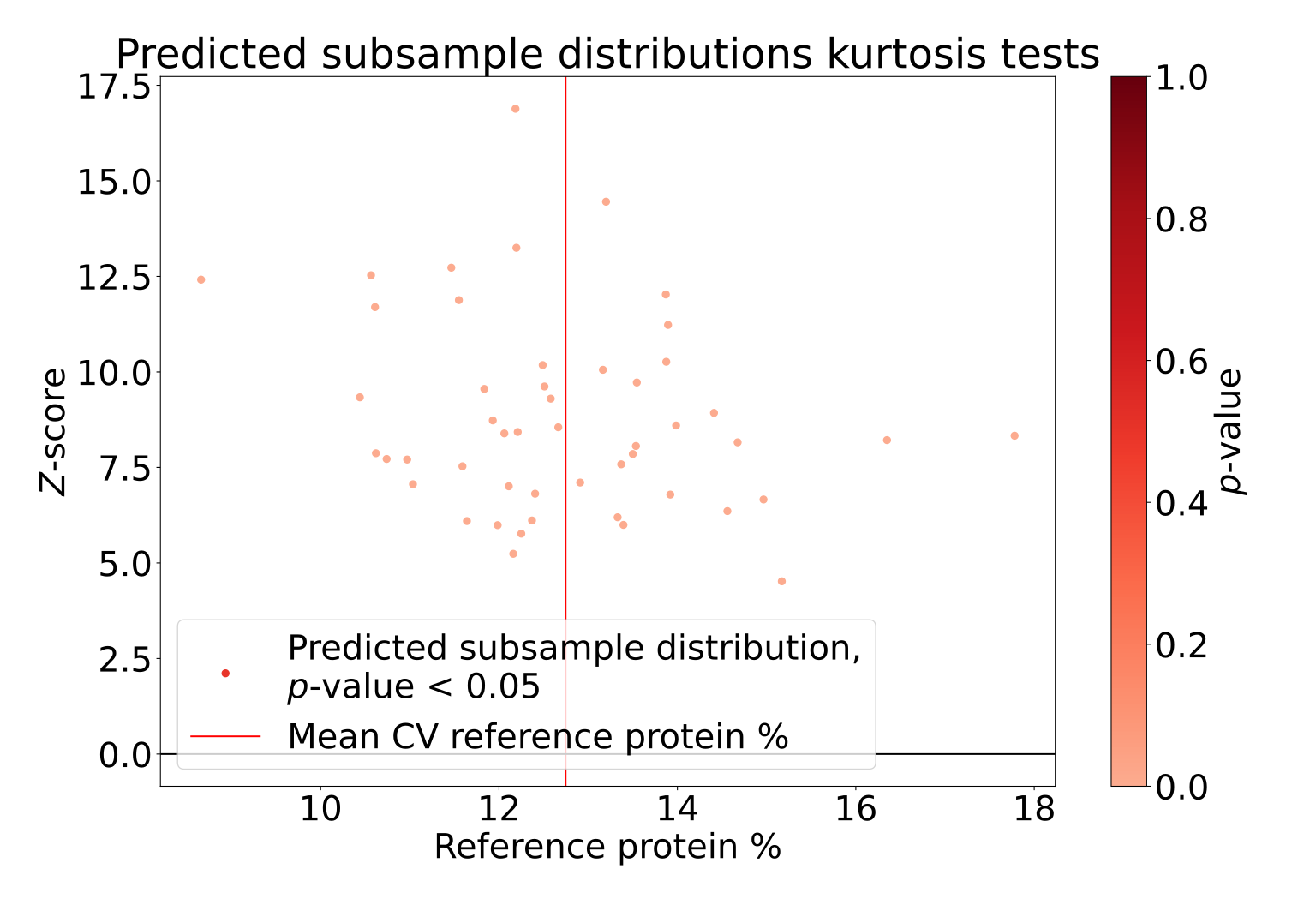}
        \caption{Validation kurtosis $Z$-tests.}
    \end{subfigure}
    \begin{subfigure}[b]{0.33\textwidth}
        \centering
        \includegraphics[width=\textwidth]{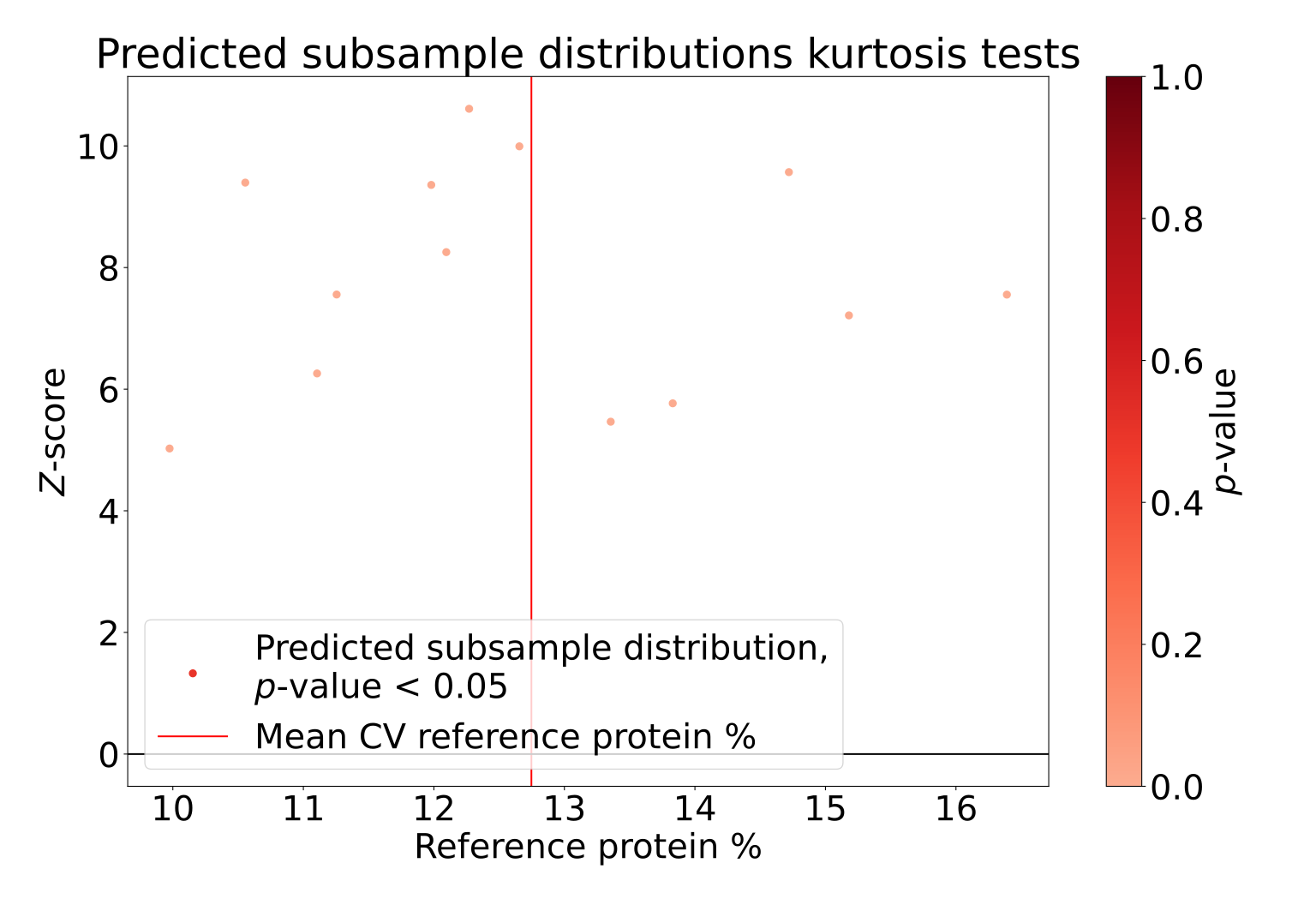}
        \caption{Test kurtosis $Z$-tests.}
    \end{subfigure}
    \caption{PLS-R subsample prediction distribution kurtoses and kurtosis tests for each bulk sample as a function of the reference protein of the bulk sample.}
    \label{fig:prediction_excess_kurtosis_plsr}
\end{figure}

\begin{figure}[htbp]
    \centering
    \begin{subfigure}[b]{0.33\textwidth}
        \centering
        \includegraphics[width=\textwidth]{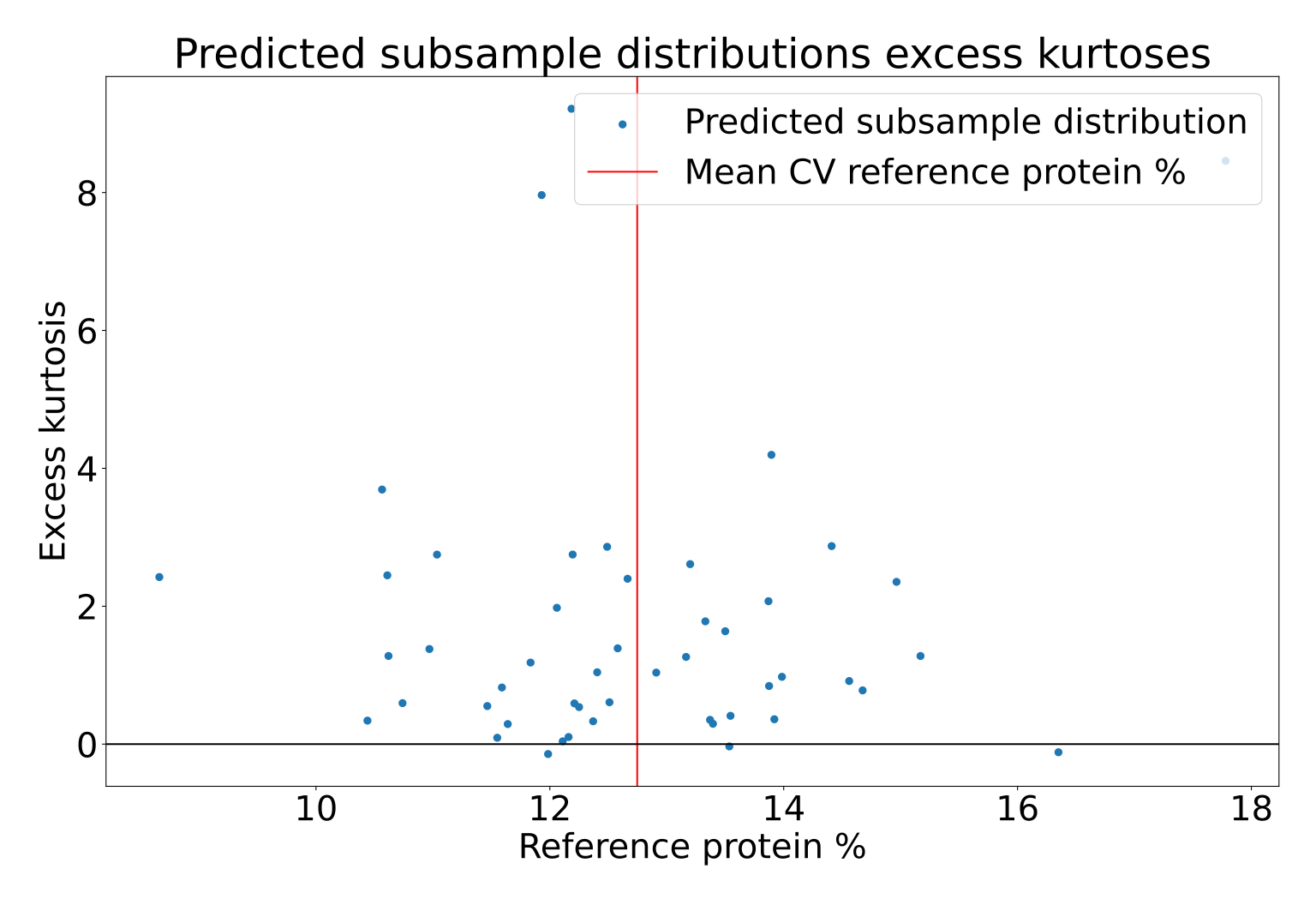}
        \caption{Training kurtoses.}
    \end{subfigure}
    \hfill
    \begin{subfigure}[b]{0.33\textwidth}
        \centering
        \includegraphics[width=\textwidth]{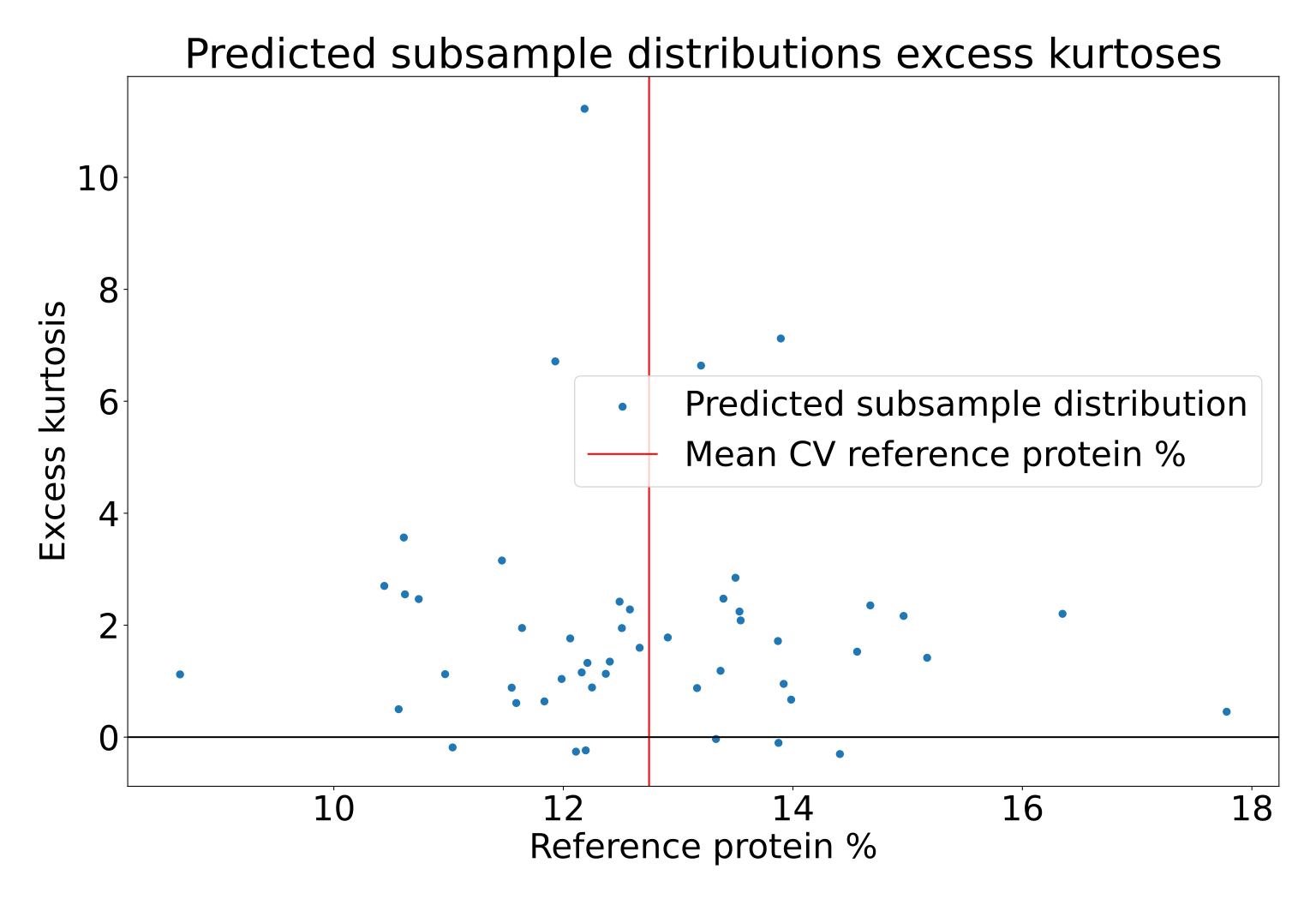}
        \caption{Validation kurtoses.}
    \end{subfigure}
    \begin{subfigure}[b]{0.33\textwidth}
        \centering
        \includegraphics[width=\textwidth]{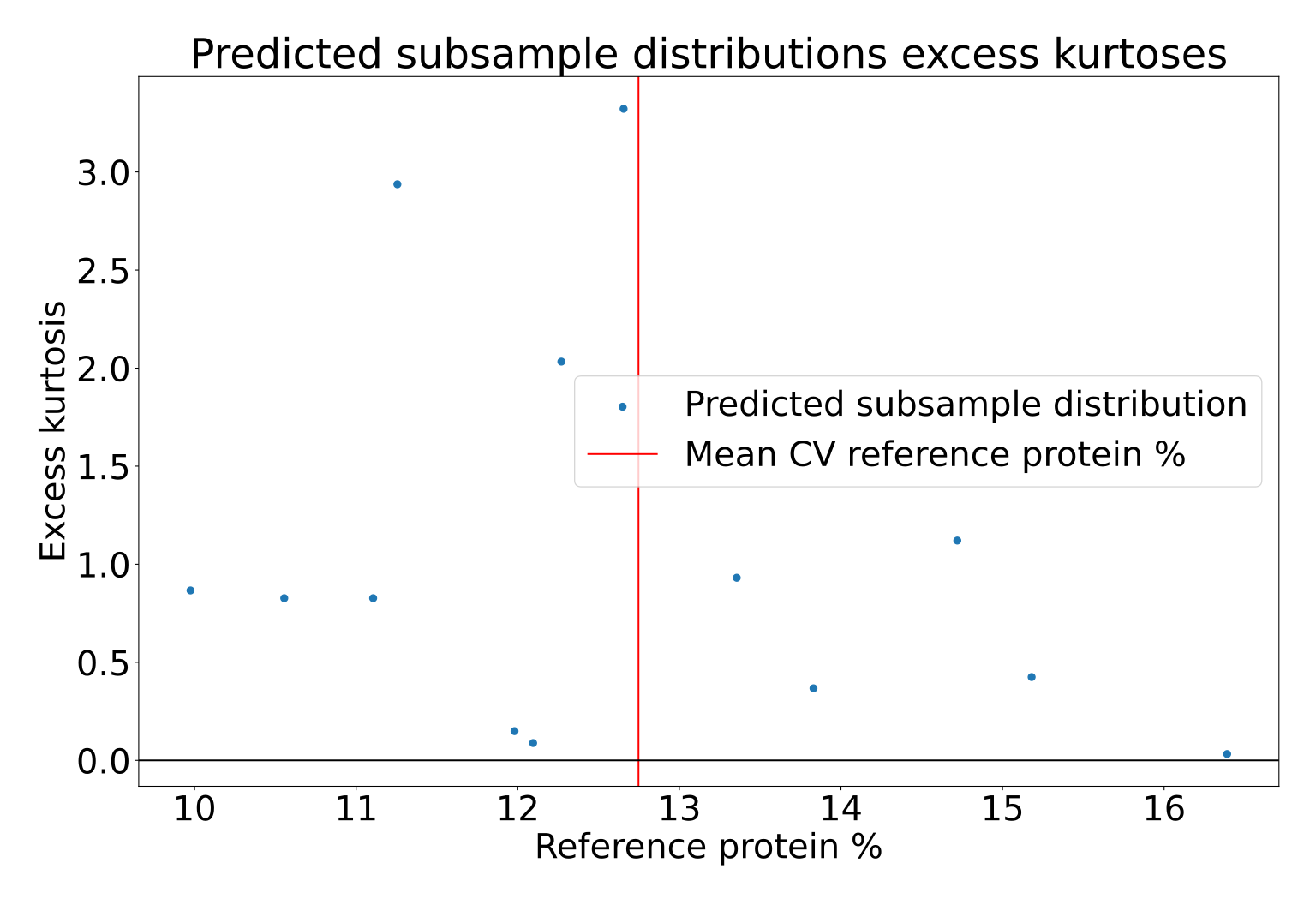}
        \caption{Test kurtoses.}
    \end{subfigure}
    \\
    \centering
    \begin{subfigure}[b]{0.33\textwidth}
        \centering
        \includegraphics[width=\textwidth]{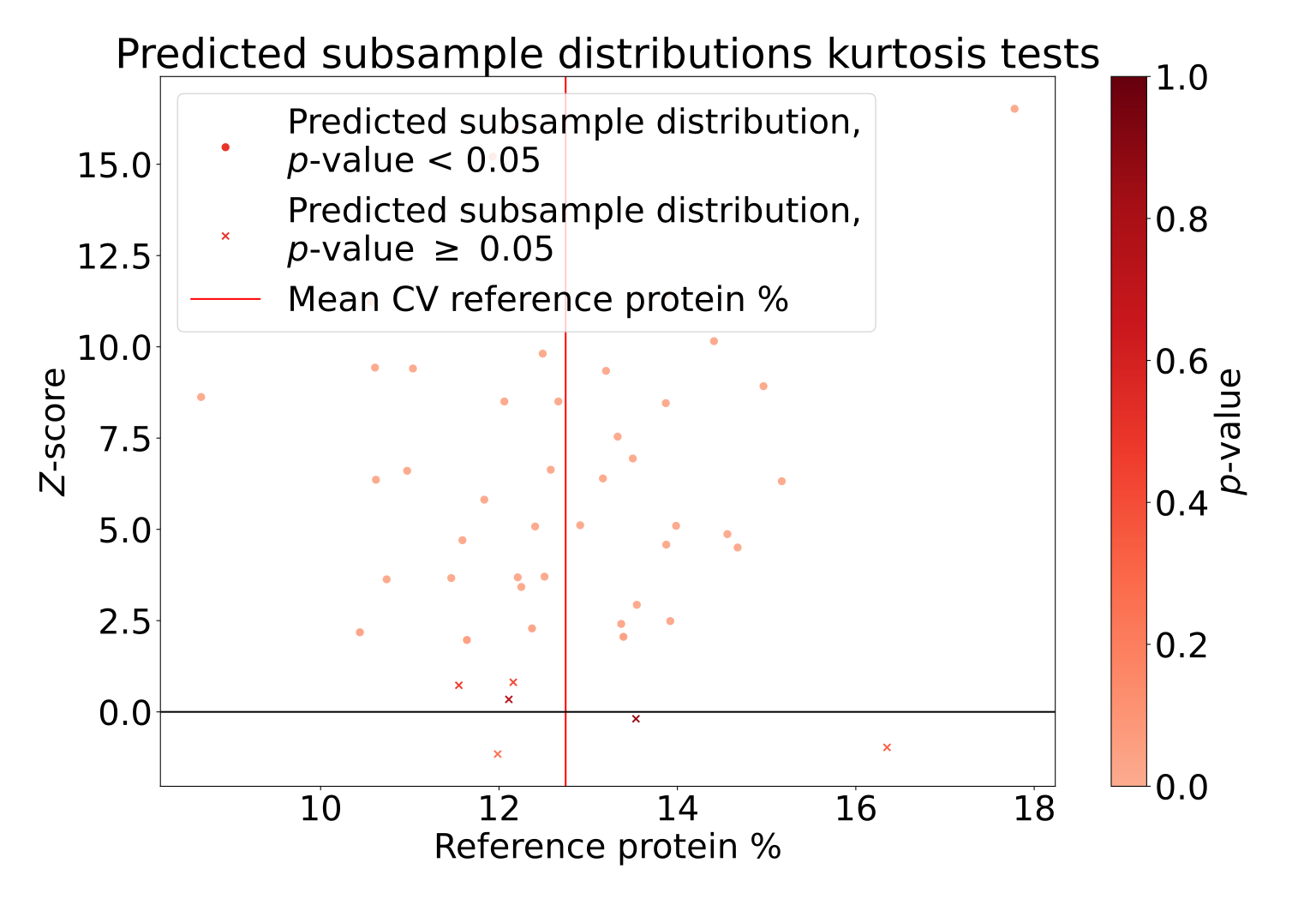}
        \caption{Training kurtosis $Z$-tests.}
    \end{subfigure}
    \hfill
    \begin{subfigure}[b]{0.33\textwidth}
        \centering
        \includegraphics[width=\textwidth]{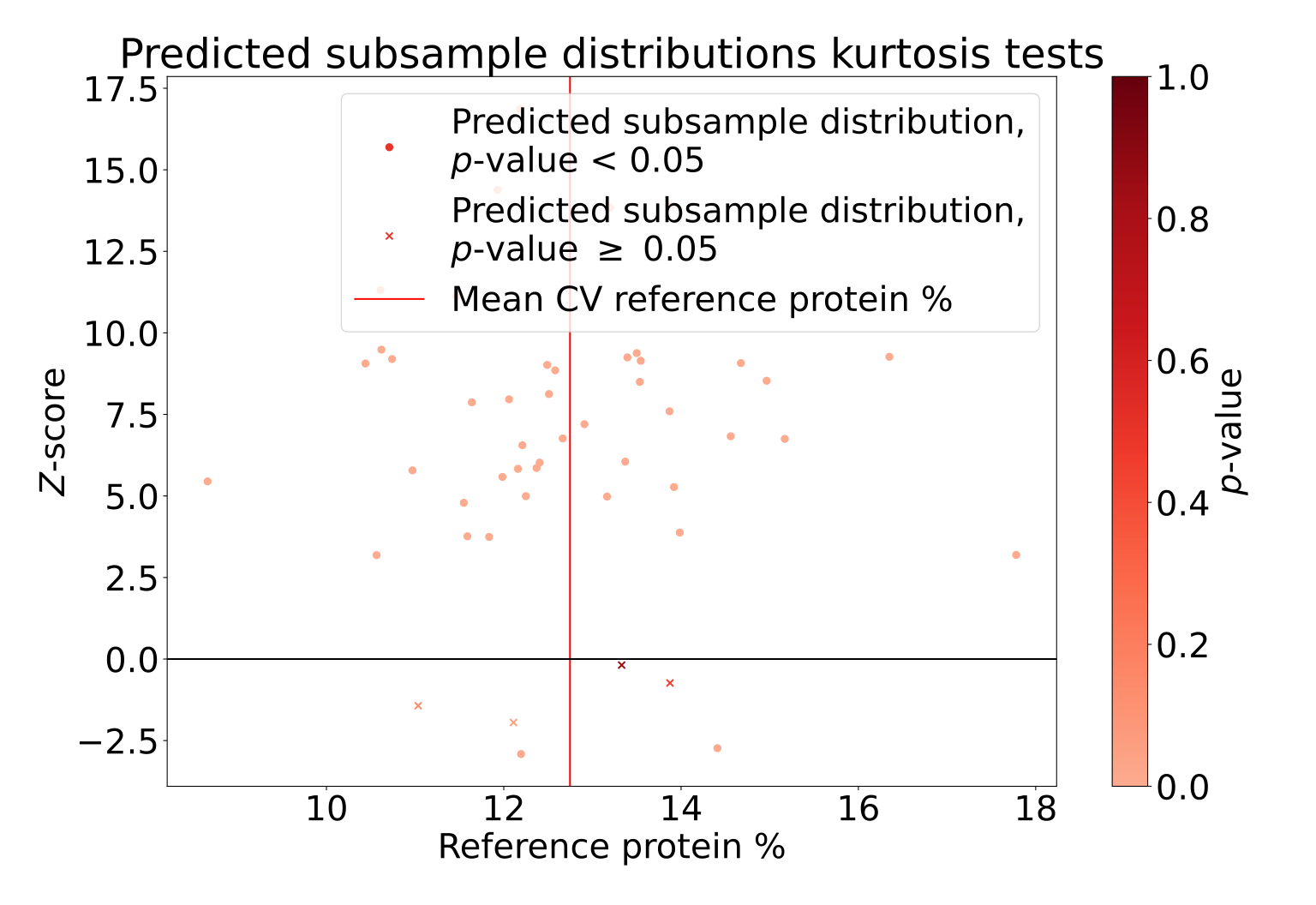}
        \caption{Validation kurtosis $Z$-tests.}
    \end{subfigure}
    \begin{subfigure}[b]{0.33\textwidth}
        \centering
        \includegraphics[width=\textwidth]{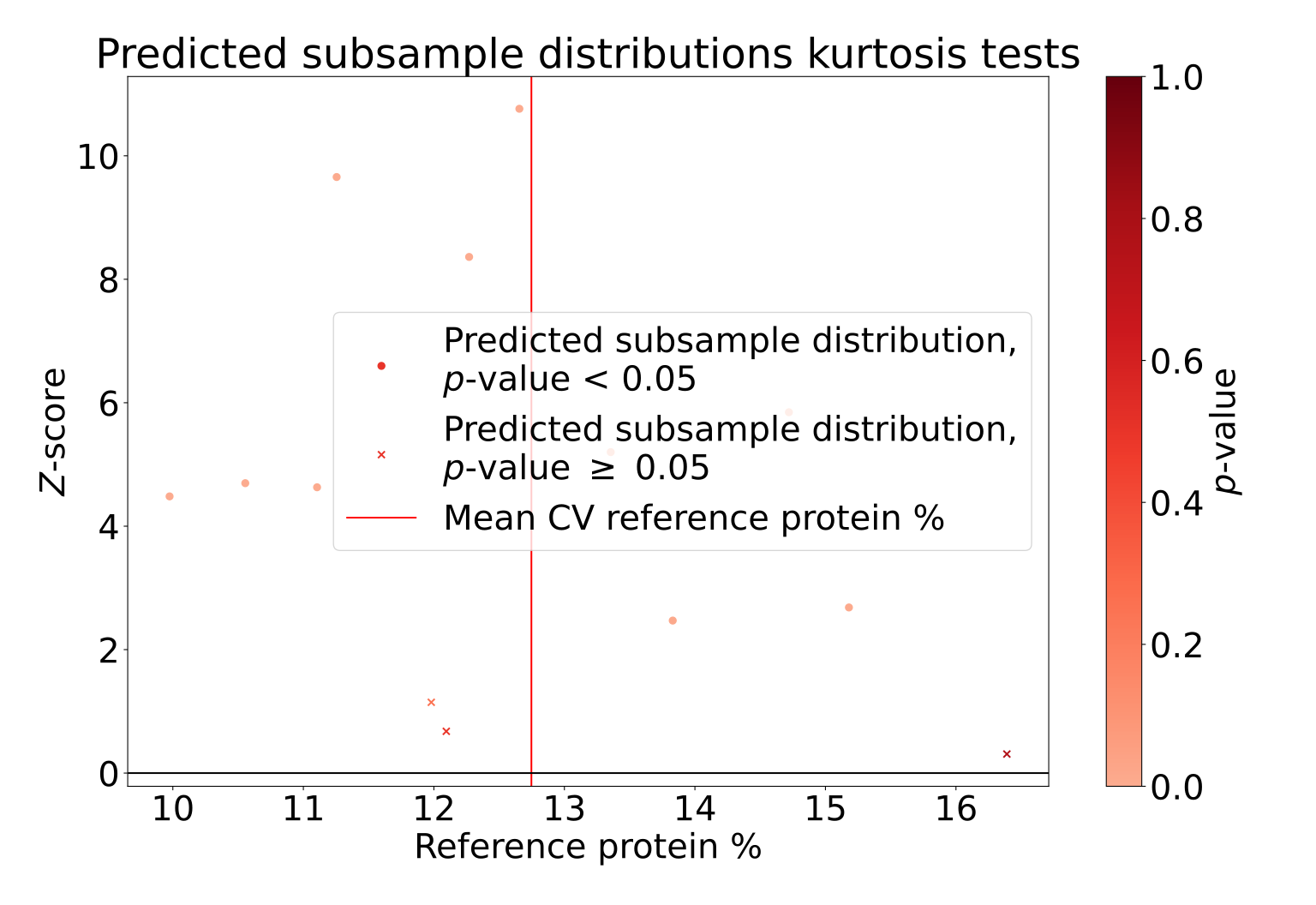}
        \caption{Test kurtosis $Z$-tests.}
    \end{subfigure}
    \caption{PLS-R$_{\text{bulk}}$ subsample prediction distribution kurtoses and kurtosis tests for each bulk sample as a function of the reference protein of the bulk sample.}
    \label{fig:prediction_excess_kurtosis_bulk_plsr}
\end{figure}

We would also like to inspect the excess kurtosis, which is $0$ for a normal distribution, is bounded below by $-2$, and is unbounded above. Kurtosis measures a distribution's tendency to produce outliers and not, as otherwise often stated, its peakedness \cite{westfall2014kurtosis}. Negative excess kurtosis implies a platykurtic distribution that produces fewer or less extreme outliers than a normal distribution. Likewise, a positive excess kurtosis implies a leptokurtic distribution that produces more or more extreme outliers than a normal distribution. In turn, a distribution with an excess kurtosis of $0$ is called mesokurtic. We show the excess kurtoses and associated kurtosis tests for Modified ResNet-18 Regressor, PLS-R, and PLS-R$_{\text{bulk}}$ in \myfigref{fig:prediction_excess_kurtosis_resnet}, \myfigref{fig:prediction_excess_kurtosis_plsr}, and \myfigref{fig:prediction_excess_kurtosis_bulk_plsr}, respectively. These tests imply that, beyond any reasonable doubt, most of the prediction distributions are leptokurtic and produce too many or extreme outliers than would be expected from a normal distribution.

\begin{figure}[htbp]
    \centering
    \begin{subfigure}[b]{0.33\textwidth}
        \centering
        \includegraphics[width=\textwidth]{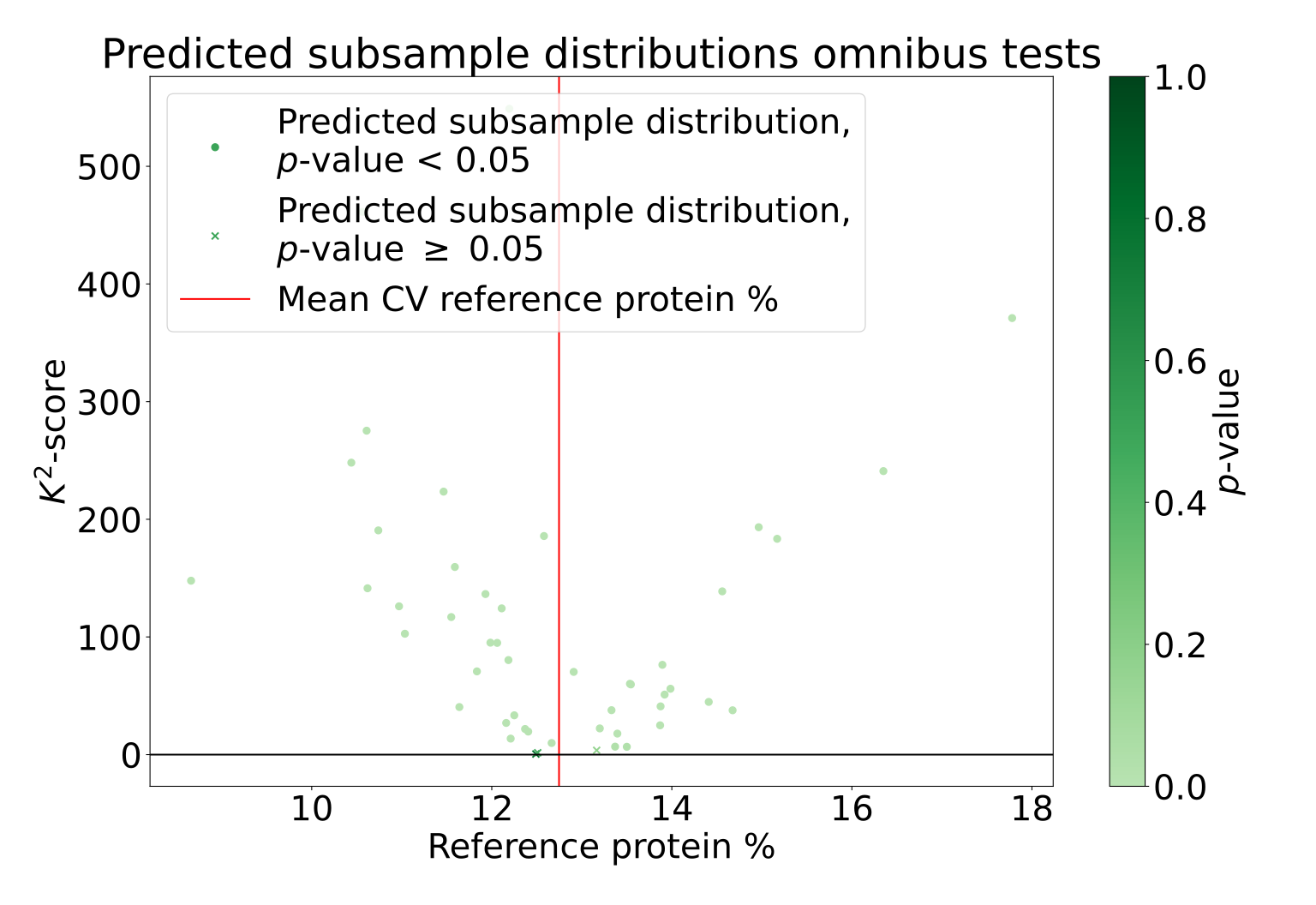}
        \caption{Modified ResNet-18 Regressor training omnibus $K^2$-test.}
    \end{subfigure}
    \hfill
    \begin{subfigure}[b]{0.33\textwidth}
        \centering
        \includegraphics[width=\textwidth]{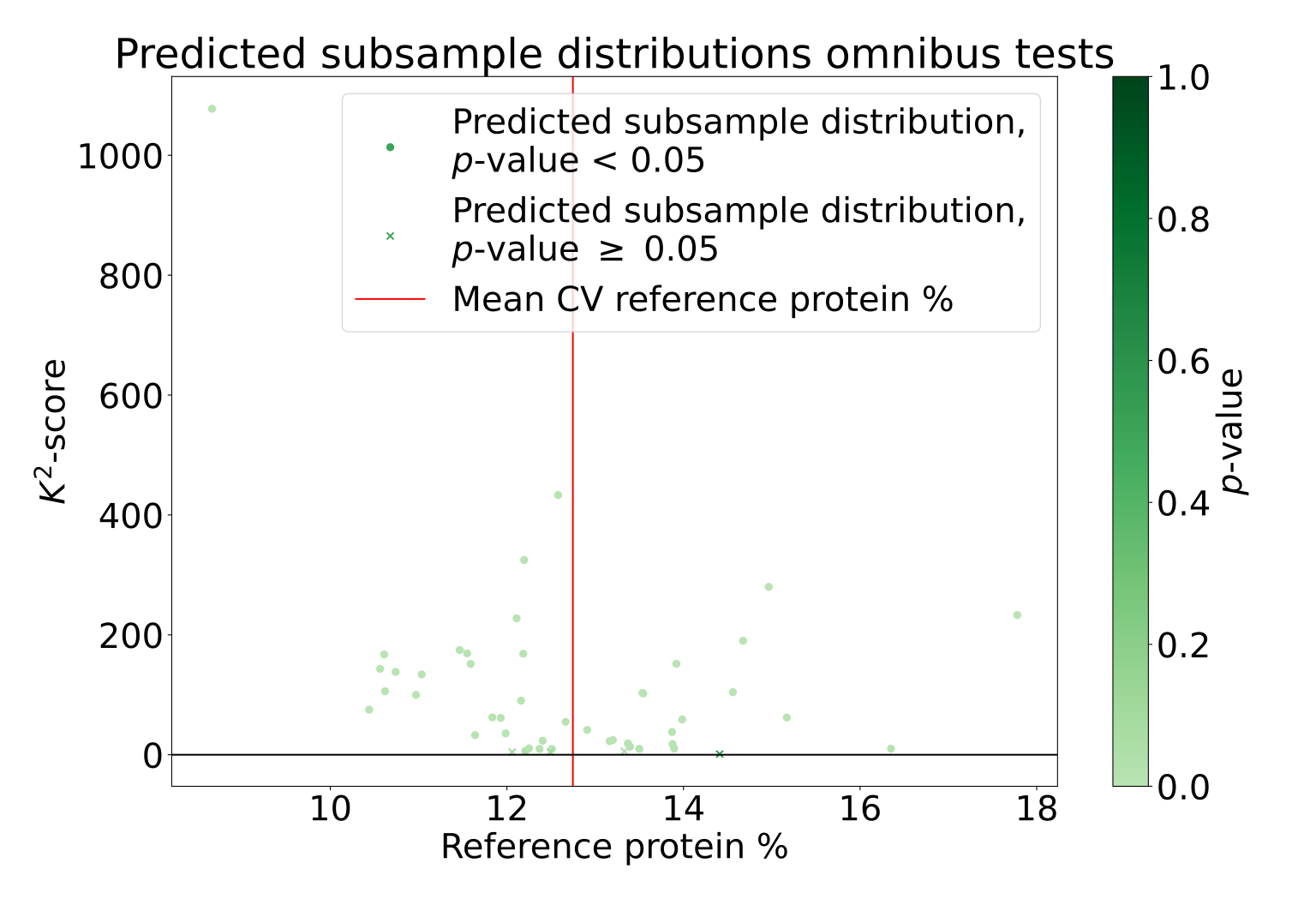}
        \caption{Modified ResNet-18 Regressor validation omnibus $K^2$-test.}
    \end{subfigure}
    \begin{subfigure}[b]{0.33\textwidth}
        \centering
        \includegraphics[width=\textwidth]{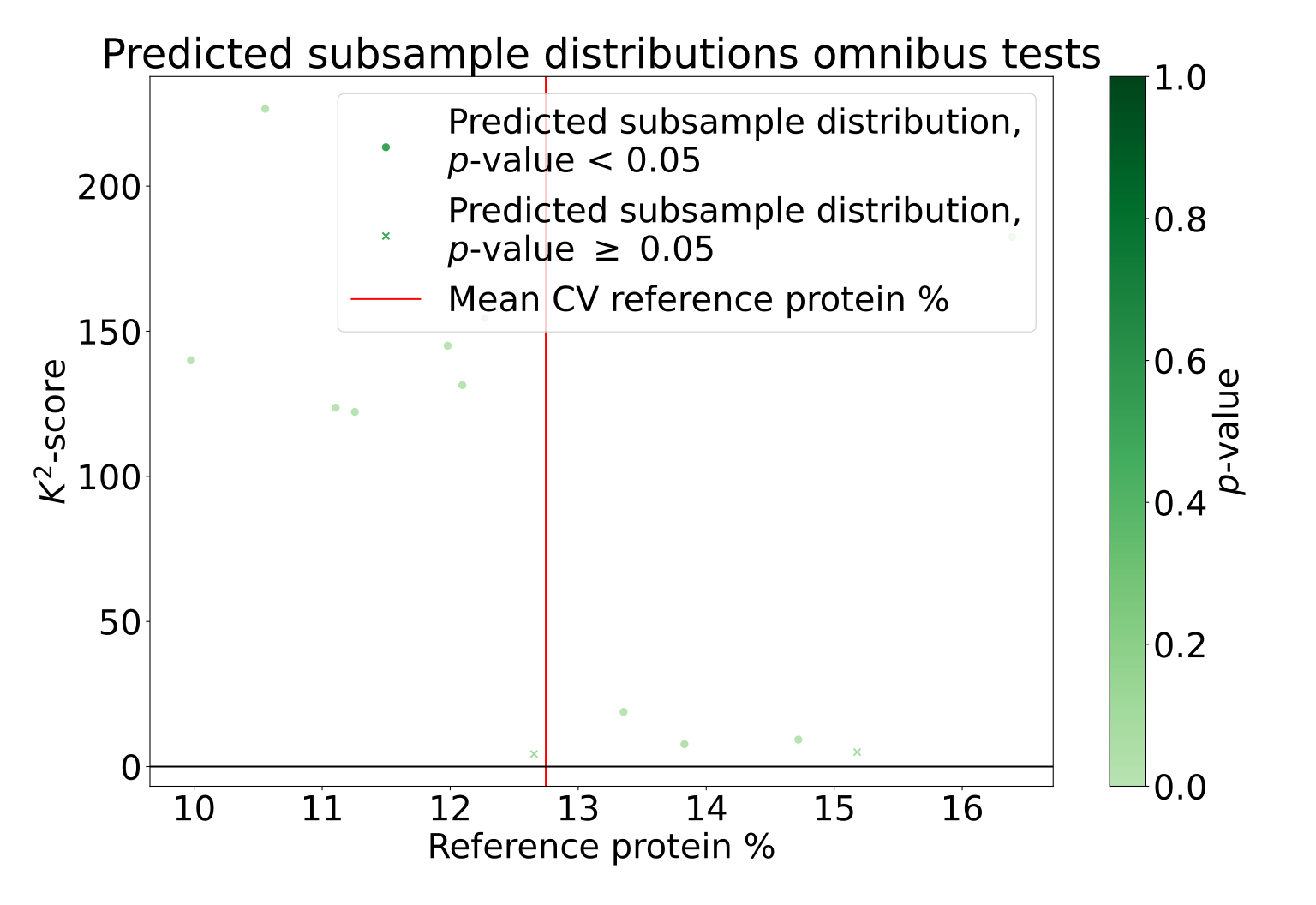}
        \caption{Modified ResNet-18 Regressor test omnibus $K^2$-test.}
    \end{subfigure}
    \\
    \centering
    \begin{subfigure}[b]{0.33\textwidth}
        \centering
        \includegraphics[width=\textwidth]{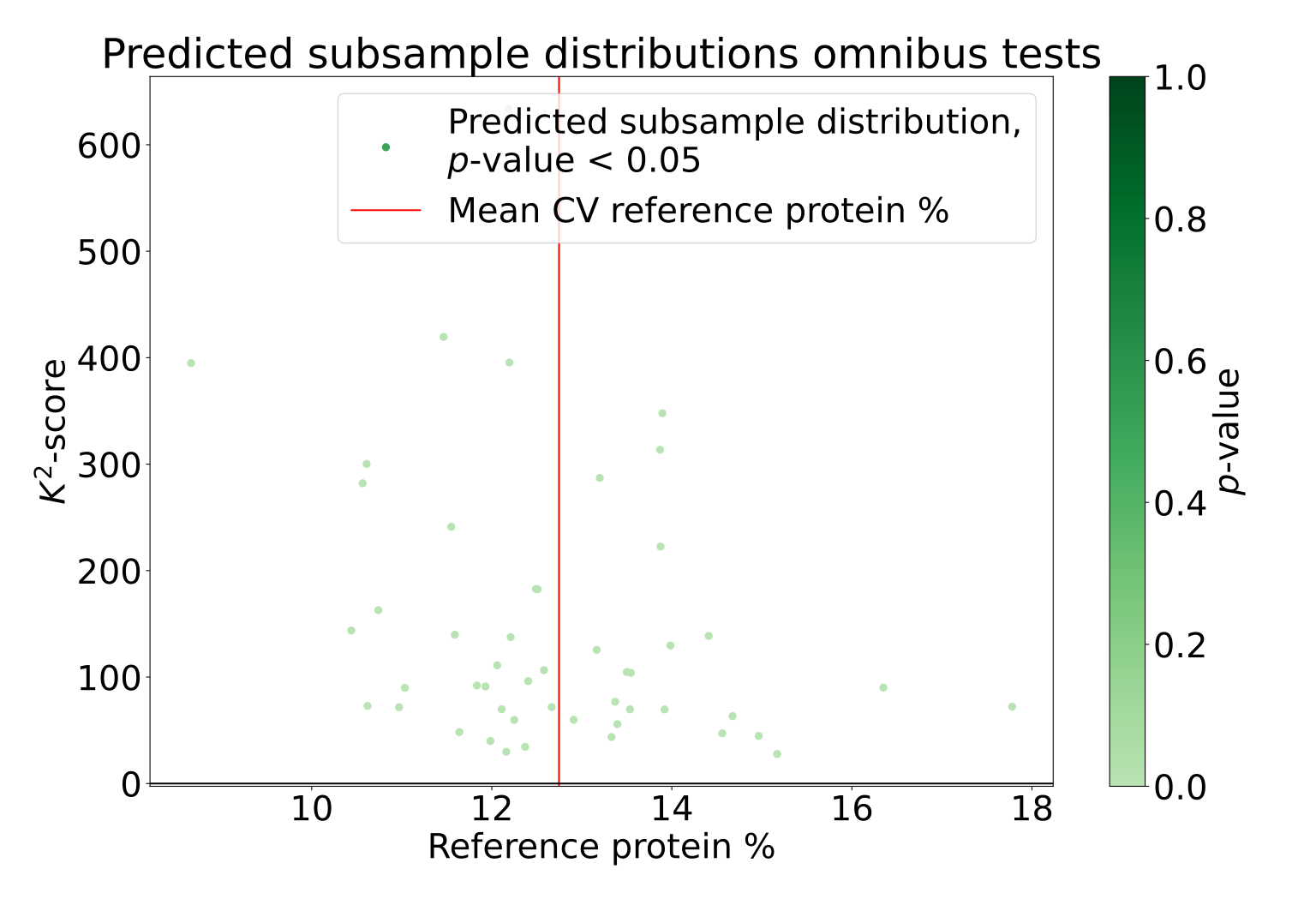}
        \caption{PLS-R training omnibus $K^2$-test.}
    \end{subfigure}
    \hfill
    \begin{subfigure}[b]{0.33\textwidth}
        \centering
        \includegraphics[width=\textwidth]{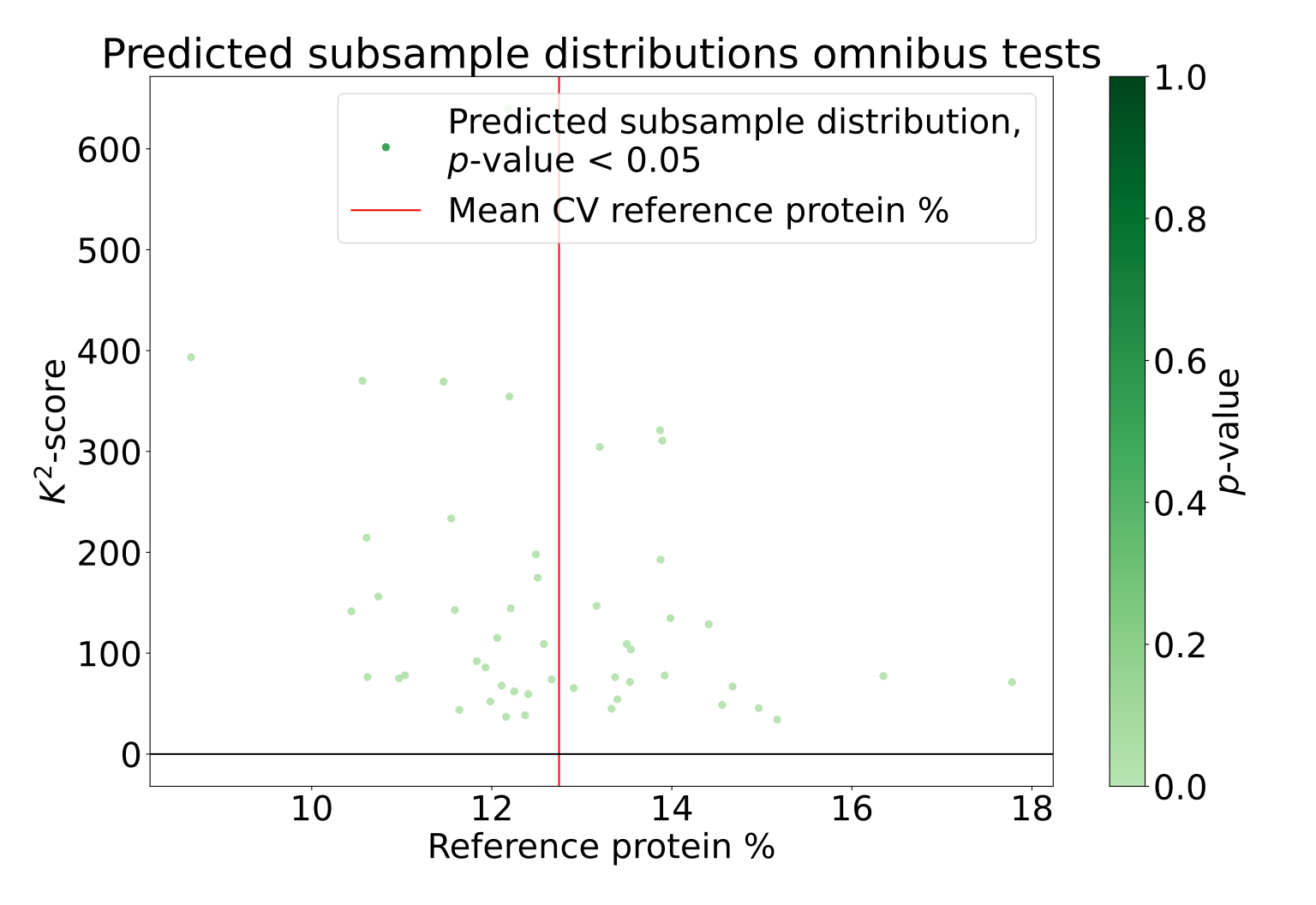}
        \caption{PLS-R validation omnibus $K^2$-test.}
    \end{subfigure}
    \begin{subfigure}[b]{0.33\textwidth}
        \centering
        \includegraphics[width=\textwidth]{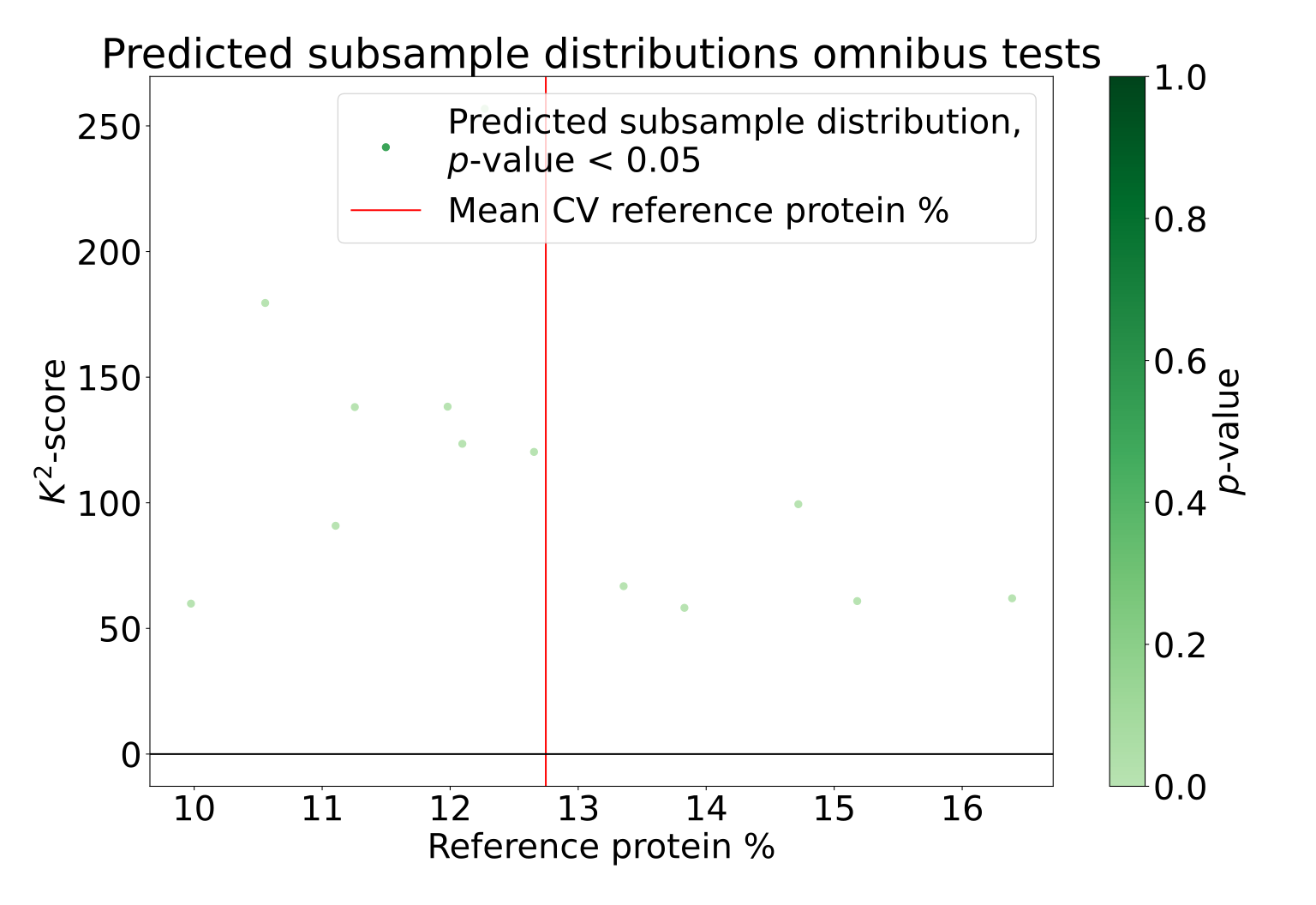}
        \caption{PLS-R test omnibus $K^2$-test.}
    \end{subfigure}
    \\
    \centering
    \begin{subfigure}[b]{0.33\textwidth}
        \centering
        \includegraphics[width=\textwidth]{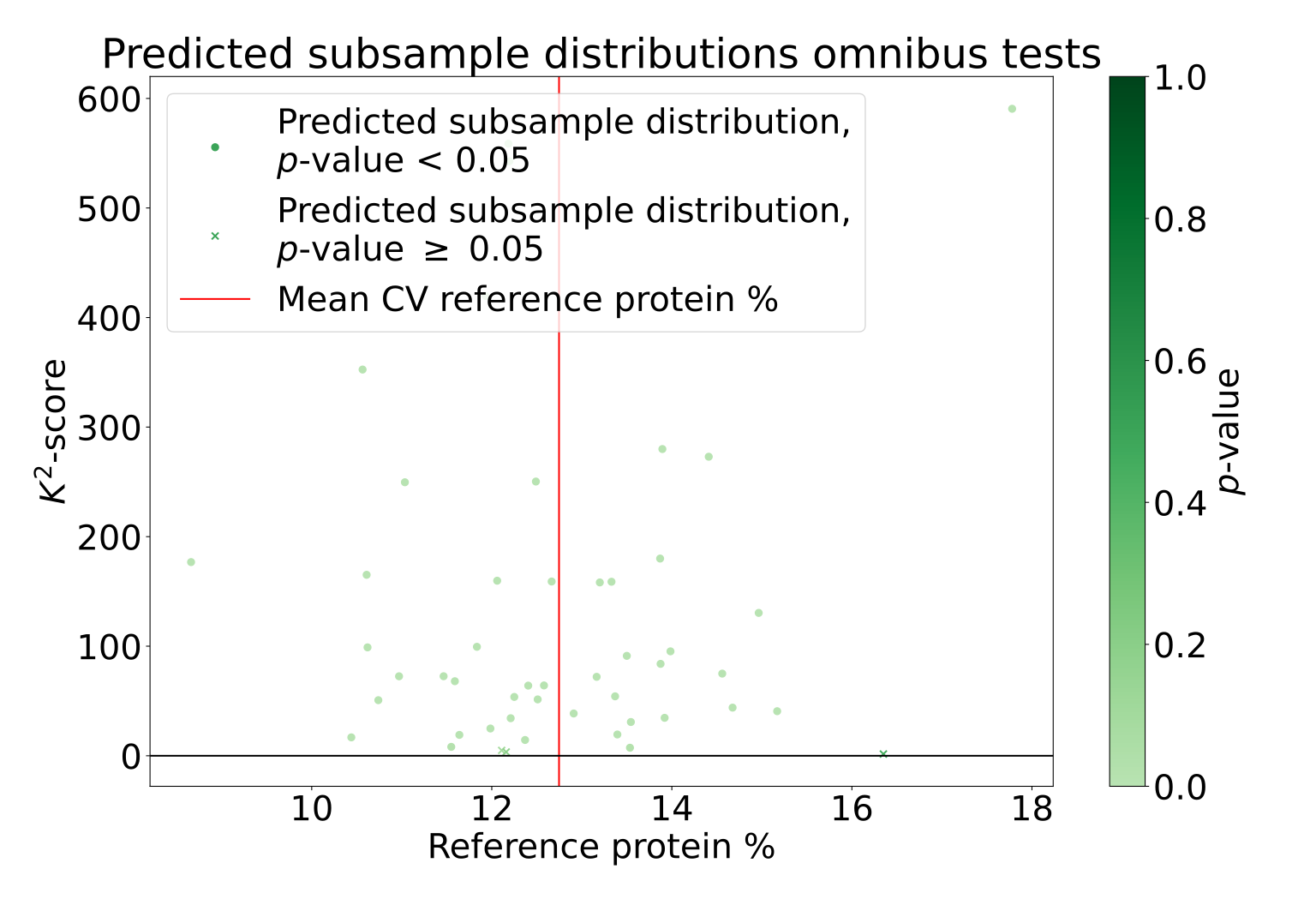}
        \caption{PLS-R$_{\text{bulk}}$ training omnibus $K^2$-test.}
    \end{subfigure}
    \hfill
    \begin{subfigure}[b]{0.33\textwidth}
        \centering
        \includegraphics[width=\textwidth]{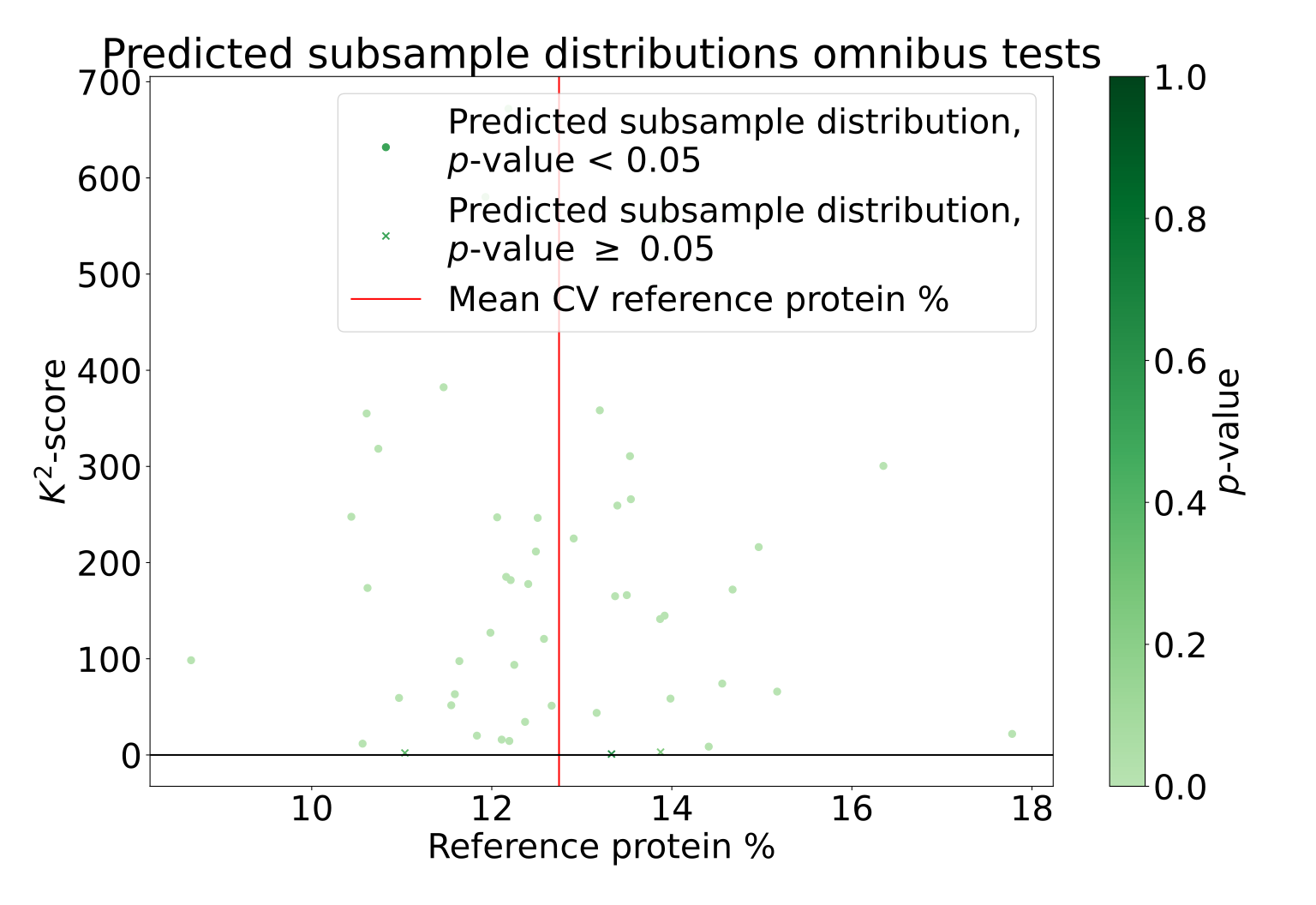}
        \caption{PLS-R$_{\text{bulk}}$ validation omnibus $K^2$-test.}
    \end{subfigure}
    \begin{subfigure}[b]{0.33\textwidth}
        \centering
        \includegraphics[width=\textwidth]{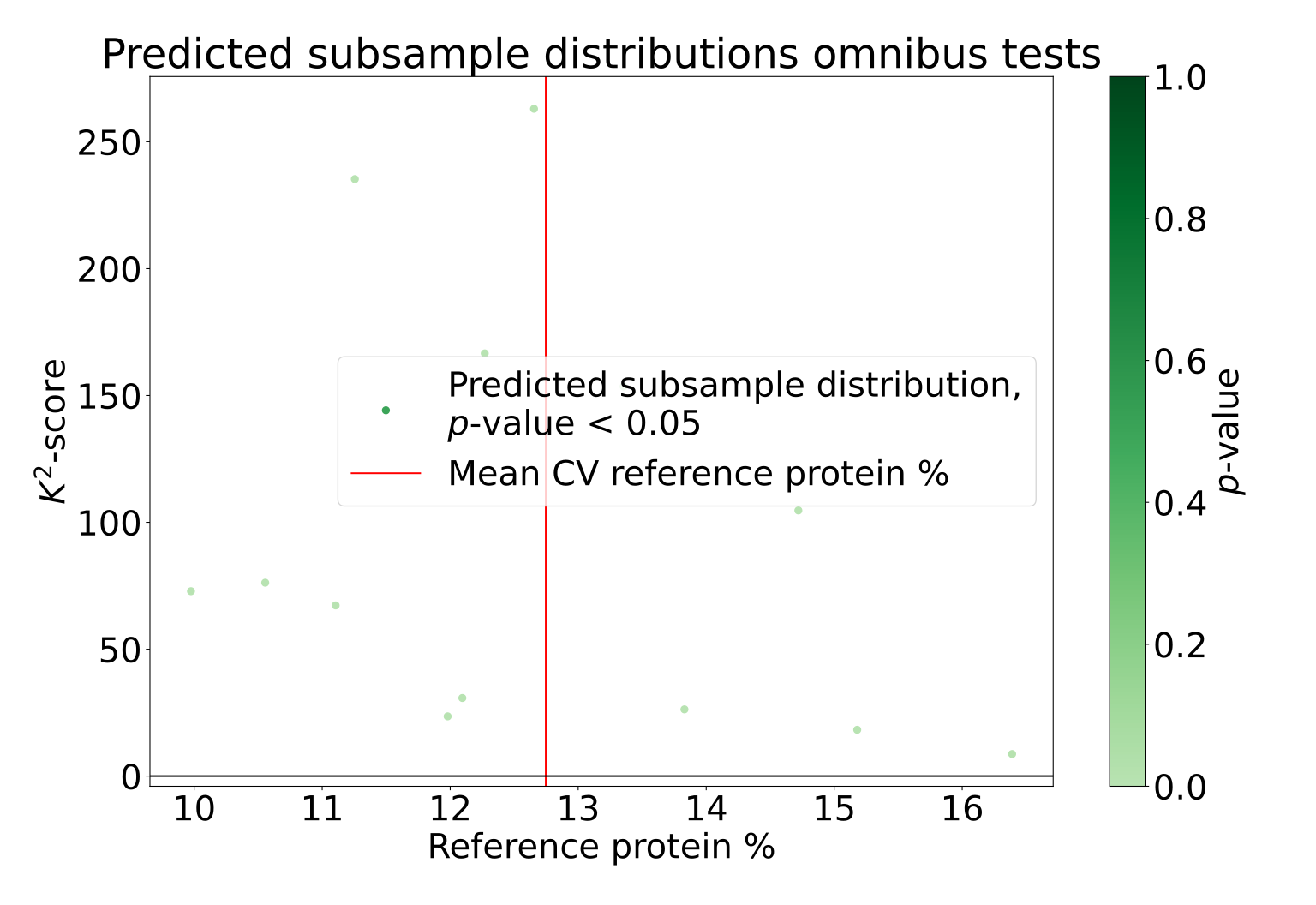}
        \caption{PLS-R$_{\text{bulk}}$ test omnibus $K^2$-test.}
    \end{subfigure}
    \caption{Modified ResNet-18 Regressor, PLS-R, and PLS-R$_{\text{bulk}}$ subsample prediction distribution omnibus tests.}
    \label{fig:prediction_omnibus_tests}
\end{figure}

Finally, we show the omnibus tests for normality for all three models in \myfigref{fig:prediction_omnibus_tests}. Here, it is evident that most of the prediction distributions for all the models are, in fact, not normal. The non-zero skewnesses combined with the high excess kurtoses imply that the distributions contain more or more extreme outliers than the normal distribution and that these outliers tend to show up on one side of the means of the distributions. Recall that RMSE is a measure that weights outliers heavily due to the square term in the deviation from the mean. In contrast, the arithmetic mean only weights outliers linearly. Thus, a model that is calibrated to minimize RMSE$(\mathbf{y}_{\text{ss}}, \hat{\mathbf{y}}_{\text{ss}})$, where outliers are weighted heavily, will yield a different set of model parameters than those achieved by minimizing RMSE$(\mathbf{y}_{\text{bm}}, \hat{\mathbf{y}}_{\text{bm}})$, where outliers are weighted less heavily.

While our analysis reveals the skewed and leptokurtic distributions that give rise to the discrepancy between the two minimization problems, the underlying reasons for the shapes of these distributions remain unknown. We note that, especially for Modified ResNet-18 Regressor, the distributions seem to be skewed towards a protein \% that is slightly higher than the CV reference mean value, and the skewness gets more extreme as the bulk sample mean reference value deviates further from the CV reference mean value. Additionally, for Modified ResNet-18 Regressor, the training set kurtoses are lower for bulk samples that have a reference protein \% close to the mean CV reference protein. These effects could indicate that the Modified ResNet-18 Regressor has learned to recognize grain from the different bulk samples in the training set and map them close to the reference value, thus circumventing the features that explain protein content, severely overfitting, and rendering ineffective the strategy that assigns the bulk sample mean reference values to bulk subsamples.

\section{The influence of grain density on model performances}\label{sec:grain_density}
In this section, we investigate the effect of grain density on protein content regression on Dataset \#1 and grain variety classification on Dataset \#2. Recall that the grain density of an image crop is the ratio of its amount of pixels containing grain to its total amount of pixels. We do not analyze the predictions of the ensemble models on Dataset \#1. Instead, we analyze the ensembles' constituents individually. In particular, we are interested in measuring the average performance of each constituent as a function of grain density. This allows us to compute a mean RMSE$(\mathbf{y}_{\text{ss,test}}, \hat{\mathbf{y}}_{\text{ss,test}})$ and an associated SEM. Engstrøm \etal~\cite{engstrom2023improving} use a grain density of at least $0.1$ for all their model calibrations and evaluations on both datasets.

\begin{figure}[htbp]
    \centering
    \begin{subfigure}[b]{0.48\textwidth}
        \centering
        \includegraphics[width=\textwidth]{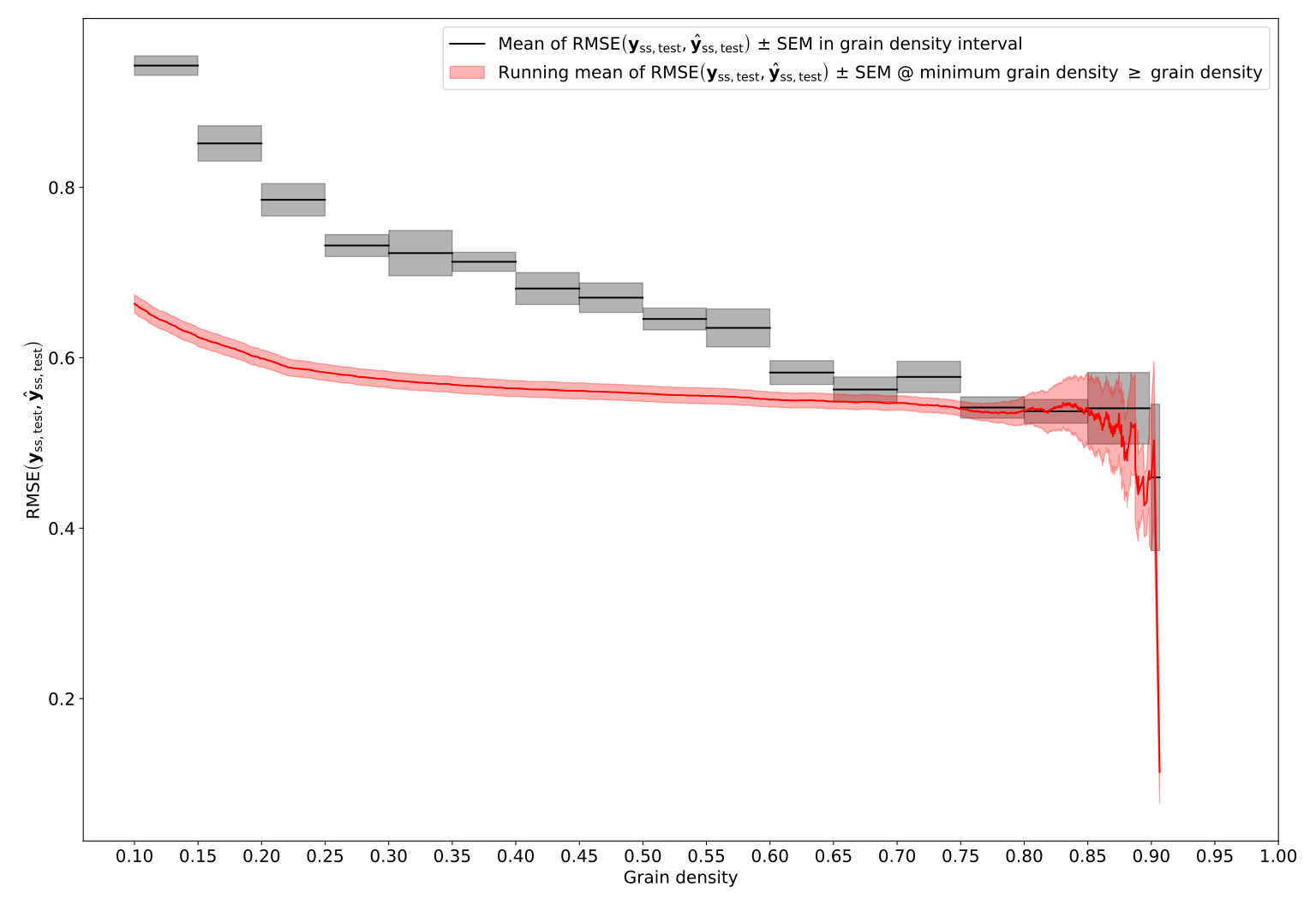}
        \caption{Modified ResNet-18 Regressor.}
    \end{subfigure}
    \hfill
    \begin{subfigure}[b]{0.48\textwidth}
        \centering
        \includegraphics[width=\textwidth]{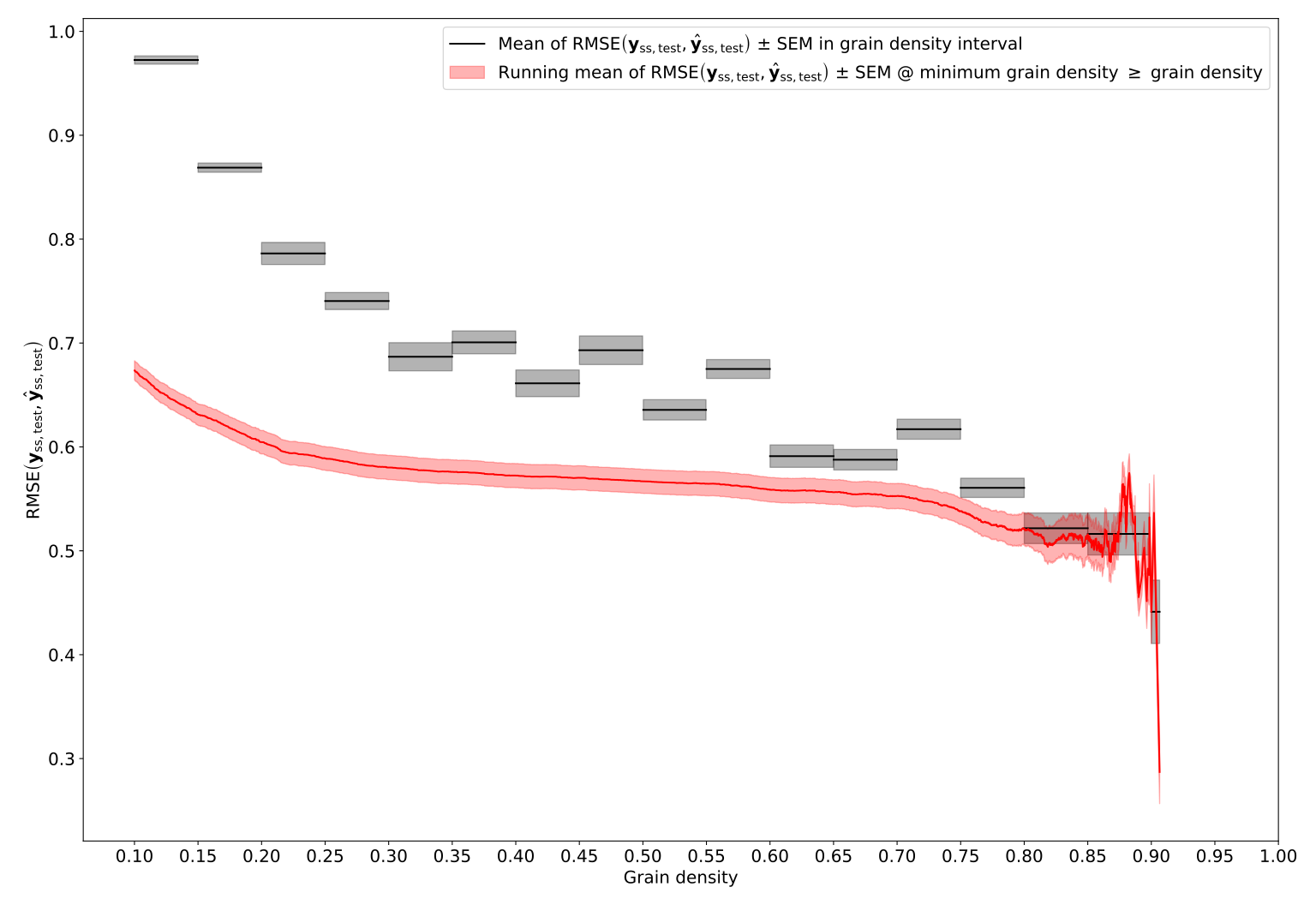}
        \caption{PLS-R.}
    \end{subfigure}
    \\
    \begin{subfigure}[b]{0.48\textwidth}
        \centering
        \includegraphics[width=\textwidth]{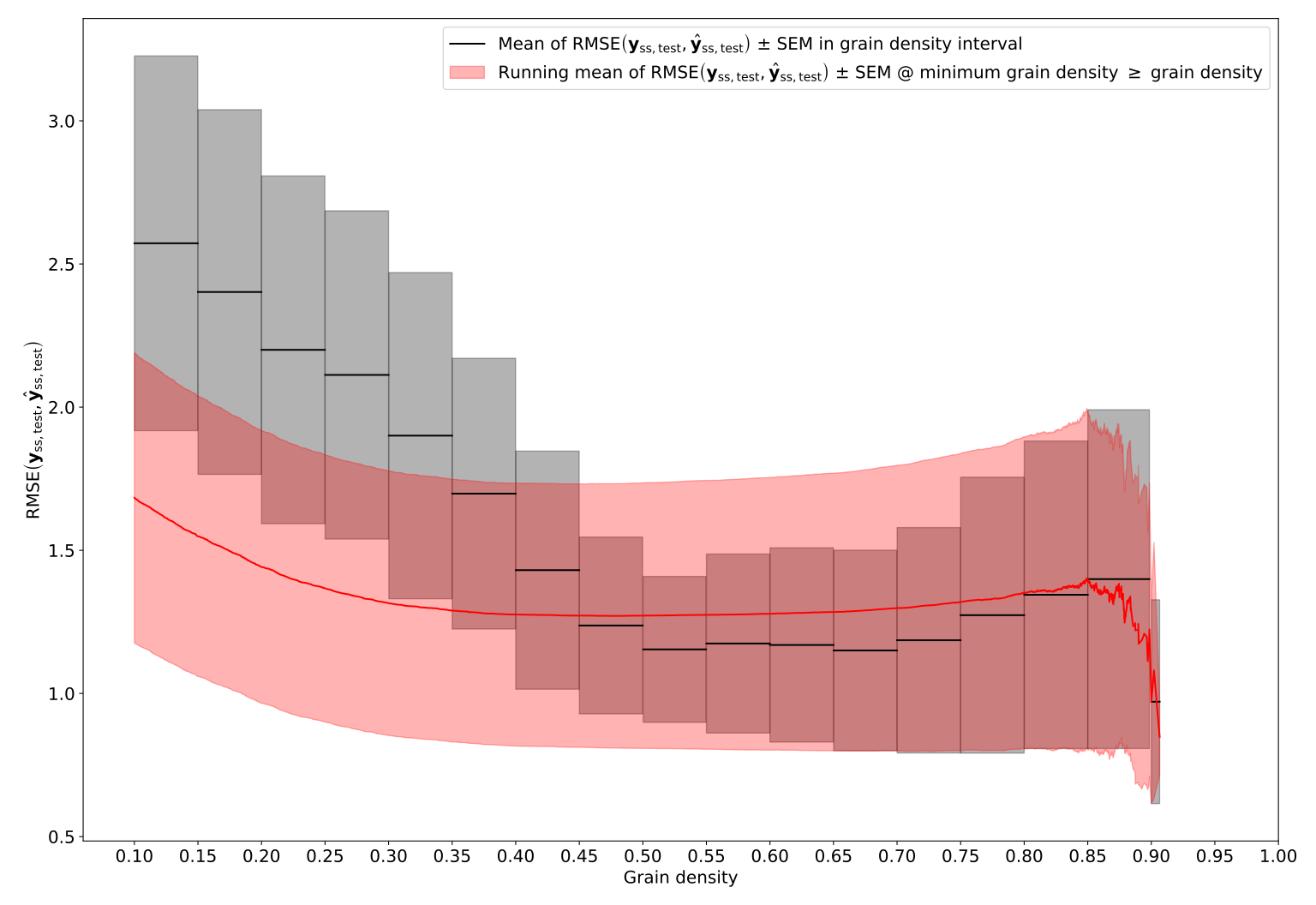}
        \caption{PLS-R$_{\text{bulk}}$}
    \end{subfigure}
    \caption{RMSE$(\mathbf{y}_{\text{ss,test}}, \hat{\mathbf{y}}_{\text{ss,test}})$ as a function of the grain density. The red graphs should be interpreted right-to-left, i.e., for each of the three regression model types at any grain density, they represent the mean RMSE and associated SEM achieved by the five ensemble model constituents by predicting test set subsamples with grain density at least that high.}
    \label{fig:rmse_at_density}
\end{figure}

In \myfigref{fig:rmse_at_density} we show the RMSE$(\mathbf{y}_{\text{ss,test}}, \hat{\mathbf{y}}_{\text{ss,test}})$ for grain density intervals of $0.05$ and for any given value of minimum grain density. Here, it is clear that for all models, RMSE$(\mathbf{y}_{\text{ss,test}}, \hat{\mathbf{y}}_{\text{ss,test}})$ deterioates as grain density decreases. This is unsurprising as the reference value, computed as a mean over a bulk sample, is expected to be more accurate as grain density increases, and, likewise, any reasonably well-calibrated model will reflect this relationship in its predictions.

\begin{figure}[htbp]
    \centering
    \begin{subfigure}[b]{0.48\textwidth}
        \centering
        \includegraphics[width=\textwidth]{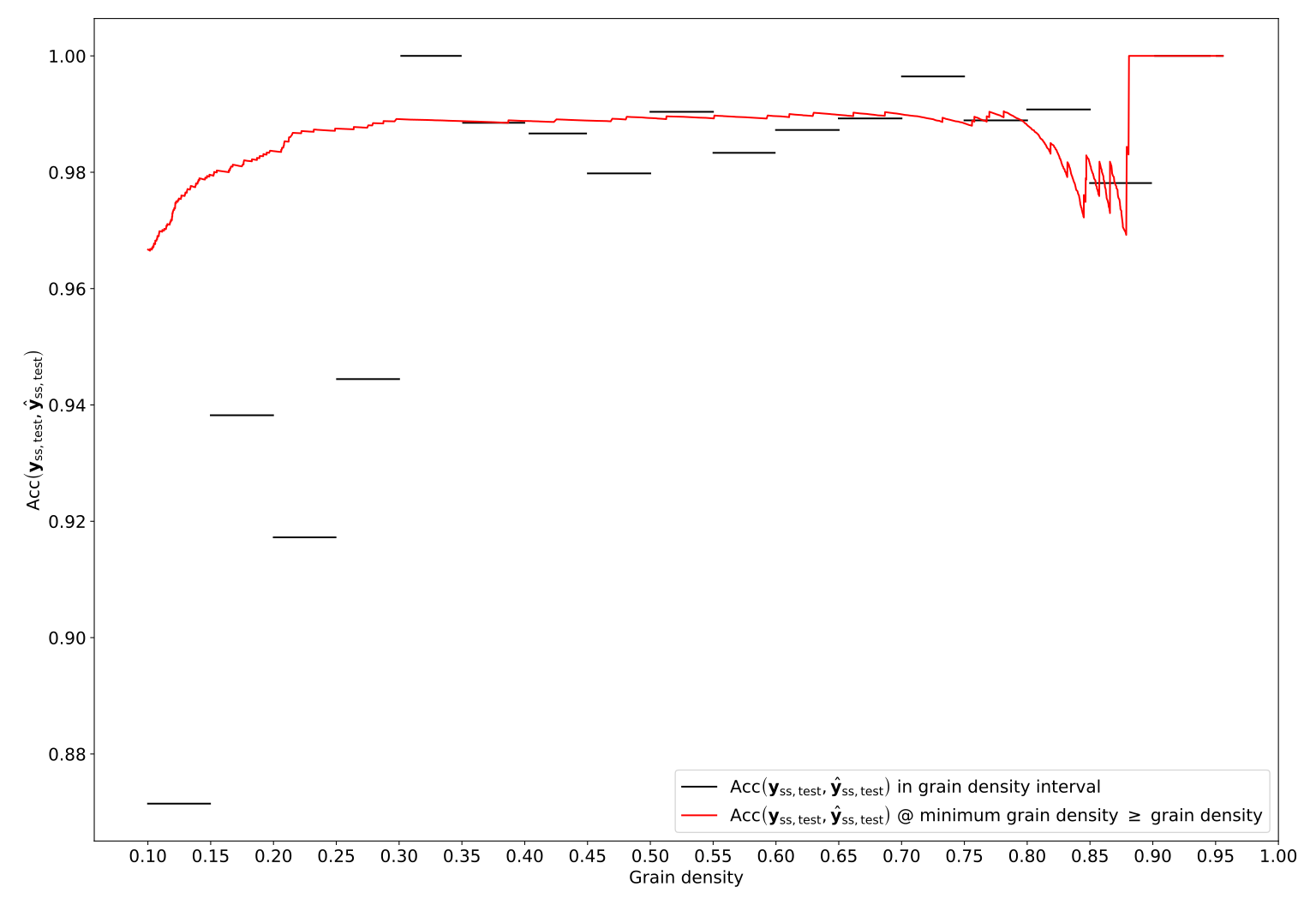}
        \caption{Modified ResNet-18 Classifier.}
    \end{subfigure}
    \hfill
    \begin{subfigure}[b]{0.48\textwidth}
        \centering
        \includegraphics[width=\textwidth]{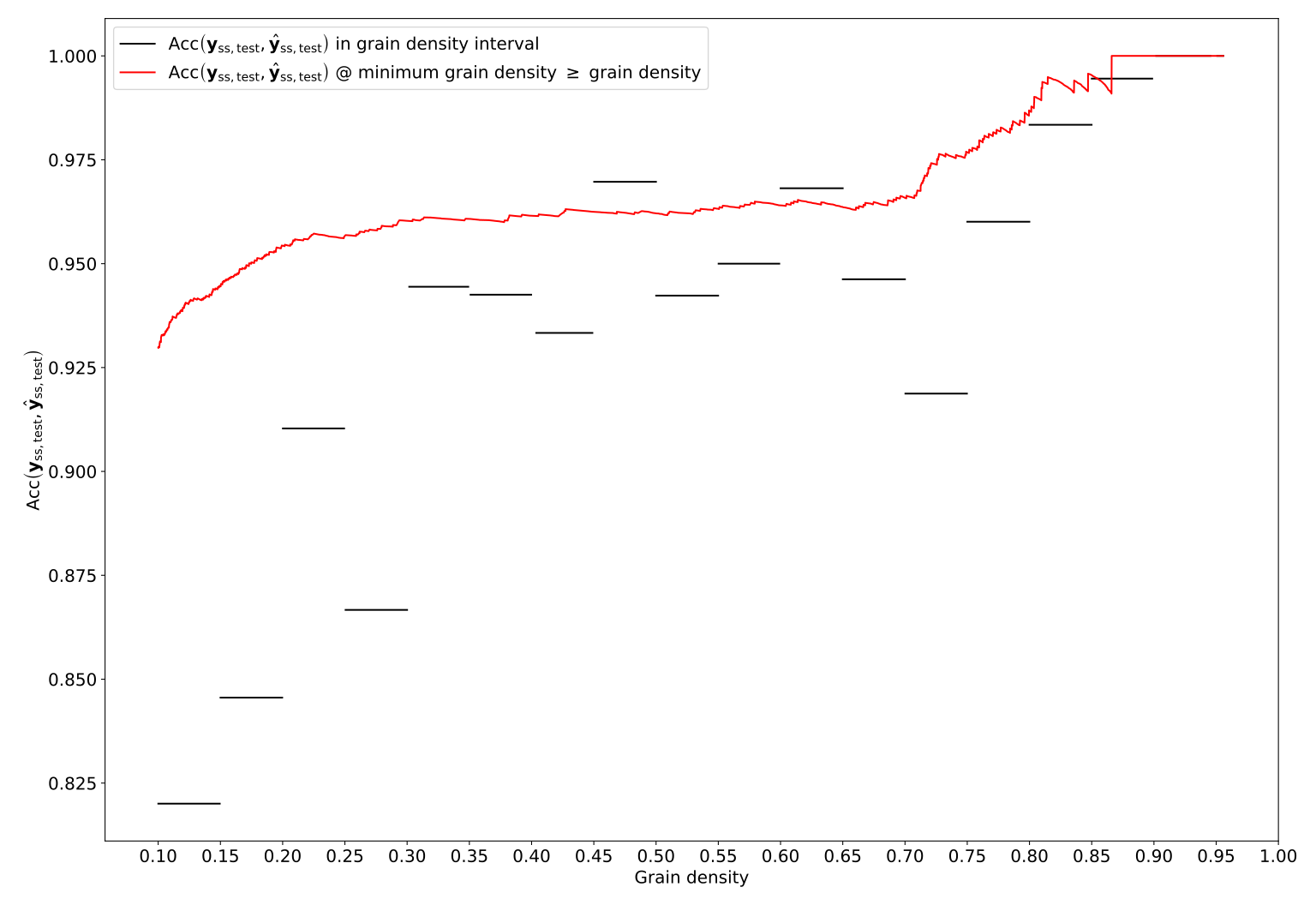}
        \caption{PLS2-DA.}
    \end{subfigure}
    \caption{Acc$(\mathbf{y}_{\text{ss,test}}, \hat{\mathbf{y}}_{\text{ss,test}})$ as a function of the grain density. The red graphs should be interpreted right-to-left, i.e., for both of the classification model types at any grain density, they represent the mean RMSE and associated SEM achieved by the five ensemble model constituents by predicting test set subsamples with grain density at least that high.}
    \label{fig:acc_at_density}
\end{figure}

On Dataset \#2, Dreier \etal~\cite{dreier2022hyperspectral} use a grain density of at least $0.5$ for all their calibrations but also do a test set evaluation for grain density $<0.5$ and show that accuracy rapidly diminishes as grain density decreases. Additionally, on Dataset \#2, there are no ensemble models as the models were trained with a single training and validation dataset split. In \myfigref{fig:acc_at_density} we show the Acc$(\mathbf{y}_{\text{ss,test}}, \hat{\mathbf{y}}_{\text{ss,test}})$, defined in \myeqnref{eq:acc}, for Modified ResNet-18 Classifier and PLS2-DA. Recall that for Dataset \#2, unlike for Dataset \#1, the reference value is equally accurate for a low and a high grain density as only a single grain variety is present in any given subsample. While Modified ResNet-18 Classifier is better than PLS2-DA, both models have a relatively stable performance until grain density decreases below $0.3$. At the interval with the lowest grain density of $0.1$ to $0.15$, Modified ResNet-18 Classifier reaches its minimum accuracy of $\sim 0.87$, and PLS2-DA reaches its minimum of $\sim 0.82$. Interestingly, however, both test accuracies are much higher than the Acc$(\mathbf{y}_{\text{ss,val}}, \hat{\mathbf{y}}_{\text{ss,val}})$ of $\sim 0.5$ achieved by Dreier \etal~\cite{dreier2022hyperspectral} on the validation split with their models that were trained on a grain density $\geq0.5$. These results reveal that while easier with a high grain density, getting accurate predictions on a low grain density subsample is feasible as long as the model has been calibrated on low grain density subsamples. Our results also show that both the spatio-spectral CNN and the purely spectral PLS2-DA benefit from a larger grain density.

\begin{equation}\label{eq:acc}
    \text{Acc}(\mathbf{y}, \hat{\mathbf{y}}) = \frac{1}{n}\sum_{i=0}^{n-1} \mathbb{1}\left(\mathbf{y}_{i}=\hat{\mathbf{y}}_{i}\right)
\end{equation}

\pagebreak

\section{Conclusion}\label{sec:conclusion}
This report shows that calibrating CNNs and PLS models to predict protein content in grain bulk subsamples using mean references from grain bulk samples can lead to biased average predictions. We have investigated the models' prediction distributions of the protein content in the grain bulk subsamples and found that they are both leptokurtic and skewed, implying that they produce more or more extreme outliers than a normal distribution and that these outliers typically occur on one side of the mean value. The underlying reason for the existence of the skewed leptokurtic distribution remains unknown. However, their existence implies that minimizing the outlier-sensitive RMSE on grain bulk subsamples leads to incorrect mean predictions on the grain bulk samples. We have devised two strategies to mitigate this issue. The first strategy works for both CNNs and PLS models and consists of a linear correction of the mean predictions on the grain bulk samples. The second strategy works for PLS only and consists of calibrating it directly on the bulk samples instead of the bulk subsamples. Either strategy will lead to approximately equally good performance, and both strategies lead to much more accurate mean protein content regression than just calibrating on the bulk subsamples using the mean reference values.

Additionally, this report shows that calibrating models using a grain density of at least 10 percent is feasible for both protein regression and grain type classification using both CNNs and PLS. We have shown that performance increases with grain density for both protein content regression and grain type classification. However, compared with previous work, we have shown that for grain variety classification, including low grain density samples in the model calibration leads to more accurate predictions on low grain density samples. Simultaneously, on high grain density samples, our classifiers are competitive with previous work that calibrated models only on high grain density samples.

\selectlanguage{english}
\bibliography{converted_to_latex.bib}
\end{document}